\documentclass[letterpaper, 10 pt, conference]{ieeeconf}  % Comment this line out if you need a4paper
\usepackage{amsmath}
\usepackage{graphicx}
\usepackage[obeyspaces]{url}
\usepackage{algorithmicx}
\usepackage{algorithm}
\usepackage[noend]{algpseudocode}

\IEEEoverridecommandlockouts                              % This command is only needed if 
\usepackage{subfig}
\title{\LARGE \bf
Configuration-Space Flipper Planning for Rescue Robots
}

\author{Yijun Yuan$^{1}$, Letong Wang$^{1}$ and S\"oren Schwertfeger$^{1}$% <-this % stops a space
\thanks{$^{1}$Both authors are with the School of Information Science and Technology, 
ShanghaiTech University, China.
        {\tt\small [yuanwj, wanglt, soerensch]@shanghaitech.edu.cn}}%
}

\begin{document}

\setlength{\belowcaptionskip}{-5pt}

\maketitle
\thispagestyle{empty}
\pagestyle{empty}

%%%%%%%%%%%%%%%%%%%%%%%%%%%%%%%%%%%%%%%%%%%%%%%%%%%%%%%%%%%%%%%%%%%%%%%%%%%%%%%%
\begin{abstract}
For rescue robots, flipper endows the robot with additional ability to pass through various terrain. Autonomous motion becomes more important. In recent work autonomy is done by either planning with several special states or based on collected data. We are considering if it is possible to find a way to build continues states without collecting old trail data. In this paper, we first model the possible states as a global planning path with parameter configuration of the scene. Then, we follows the path to achieve the autonomous run. We plot the morphology of each path points to show the correctness of the path and implement a simple path following on real robot to demonstrate the performance of our algorithm.
\end{abstract}

%%%%%%%%%%%%%%%%%%%%%%%%%%%%%%%%%%%%%%%%%%%%%%%%%%%%%%%%%%%%%%%%%%%%%%%%%%%%%%%%
\section{INTRODUCTION}
%0. Rescue robot
In robotics, one of the most important mission is to mitigate after disaster happened. Robots have to possess various abilities to tackle the hostile environment. The RoboCup Rescue competition, that has been held 19 years since 2000, provides a stage to evaluate the potential of rescue robots. It aims at assessing the robot capabilities. Various arenas, such as Curb or Stair Debris were designed to test the maneuvering and mobility of the robots \cite{sheh201616}.

In RoboCup, most of the teams are using tracked vehicles. As discussed in \cite{mourikis2007autonomous}, tracked robot has its own advantages. Comparing with wheeled robot, it has a larger ground contact surface. And tracked robot would be more stable for the lower center of gravity than bipeds. 

%1. flipper robot 
For the maneuvering and mobility, it is a popular strategy to mount flippers on rescue robots to increase its ability to pass trough various terrain. The most commonly designs used in research consists of front (and back) subtracks mounted on the main body with rotary joints \cite{steplight2000mode}\cite{mourikis2007autonomous}\cite{ mihankhah2009autonomous}\cite{endo2017stair}\cite{pecka2017controlling}\cite{wiley2016planning}. There are also people working with reconfigurable robots \cite{liu2005analysis}, transformable robots \cite{li2012online}, four tracked robots \cite{vu2008autonomous} and others. Those more specially designed robot are not majorly involved in research, competition and product. In this work, we utilize our algorithm on the robot shown in Fig. \ref{fig:robot}, which is following the common flipper design principle.

\begin{figure}[tpb!]
	\centering
	\subfloat[]{
		\label{fig:origin}
		\includegraphics[width=0.85\linewidth]{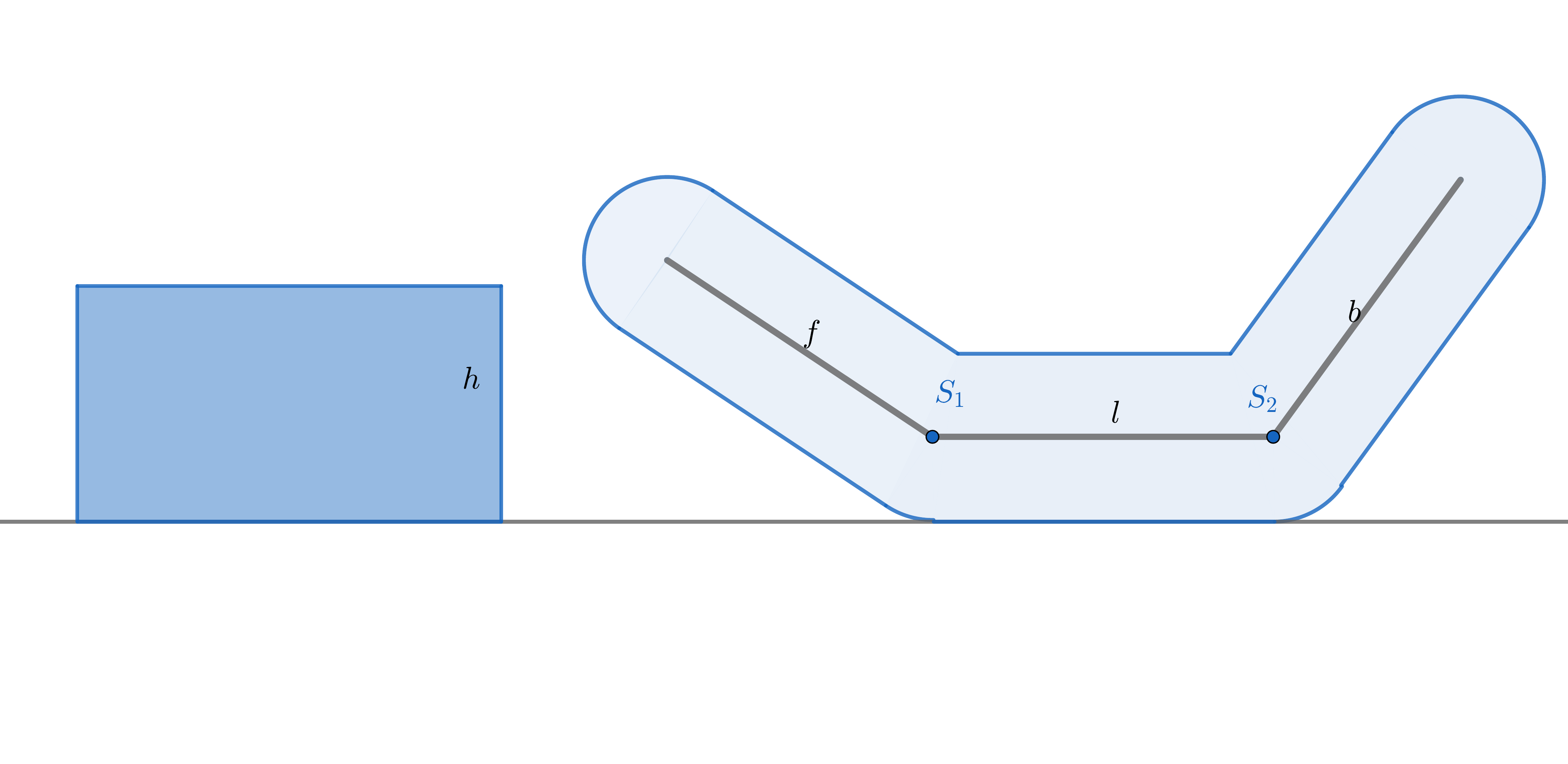}}\\
	\subfloat[]{
		\label{fig:simplify}
		\includegraphics[width=0.85\linewidth]{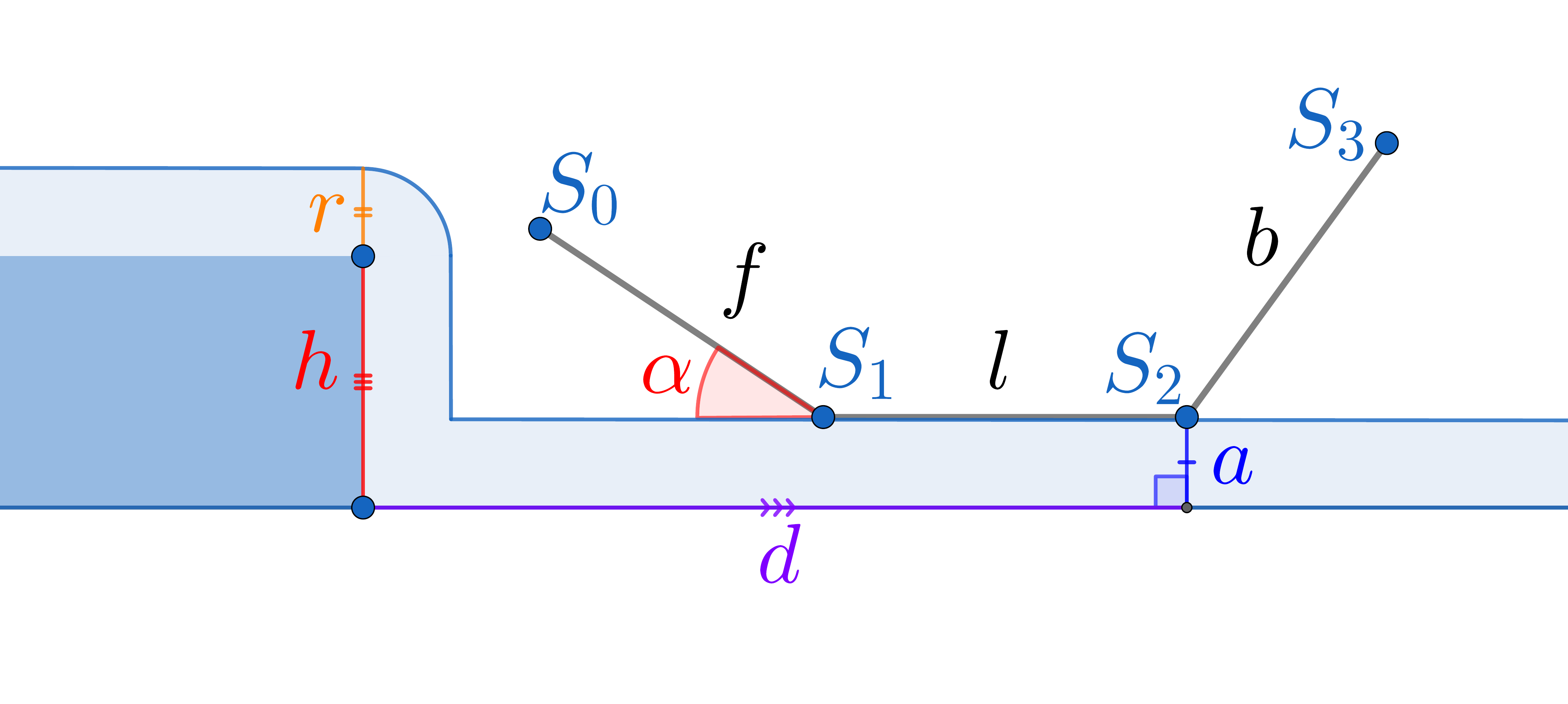}}
	\caption{Simplification of robot-ground scene to skeleton-dilated ground scene.}
	\label{fig:simplification}
\end{figure}

However, the additional operation of the flippers increases the complexity in controlling the robot. 

%2. autonomously run(the review of planning paper)
Tele-operating can be a choice to direct the robot moving. The problem is, when the communication quality can not be guaranteed, it is urgent to ensure the robot can run autonomously.
Running autonomously is an important ability for rescue robots.

In this work, we focus on the algorithms to generating a global plan for tracked robots with flippers.

Tracked robots have been explored in many papers. \cite{martens1994stabilization} note the difficulty of climbing stairs for tracked robots for the limits on operator's feedback. Then batches of works have attempted the tackle the stair climbing  problem. \cite{steplight2000mode} relies on sonar, monocular camera and two-axis accelerator, proposing a sensor fusion model. In \cite{xiong2000vision}, robot's heading can be determined by only using monocular vision. After that, \cite{mourikis2007autonomous} propose an algorithm that can performing robustly in real world scenes without the pre-assumption of the stair geometry, interaction between wheel and stair surface or the light condition. Though, the planning of subtracks is not mentioned much in those work. It does not play out the potential of flippers. Actually, the flipper planning algorithms are either with state machine adopt the task with several  simple morphology or data driven by pre-collecting trails. 
\cite{colas20133d} group the configuration of flippers into four postures and make execution on its corresponding situation. \cite{zimmermann2015adaptive} utilize reinforcement learning (RL) to accommodate the morphology to the terrain.  Sequentially, \cite{pecka2016autonomous} further enhance the capability of control with Relative Entropy Policy Search. Make effort on the observation, \cite{pecka2017controlling} model the incomplete measurement and make control on the robot morphology under RL. Similarly, \cite{sokolov2017hyperneat} describe a detailed implemented framework to learn the mobility from simulation experiments using deep neural network. \cite{wiley2017machine} learn the effect of action and make plan on reconfiguration of tracks to tackle various obstacles.

%3. our design
Our work is inspired by \cite{wiley2017machine}, that model the relation of angles for body and flipper with the extended qualitative model on climbing a step. Thus we are considering that it should be possible to model the scene and robot as a whole with parameters that are related to the obstacle. So in this paper, we model the step climbing for a tracked rescue robot with flippers as a global search, with each point in the path a possible morphology of the robot.

As in Fig. \ref{fig:simplification}, suitable simplification is taken from robot-ground scene to skeleton-dilated ground scene. This simplification takes benefit of our robot design and will be discussed in the next section. Other similar robots can also use this method.

Next we present the interrelation with several configuration parameter, with various restrictions to constraint configuration to be a possible setting of robot on ground. In this way, we can collect the whole space of our parameters and make planning over it.

Details of our algorithm can be found in Section \ref{sec::method}.  In Section \ref{sec::experiment}, we evaluate our method with several metrics. After that, in Section \ref{sec::conclusion}, we conclude this work and reemphasize our contribution.

\section{Methodology}
\label{sec::method}

In this section, we will introduce our method in detail. It consist of simplification and representation, global planning and path following. Please note, our robot is standing straight to the step, that is, the initial orientation is perpendicular to the vertical plane of the step. While this is a great restriction of our algorithm, the experiments will show that our approach works well even when this condition is violated.

\subsection{Represent the Model with Parameters}
\label{sec::representation}
%simplification
\begin{figure}[tpb]
	\centering
	\includegraphics[width=0.63\linewidth]{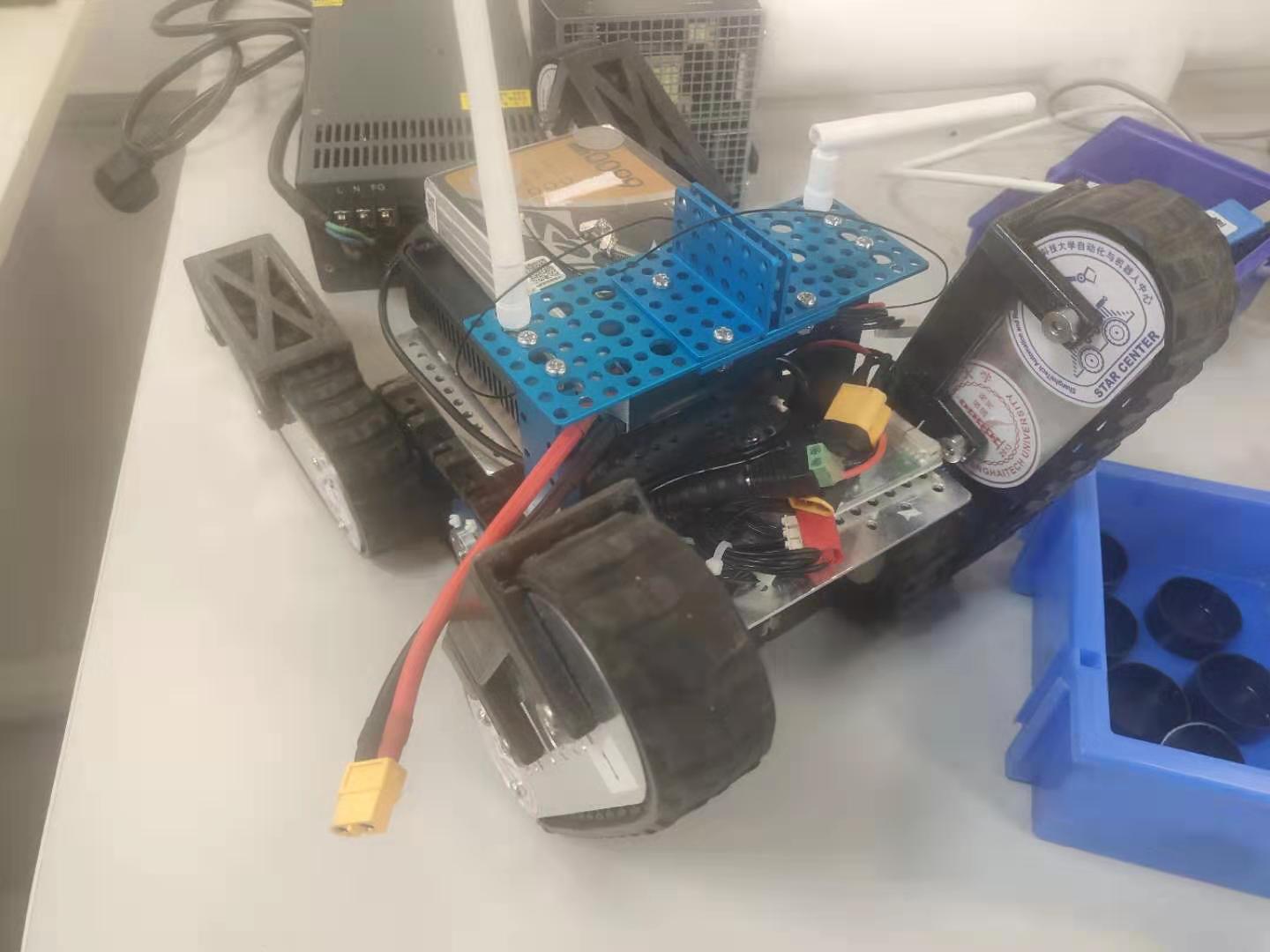}
	\caption{Rescue Robot of MARS lab, SIST.}
	\label{fig:robot}
\end{figure}

This work is derived from our rescue robot shown in Fig. \ref{fig:robot}. Its side view is as in Fig. \ref{fig:origin}. To make things clear, we draw lines to connect the neighbor motor joints and call those lines the skeleton. We observe that the closest distance from skeleton to track is always the wheel radius $r$. Taking the advantage of our rescue robot design, we can simplify the robot as its skeleton while dilate the ground and obstacle with track radius $r$ as in Fig. \ref{fig:simplify}. We denote the front and back joints as $S_1$ and $S_2$. The mobile base is line $S_1S_2$, the front and back flippers are line $S_0S_1$ and $S_2S_3$, respectively. We also call the quarter circle as curve and it will be used in following parts.

Then we represent the possible states of the robot in the world with three parameters $d$, $a$ and $\alpha$ as plotted in Fig. \ref{fig:simplify}. Where $d$ is the horizontal distance from $s_2$ to the stair, $a$ is the vertical distance from $s_2$ to ground, $\alpha$ is the angle between line $S_2S_1$ and $S_1S_0$. We consider $\alpha$ to be positive if  $S_0$ is on the right side of ray $S_2S_1$ and negative otherwise.

Please note that those three parameters are adequate to represent the state since the back flipper angle and angle of elevation can be uniquely determined given corresponding $d$, $a$ and $\alpha$, which will be discussed in Section \ref{sec::local}.
%parameters

\subsection{Global Planning for Rescue Robot}
\label{sec::global}

Actually, the main idea for our Global Planning is using a function to determine if one configuration is available and search in this possible space.

The whole algorithm is shown in Algorithm \ref{alg::globalPlan} and its detail are discussed in following subsection.

\begin{algorithm}[h]
	\caption{globalPlan: Global Planning for Flipper} 
	\label{alg::globalPlan}
	\begin{algorithmic}[1]
		\Require $d_0$,  $r$, $h$, $\Delta d$, $\alpha_{lb}$, $\alpha_{ub}$
		\State Init list $l$
		\State $l_p\leftarrow pathSearch(d_0, \Delta_d, r,\alpha_{lb}, \alpha_{ub})$ \Comment{Section \ref{sec::pathSearch}, \ref{sec::stateCheck}}
		\For{$p$ \textbf{in} $l_p$}
		\State $d,a,\alpha$$\leftarrow$ $p$.
		\State $q\leftarrow recoverWholeParameter(d,a,\alpha)$ \Comment{Section \ref{sec::recover}}
		\State $l.append(q)$
		\EndFor
	\end{algorithmic}
	\Return $l$
\end{algorithm}

%state check
\subsubsection{State Check}
\label{sec::stateCheck}
State check is a function that can return the feasible alpha angles by feeding certain $d$, $a$ pairs.

In our method, for safety, we assume there are some rules that the morphology should follow:

\begin{itemize}
	\item When the skeleton touches the curve part of surface and want to lift $S_1$ or $S_2$, it should be tangent to the curve.
	\item The base $S_1S_2$ should either touch the ground on $S_2$ or be tangent to the curve part of surface.
\end{itemize}

The first rule ensures that the obstacle will not block the movement of the robot. The second rule constrained the morphology into feasible group that won't slip down from step.

The main idea is that we can group morphologies into groups of states that are divided by several conditions. Then we treat each state separately. Those conditions are computed with $d$ and $a$.

For each pair of $(d,a)$, we can determine which group it belongs to. That is, we find the corresponding rough morphology for this pair. Then it will be easy to compute the valid range of $\alpha$ to obtain the accurate morphology.

%The detail can be found in Algorithm. \ref{alg::stateCheck}. To demonstrate the method more clear, we also plot the morphology in Fig. \ref{fig:geometry} that correspond to each condition and groups in Algorithm. \ref{alg::stateCheck}.

\begin{figure*}[tpb!]
	\centering
	\includegraphics[width=0.9\linewidth]{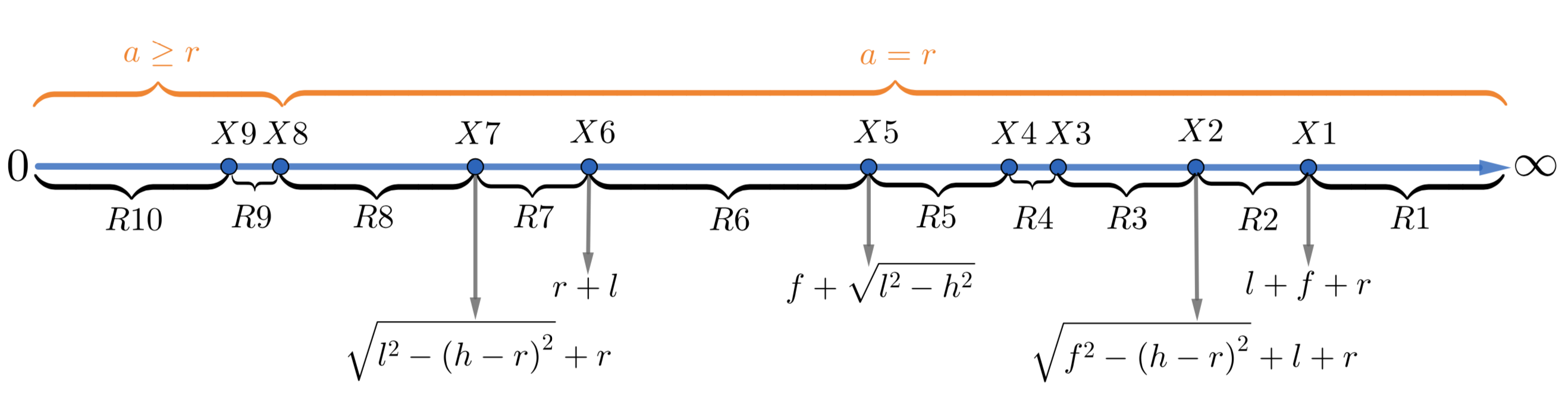}
	\caption{Possible states on axis of d. From left to right is small d, 0, to large.}
	\label{fig:pipeline}
\end{figure*}

In this algorithm we traverse $d$ from larger value to smaller ones. As shown in Fig. \ref{fig:pipeline}, the d's space is represented by an axis, where $X$s are critical points and $R$s are states between critical points. For each $(d,a)$, we check its relation to $X_i$ in ascend order of $i$ to distribute it to corresponding state $R$.

In our design, we want to ensure a stability of robot to avoid the morphology that is possible to slip down of the stairs as in Rule 1. So we pre-assume the flipper ($S_0S_1$) should be tangent to the curve if $S_1$ leave the ground. The body ($S_1S_2$) should be tangent to the curve if $S_2$ leave the ground.

%%%%%%% Letong Wang %%%%%%%%%%%%%
From the beginning, $X1$ is the critical point where $d_{X1}=f+r+l$. If the robot moves farther from the step (in state $R1$) then $d_{X1}$, its flippers could be free within their bounded range. If it moves closer (in R2 to R10), the flippers' movement will under some constrains, for example, the front flipper will be able to touch the step.
%TODO If it moves closer (in R2 to R10), the flippers' movement will  DO WHAT?? under some constrains. 

$X2$ is the critical point where the front end($s_0$) of the flipper $f$ just on the bottom end of the curve. If the robot is further, it will hold a minimum $alpha$ that $s_0$ will touch the vertical plane of the step (in state R2). If the robot moves little bit closer, $s_0$ will be possible to touch the curve (in state R3).

$X3$, as an important division, indicates that the further morphology should not lift the $S1$, since the front flipper is not tangent to the curve at all. As robot moves forward to step, $S1$ will possibly rise up as tracks climbs the stair.

Then $X4$ is utilized to cut the space that is with closer $d$ than $d_X3$. It is such a state that with straight line $S_0S_2$ cut curve on $S_0$. While not adequate to cut, it should be $R_4$ that is with a lower bound of $\alpha$ as in Fig. \ref{fig:R4}. Otherwise it will be $R5$ with a lower bound of $\alpha$ as in Fig. \ref{fig:R5} if it is not as close as $X5$.

$X5$ is a critic point that $S_0$ be able to touch the top end of curve with flipper horizontally. It is defined as the lower bound of $\alpha$ for following states, such as Fig. \ref{fig:R6R7R8LowerBound}. Then $X6$ with $d_{X6}=r+l$ set the critical point to indicate whether $S_1$ can still touch the ground. In $R6$ that is the farther region, sure, it can, and thus upper bound of $\alpha$ won't be restricted by obstacle. In $R7$, however, the upper bound is the case $S_1$ touch the vertical plane of stair while flipper perpendicular to the ground.

Keep moving forward, we will reach $X7$ that its $S_1$ on the bottom of curve. It shows the first time robot body touches the stair edge. Between $X7$ and $X8$ is the region that $S_1S_2$ won't be able to be tangent to curve. Stop here the $S_2$ won't lift from ground for the non-tangency of body $S_1S_2$ to curve. While $X8$ make a big difference that its following state are all for $a\geq r$ and the body will be always tangent to the curve.

$X9$, that both $f$ and $l$ of it are tangent to the curve, with $f$ lying on the upper plane of the step, sets a division for following state of $X8$ into $R9$ and $R10$. 

For $R9$ that is with $S_1$ lower than $h+r$ as in Fig. \ref{fig:R9}, we constraint it should also have its front flipper touch curve for stable issue. While $R10$ with higher $S_1$ than $h+r$ should have its $S_0$ touch the stair plane to avoid the sharp drop caused by gravity. 

To note that, when $h$ is high, $9.5cm$ for example, it is possible that $d_{X3}<d_{X4}$. And thus the $R4$ will disappeared or say, be part of $R3$. But we are considering the duty of $X3$ is to indicate whether $S_1$ can leave the ground. So the $R4$ that is in $R3$ should also follow the rules in $R3$.

Actually, we check the state with critical points from $X_1$ to $X_9$ in order, just to ensure the duty of critical point will carry out.

\newcommand\blaSize{0.2}
\begin{figure*}[tpb!]
	\centering
	\subfloat[R1]{
	\label{fig:R1}
	\includegraphics[width=\blaSize\linewidth]{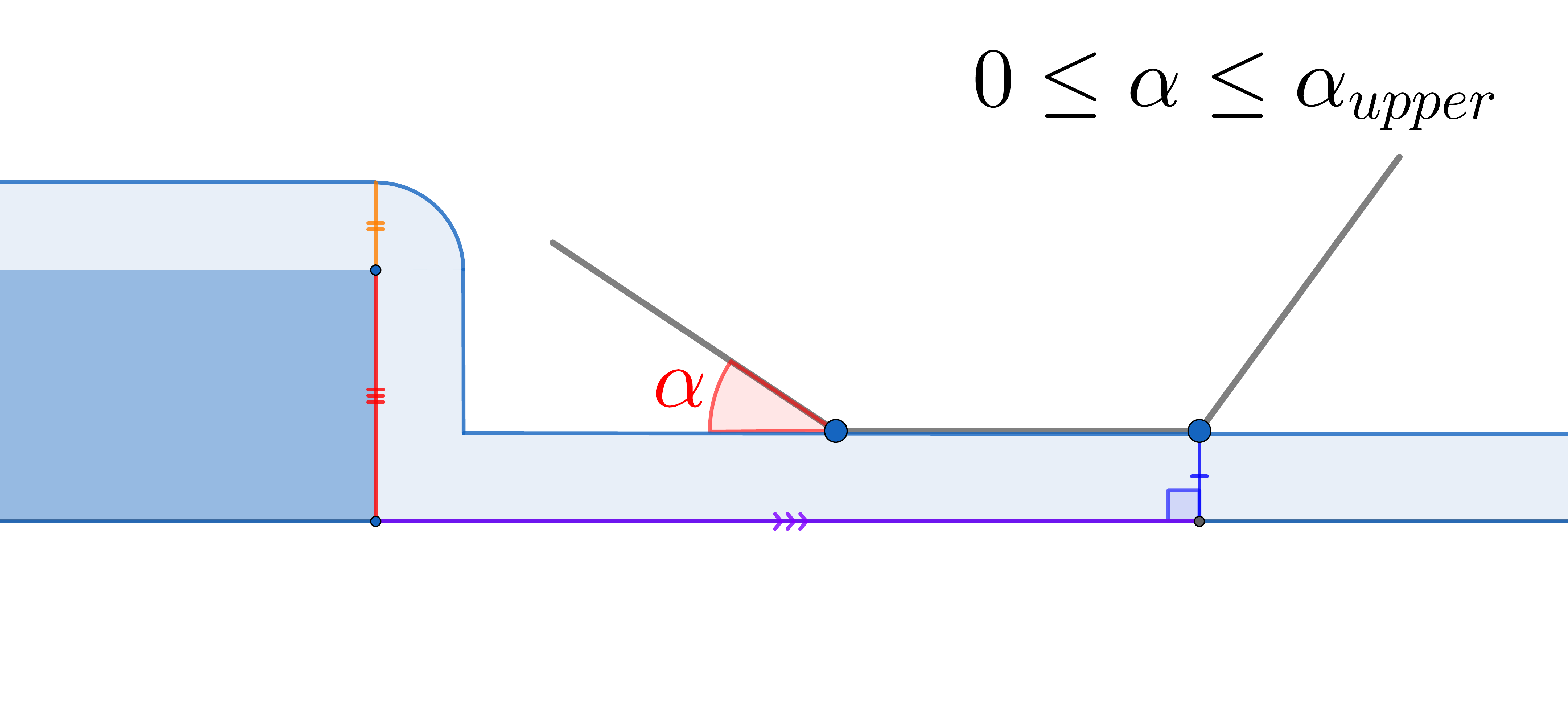}} 
	\subfloat[R2]{
	\label{fig:R2}
	\includegraphics[width=\blaSize\linewidth]{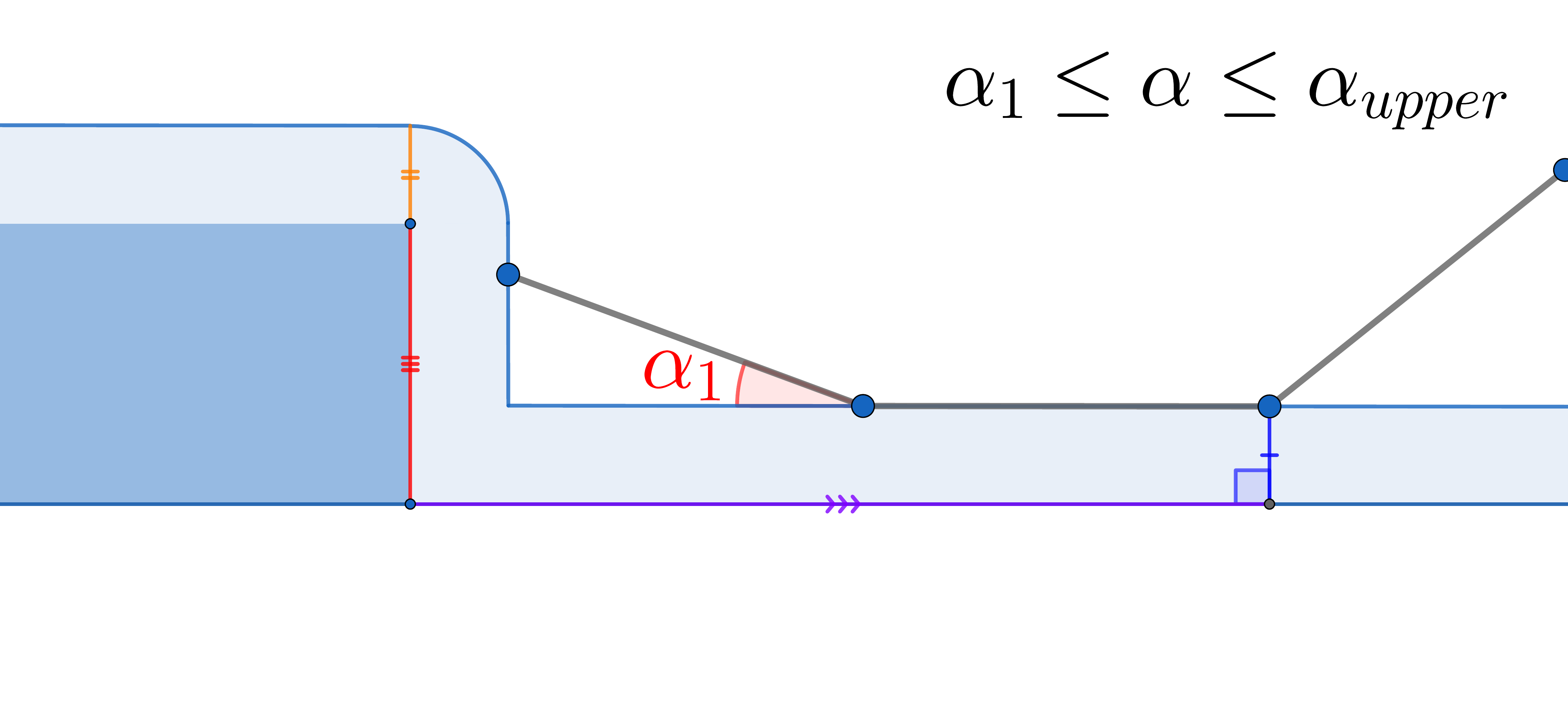}}
	\subfloat[X2]{
	\label{fig:X2}
	\includegraphics[width=\blaSize\linewidth]{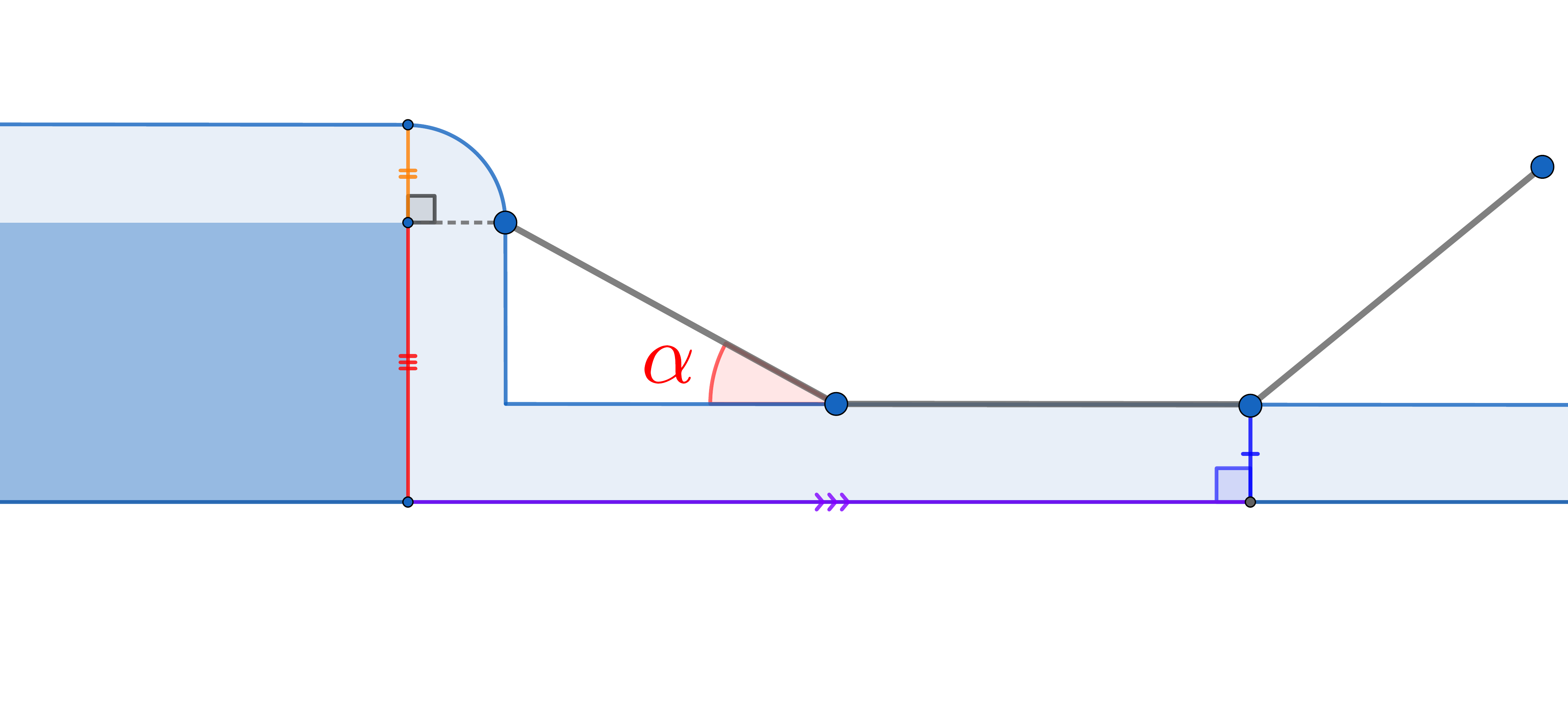}}
\subfloat[R3]{
	\label{fig:R3}
	\includegraphics[width=\blaSize\linewidth]{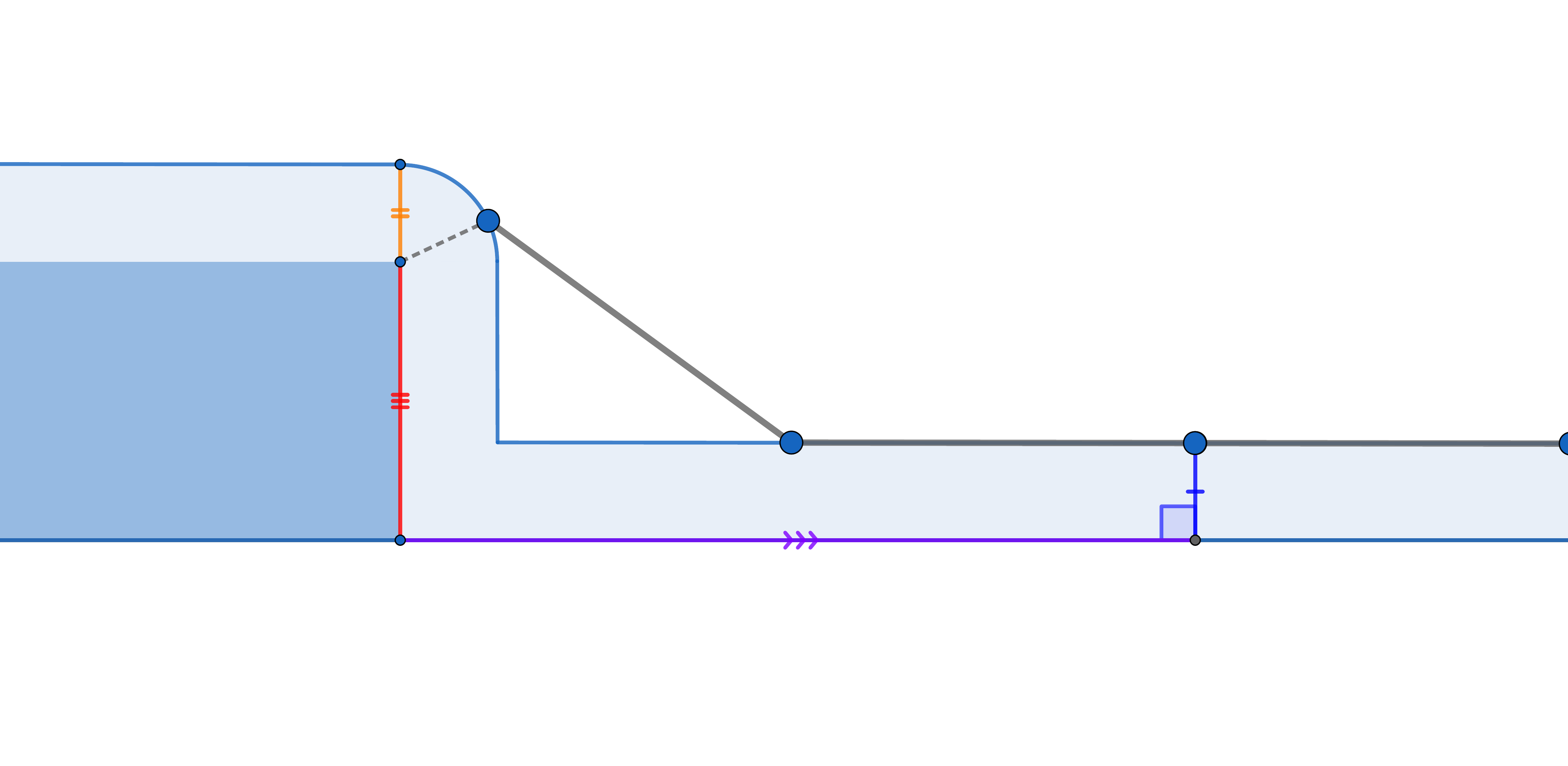}} 
\subfloat[X3]{
	\label{fig:X3}
	\includegraphics[width=\blaSize\linewidth]{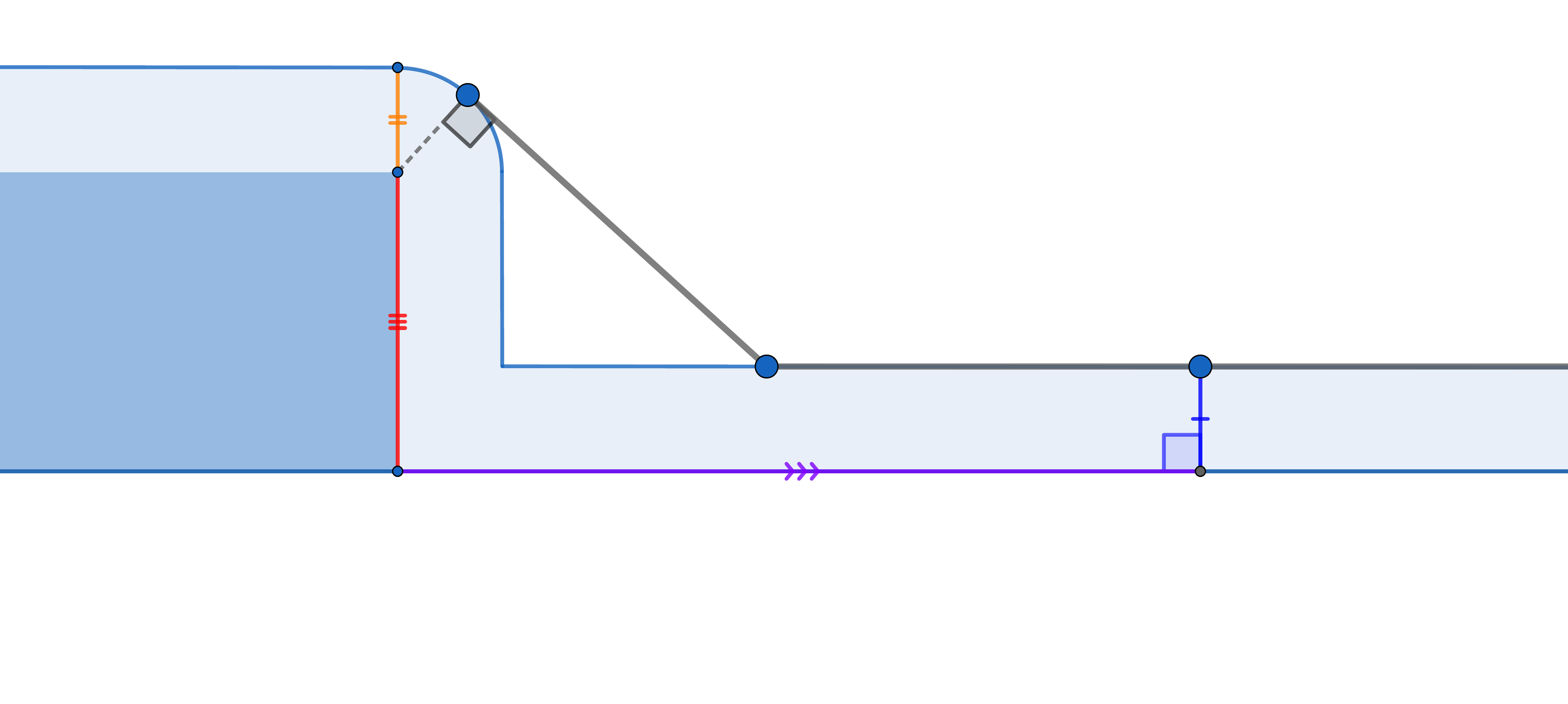}}\\
\subfloat[R4]{
	\label{fig:R4}
	\includegraphics[width=\blaSize\linewidth]{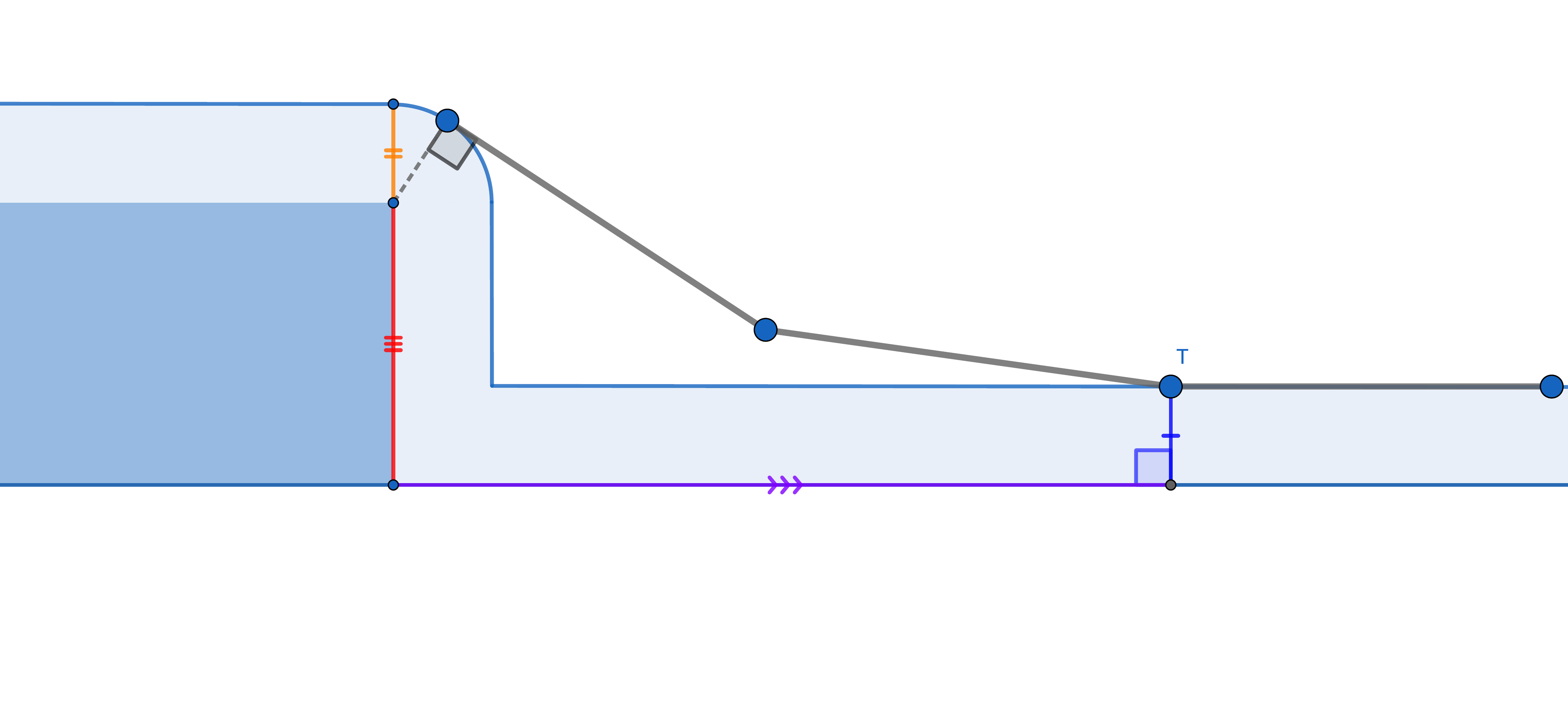}}
\subfloat[X4]{
	\label{fig:X4}
	\includegraphics[width=\blaSize\linewidth]{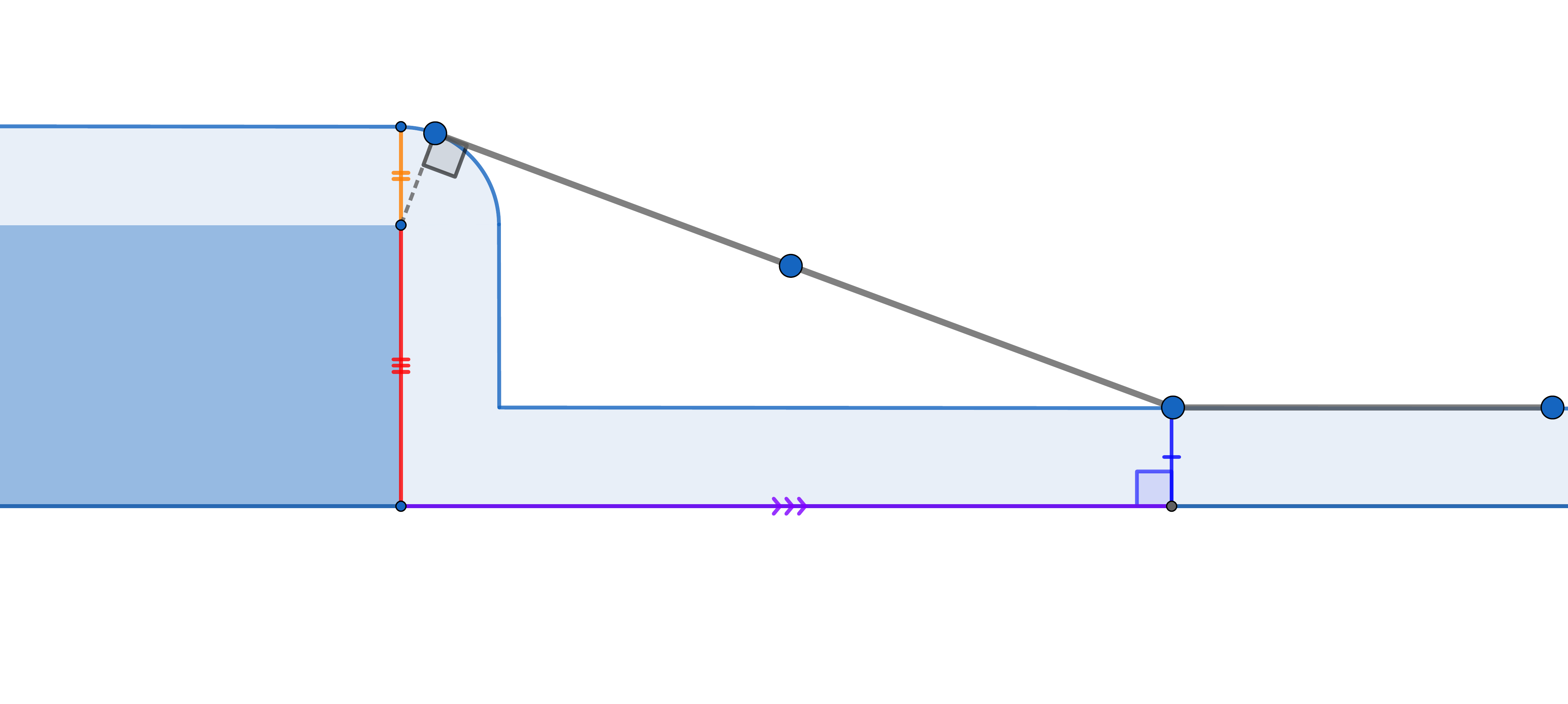}} 
\subfloat[R5]{
	\label{fig:R5}
	\includegraphics[width=\blaSize\linewidth]{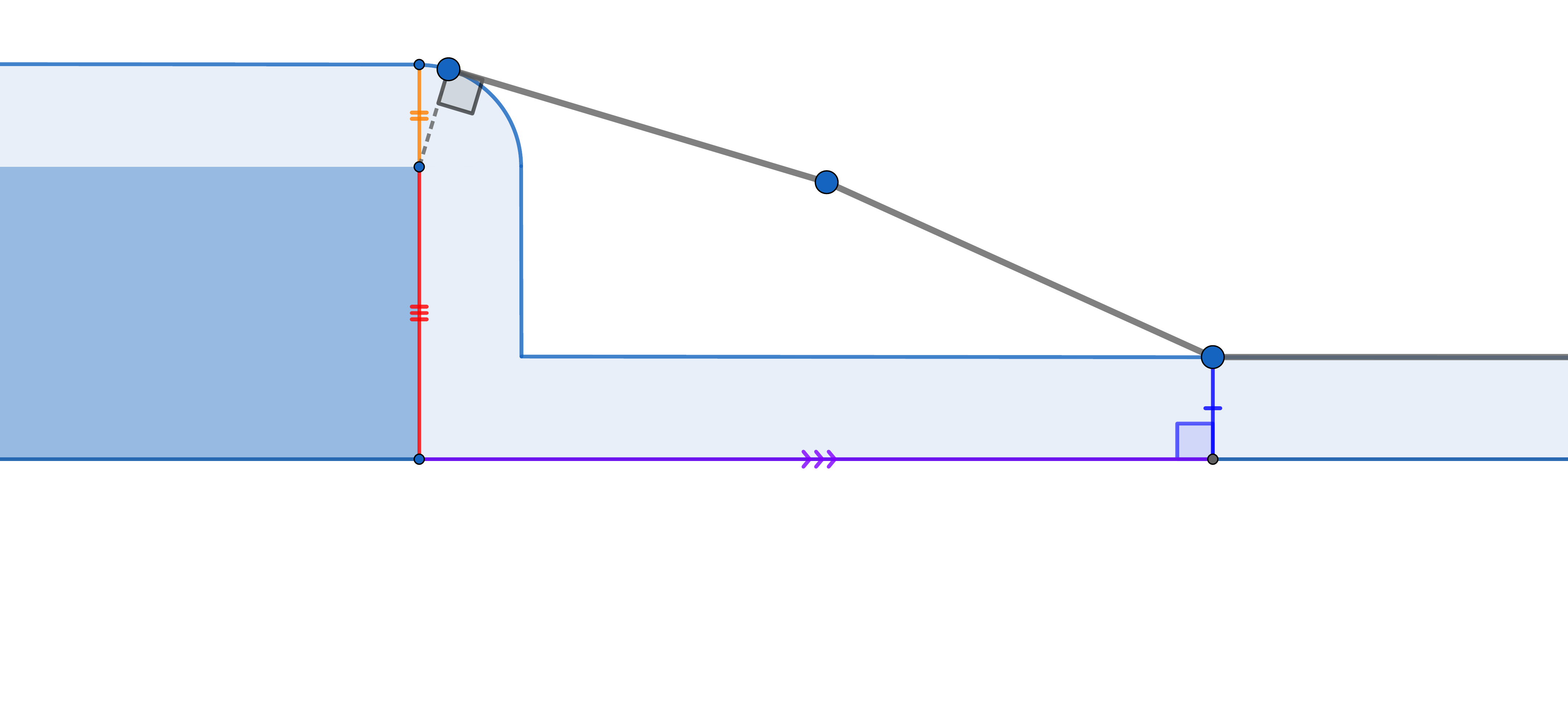}} 
\subfloat[X5]{
	\label{fig:X5}
	\includegraphics[width=\blaSize\linewidth]{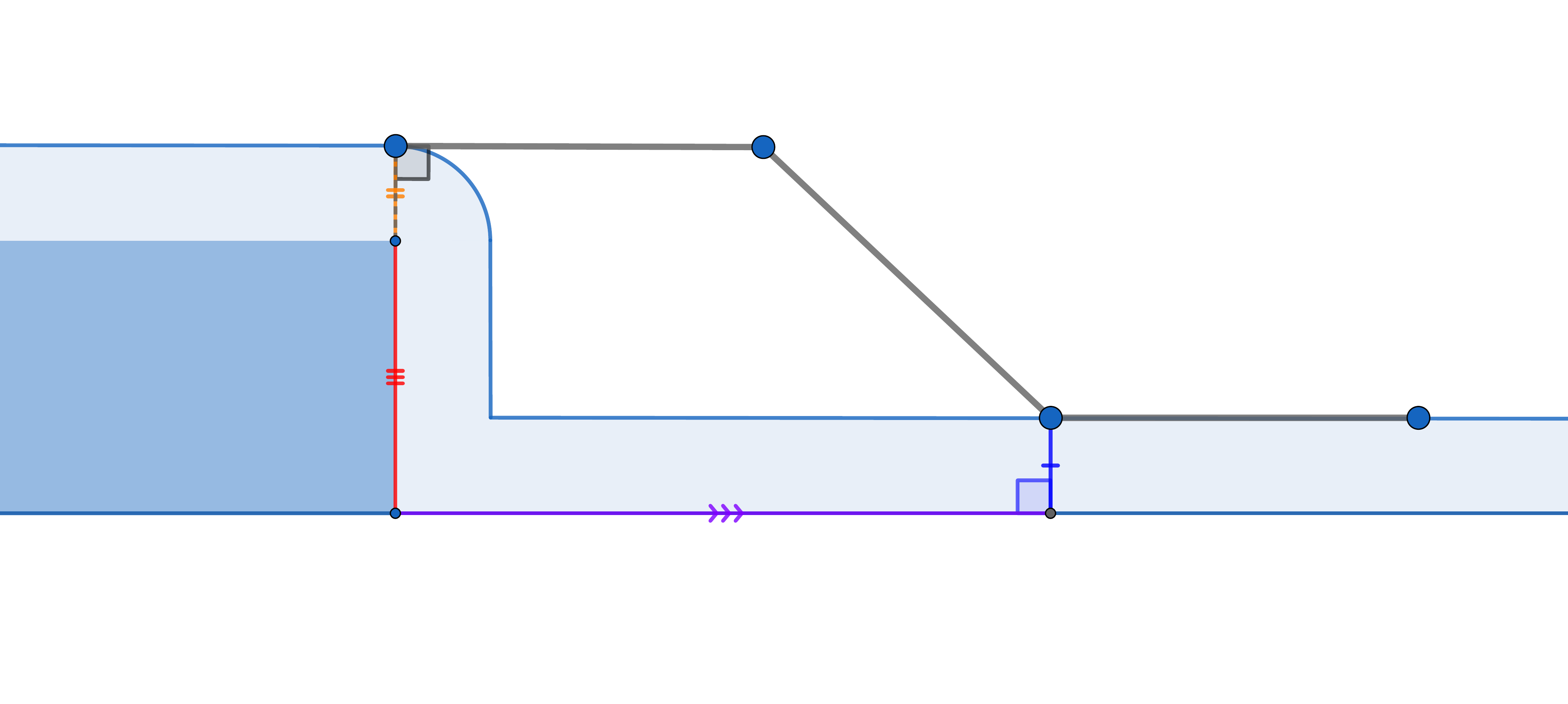}}
\subfloat[R6 R7 R8 Lower Bound]{
	\label{fig:R6R7R8LowerBound}
	\includegraphics[width=\blaSize\linewidth]{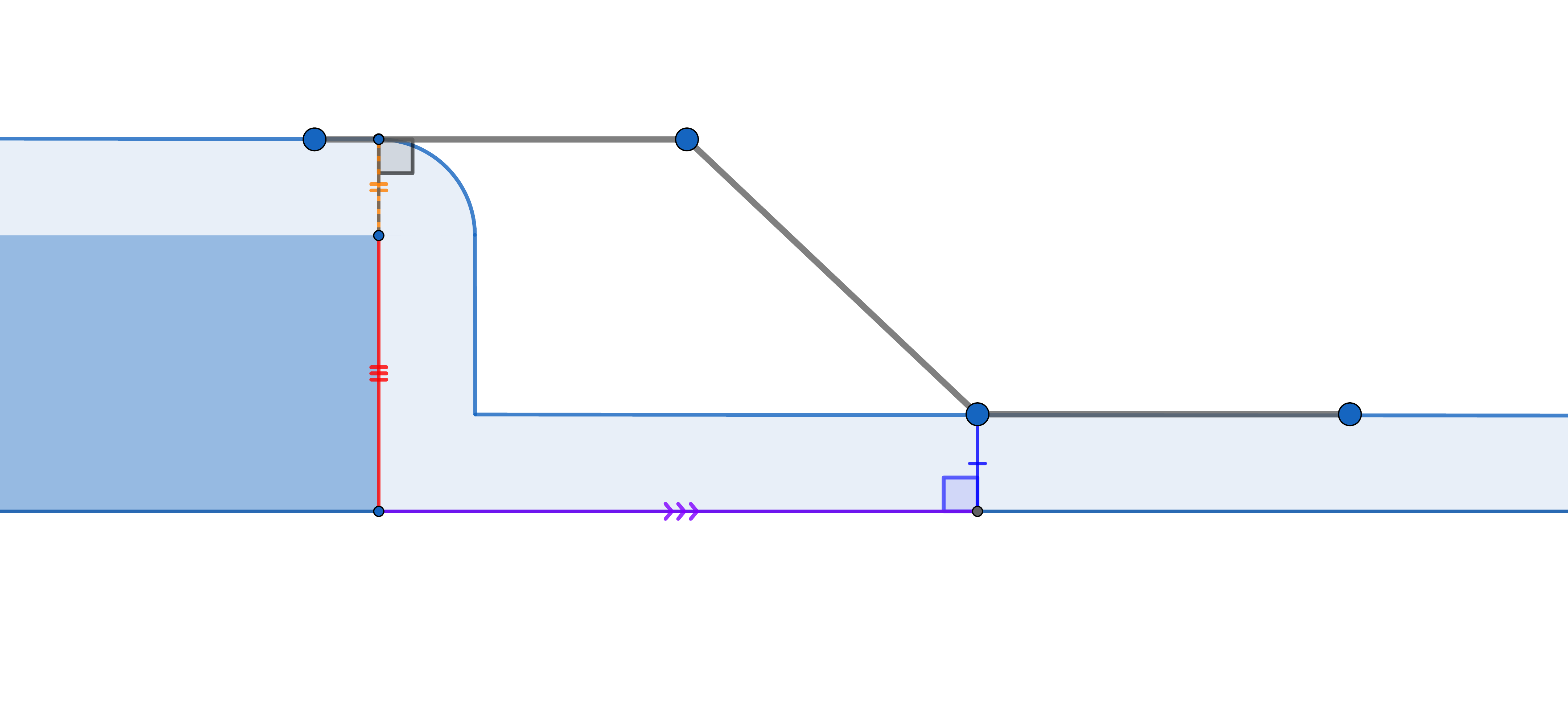}}\\
\subfloat[R7UpperBound]{
	\label{fig:R7UpperBound}
	\includegraphics[width=\blaSize\linewidth]{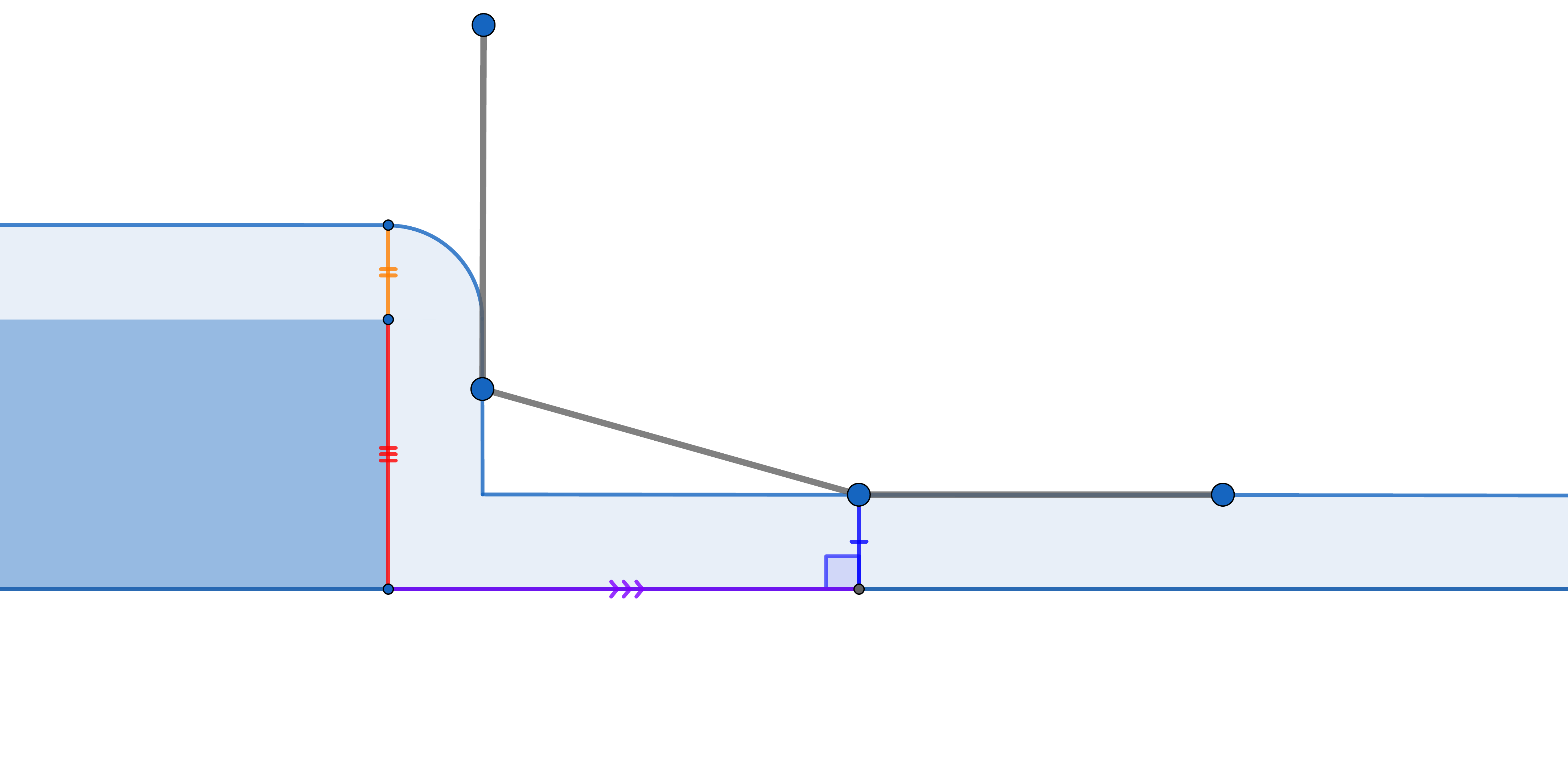}} 
%\subfloat[R7LowerBound]{
%	\label{fig:R7LowerBound}
%	\includegraphics[width=\blaSize\linewidth]{im/geometry/R7Lower.pdf}}
\subfloat[X7]{
	\label{fig:X7}
	\includegraphics[width=\blaSize\linewidth]{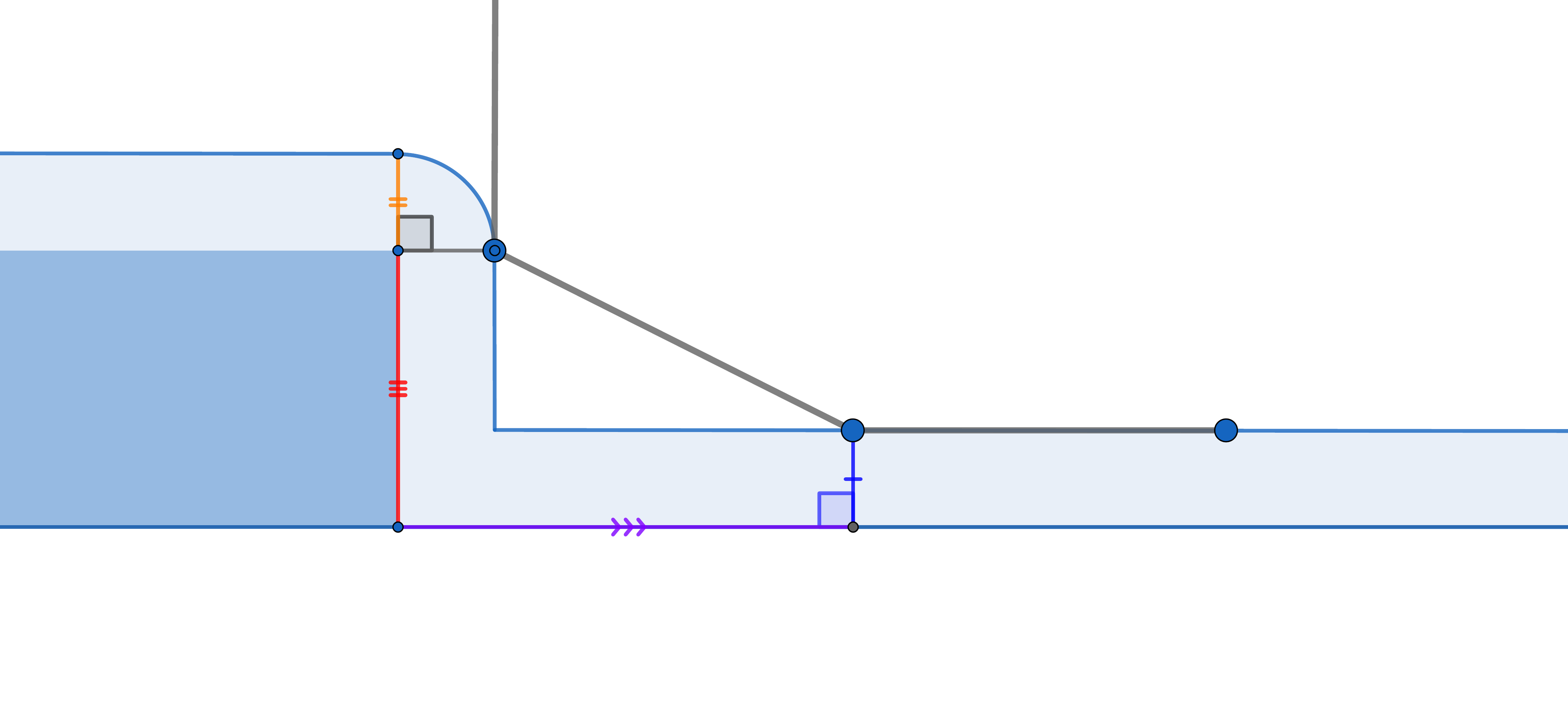}}
\subfloat[R8]{
	\label{fig:R8}
	\includegraphics[width=\blaSize\linewidth]{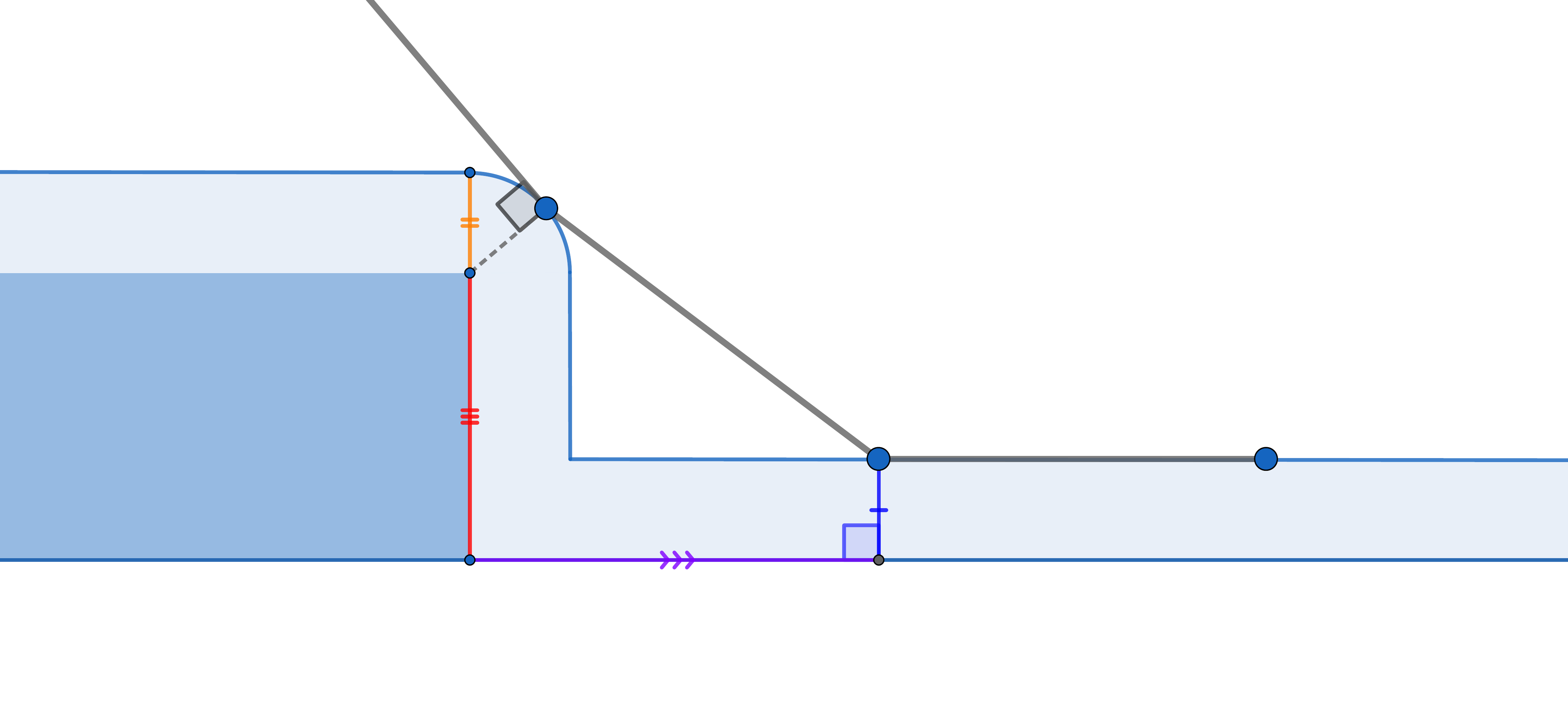}} 
\subfloat[X8]{
	\label{fig:X8}
	\includegraphics[width=\blaSize\linewidth]{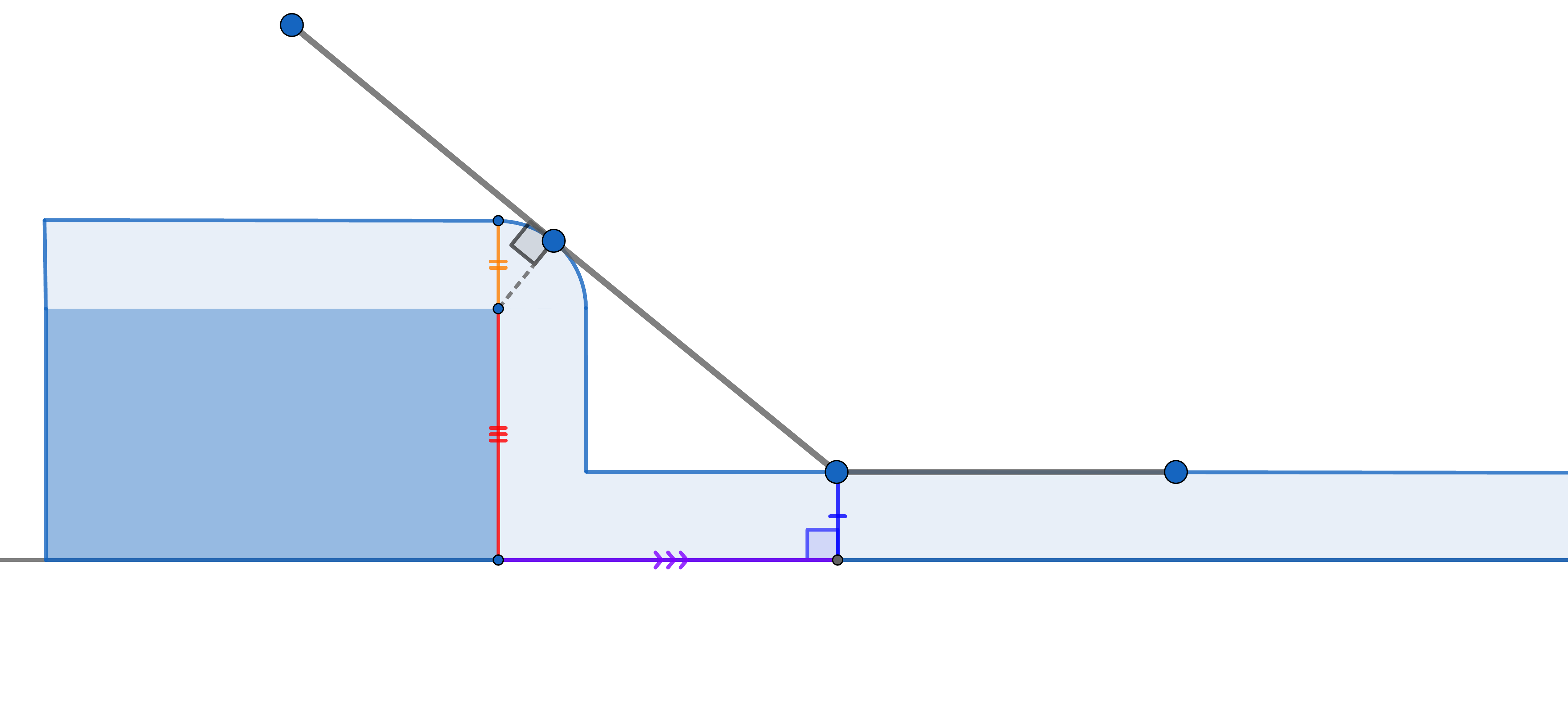}}
\subfloat[R9]{
	\label{fig:R9}
	\includegraphics[width=\blaSize\linewidth]{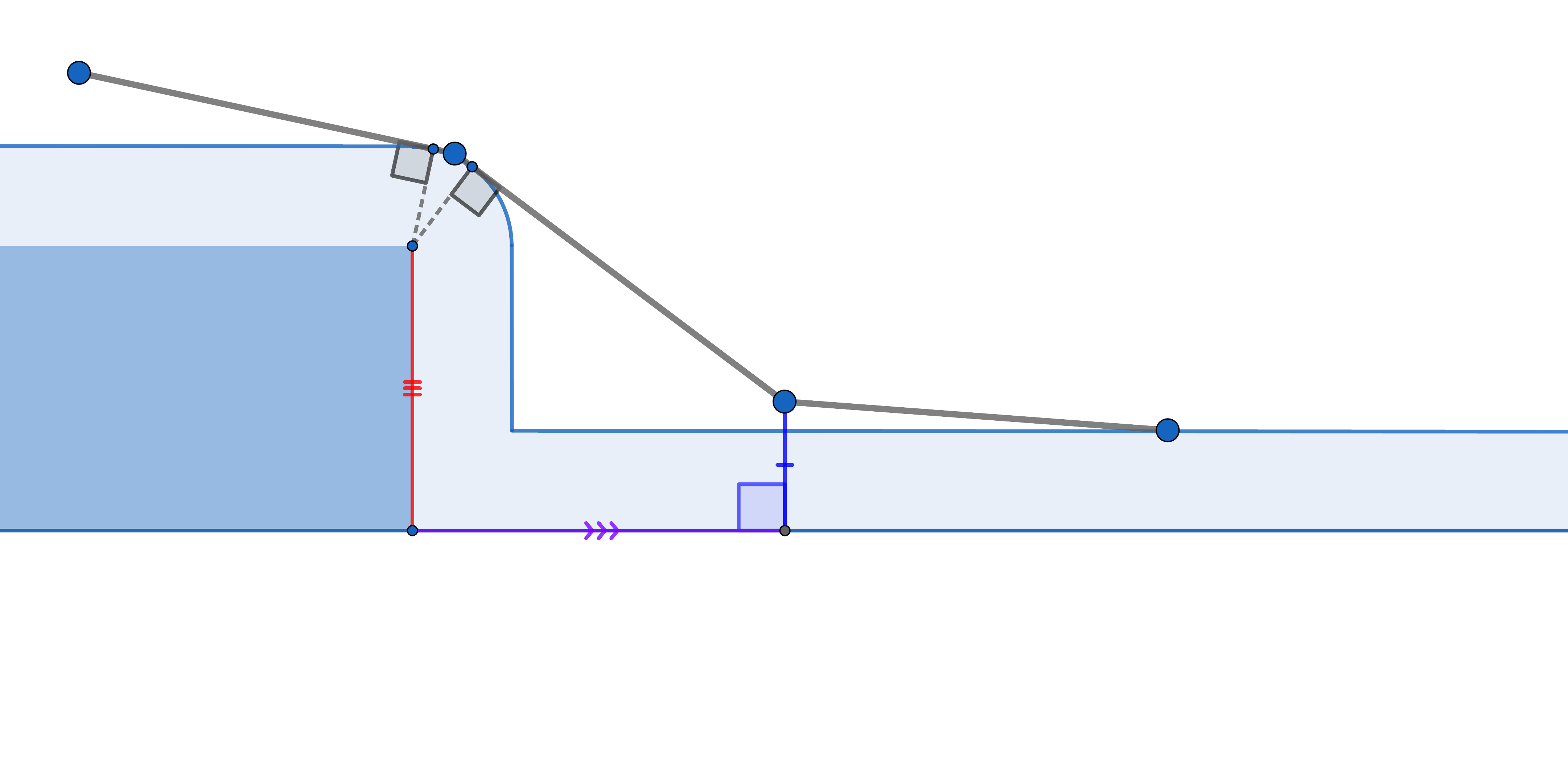}}\\
\subfloat[X9]{
	\label{fig:X9}
	\includegraphics[width=\blaSize\linewidth]{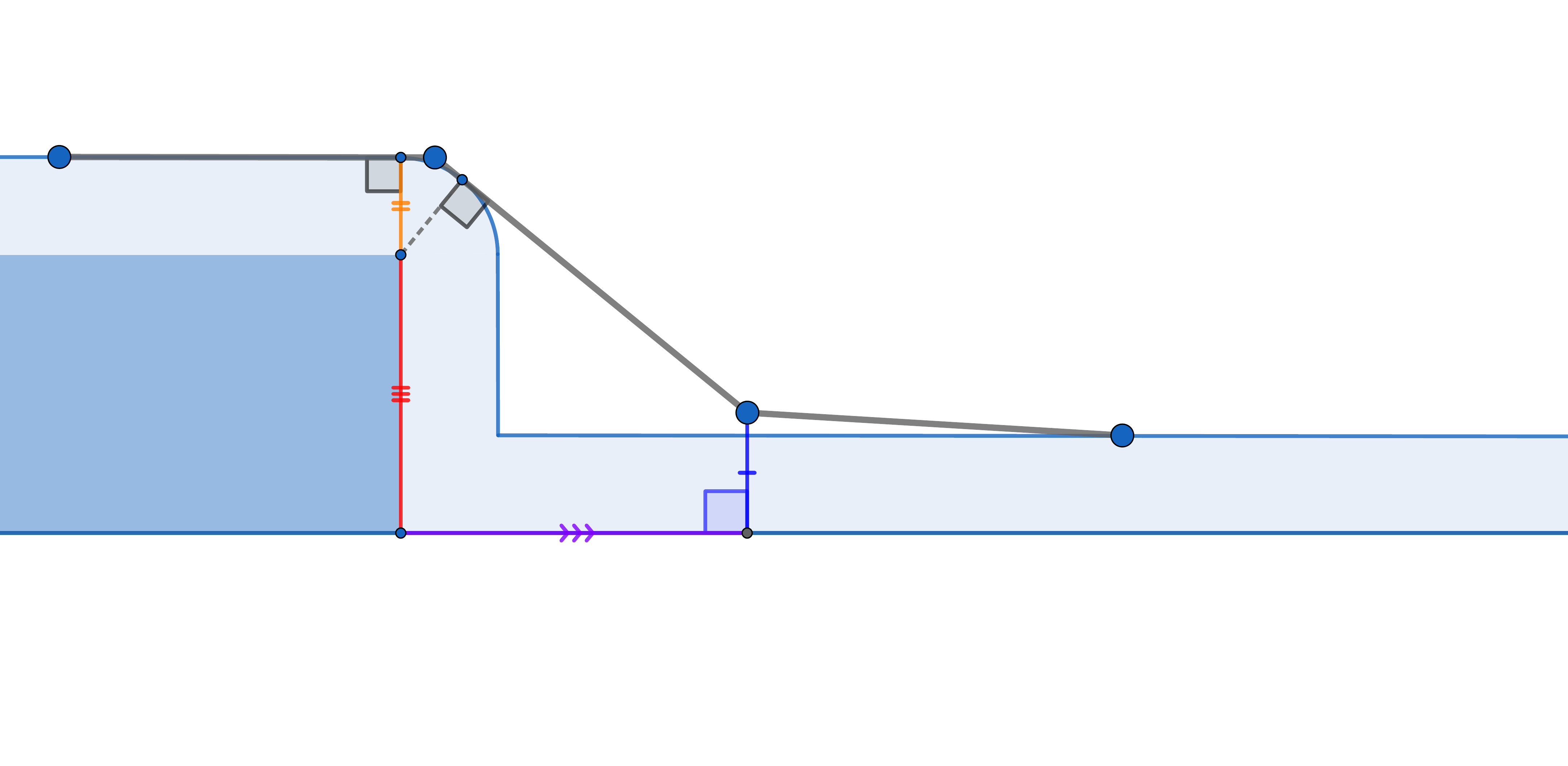}}
\subfloat[R10]{
	\label{fig:R10}
	\includegraphics[width=\blaSize\linewidth]{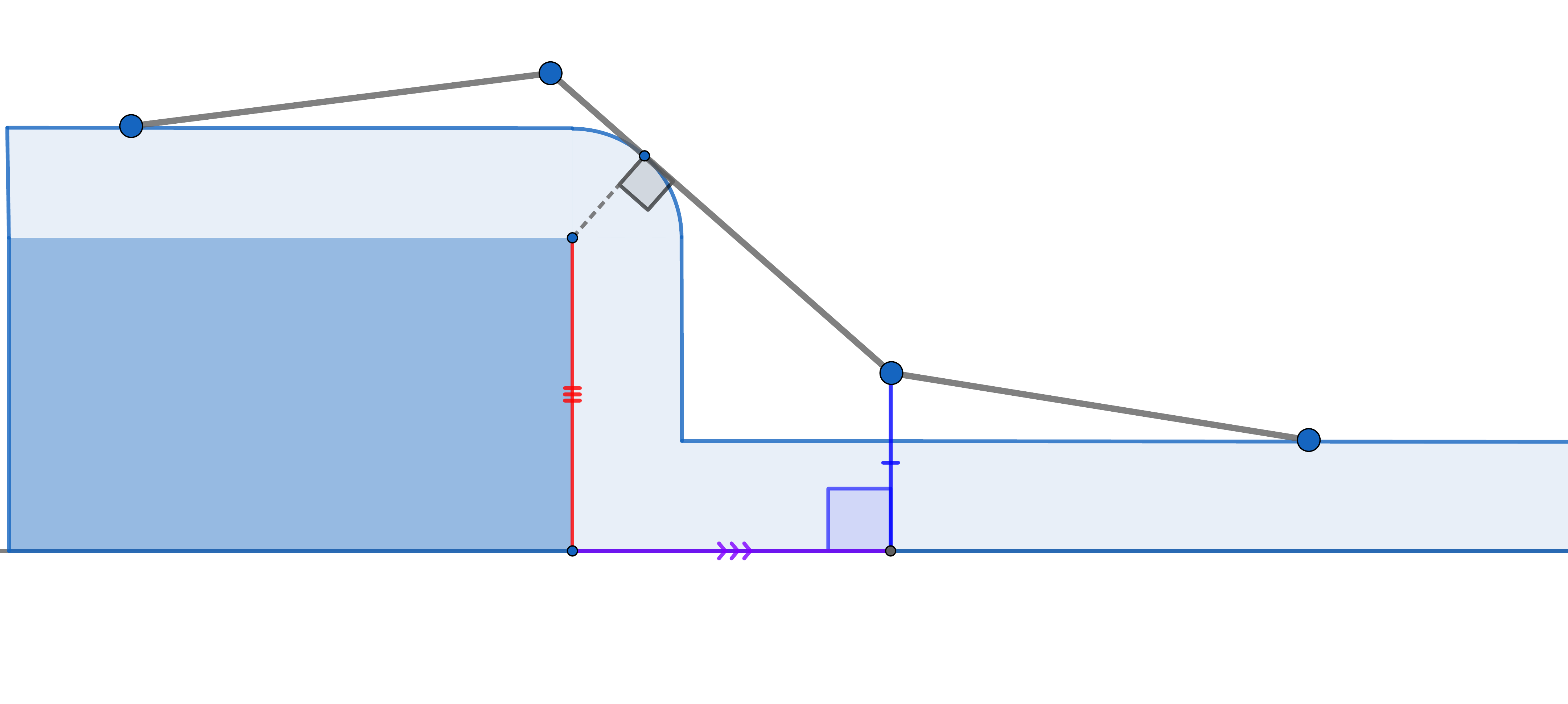}}
	\caption{Morphology of critical points and states in Fig. \ref{fig:pipeline}.}
	\label{fig:geometry}
\end{figure*}

One possible space can be found in Fig. \ref{fig:space}.

\subsubsection{Path Search}
\label{sec::pathSearch}
In this part, we do not build the whole space of parameters and make a search, because the space is large and it takes a long time to build the graph and search, which is not viable for rescue robot autonomous run. 

So alternatively we make a simple implementation. First, we discrete the $d$ from some triggered distance $d_0$ to $0$ with certain interval $\Delta_d$. Then from large to small, we find the possible state of next $d$ and find the closest triplet point to current point and update the current point. The path can be generated efficiently and will finally reach the target. It is more clear in Algorithm. \ref{alg::pathSearch}. The $FindAAlpha$ function used here take $d$ as input, and it will create batches of $(d,a)$ pairs with a range of $a$. Then we feed those batches of pairs into state check function as in Section \ref{sec::stateCheck}. The distance function we used here is \[dist(p_1,p_2) = (d_{p_1}-d_{p_2})^2 + \omega_a(a_{p_1}-a_{p_2})^2 + (\alpha_{p_1}-\alpha_{p_2})^2  \] with a weight $w_a$ to adjust the effect of $a$ on distance computing.

\begin{algorithm}[h]
	\caption{pathSearch: Path Search with State Check}
	\label{alg::pathSearch}
	\begin{algorithmic}[1]
	\Require $d_0$, $\Delta d$, r, $\alpha_{up}$, $\alpha_{low}$
	\State $p_{Current} \leftarrow (d_0, r, \alpha_{up})$
	\State Initialize path $l_p\leftarrow[p_{Current}]$
	\While{$p_{Current}[0]-\Delta d>0$}
	\State$ps \leftarrow FindAAlpha(p_{Current}[0]-\Delta d, d, r)$
	\State$p_{current} \leftarrow closest\_point(ps, p_{current})$
	\State$l_p.append(p_{current})$
	\EndWhile
	\end{algorithmic}
	\Return $l_p$
\end{algorithm}

\subsubsection{Recover the Whole Parameter}
\label{sec::recover}
From above parts, we achieved a path of triplet $(d,a,\alpha)$. However it is not adequate to reveal on the robot morphology. So here we further compute the back flipper angle $\beta$ and evaluate the robot $\theta$ and $l_t$, which is the distance from $s_2$ to 
its cut-off point. $l_t$ does not count if the base does not touch the curve part of dilated ground.

Given a triplet of $(d,a,\alpha)$, we will use state check similar to Section \ref{sec::stateCheck} to find the group of morphology it belongs and then compute its $\theta$ and $l_t$ correspondingly. For the back flipper angle, if $S_2$ does not leave the ground, we would set $\beta$ to make the whole back flipper on the ground after $X2$. Otherwise, we should ensure the end-point of it touch the ground if back flipper is not tangent to the curve. While back flipper is tangent to the curve, the $\beta$ will also be set correspondingly.

\subsection{Path Following for Rescue Robot}
\label{sec::local}
Since we have obtained the global path with each point a triplet of ($d$, $a$, $\alpha$) for a configuration, it is also important to make real robot move following the path.

For each triplet point, we send it into a similar state checking function shown in Section \ref{sec::stateCheck} to determine its current state and compute the corresponding extra parameters. 

In this part, we will also use these three additional parameters $\beta$, $\theta$ and $l_t$.

For the path following, we use $\Delta \alpha$, $\Delta \beta$ and $\Delta m$ to determine the wanted angle changing and track moving.

$\Delta \alpha$ and $\Delta \beta$ is easy to compute while $\Delta m$ requires more concern because we should also consider the touched point on the track.

While $a=r$, $s_2$ is still on the ground, the $\theta$ will effect the movement as positive $\Delta \theta$ will make $s_2$ move back a little bit. So we compute \[\Delta m = \Delta d - \Delta \theta\cdot r \]. When $a>r$, from our design, base should be on the stair edges. So we use $l_t$ to compute $\Delta m$ as \[\Delta m = \Delta l_t\].

\section{Experiment}
\label{sec::experiment}

\subsection{Setting}
%robot
The robot we use is shown in Fig. \ref{fig:robot}. It is a small size tracked robot. Its wheels on the track and flipper are with the same radius. Which means we can draw a line between joints and thus its structure fit for our design of simplification as in Fig. \ref{fig:simplification}. For this robot, the length of robot track, $l$ (from $S_1$ to $S_2$) is $14.5$ cm, the length of front flipper, $f$ (from $S_0$ to $S_1$) and of back flipper (from $S_2$ to $S_3$), $b$ are both $13.5$ cm. The radius of the robot wheel is $3.5$ cm.

%tracking system
To record the pose of the robot while moving to provide ground truth, a  tracking system, OptiTrack, has been utilized to locate the position of joints. vrpn server is activated with Motive software under the same network as robot.

%software
The implementation of our algorithm is with Python. We use ROS to coordinate with the robot. On the motor control, ROS dynamixel\_workbench\_control package has been utilized. To receive the true location from tracking system together with robot parameter, vrpn client runs under the same ROS master as the robot.

%param
During the experiment, the $\alpha$ lower bound is set $-90^{\circ}$ and upper bound $56^{\circ}$ that are restricted by the installation of motors. On the flipper planning part, we discrete $d$ with an interval $0.01$m, $a$ with interval $0.001$ and $\alpha$ with interval $0.01$ rad if result $\alpha$ for given $(d,a)$ pair is a range of value.

%arena
Because the upper bound of $\alpha$ dominates the height it can climb, so we first choose a $9.5$ cm step that is considered to be high. Then $6.7$ cm and $4.7$ cm are also included in the experiment. Because the start position $d_0$ does not effect the result of planning, we unified the initial distance to $0.4$ m for all cases. Also, we set $\omega_a$ in distance function as $100$.

This work is only analyzed on the straight to obstacle case. We are wondering if it is possible to climb on a step if it is not straight. For that we rotate the stair as the middle line to stair still $0.4$ m and run. The rotation angle $\Omega$ is from $5^{\circ}$ to $40^{\circ}$ with $5^{\circ}$ interval. 

The implementation can be found on github\footnote{\url{will be released if paper gets accepted.}}.

\subsection{Evaluation}
%experiment:
%1. 3D space with different d0, h
%2. path with various d0, h
%3. path with various principle
%4. diagram of alpha in each d step with various h
	%we have alpha<0 our method can avoid the sharp drop of the front of main body.
%5. tracking system to compare with the predicted theta, d, alpha
%6. human designed(fix angle), record its pose change, eg. elevation(our method should be small).

\begin{figure}
	\centering
	\subfloat[Space with interval of d $0.001m$.]{
	\label{fig:space}
	\includegraphics[width=0.5\linewidth]{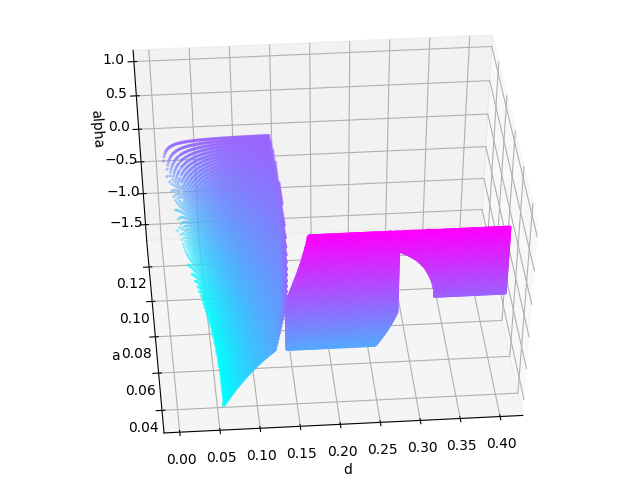}}
	\subfloat[A path with interval of d $0.01m$.]{
	\label{fig:path}
	\includegraphics[width=0.5\linewidth]{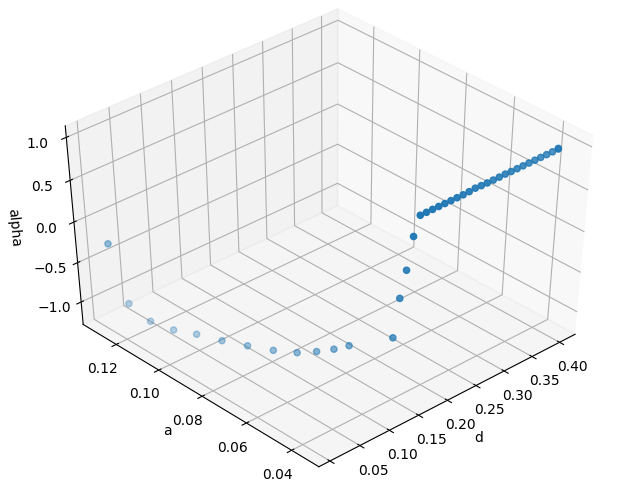}}
\caption{Space and path.}
\end{figure}

After the state check for a mesh of $d$ and $a$, we can achieve the space of possible morphology represented by triplet $(d,a,\alpha)$. We discrete $d$ with $0.001$ for demonstration and default setting for $a$ and $\alpha$. Then the generated space is as Fig. \ref{fig:space}. It demonstrates how the configuration space looks like. We can observe that it consists of one plane and one curve surface that is from far to close. The joint between two surface is a point that $S2$ can start leave the ground. And thus it is where the critical point $X8$ with $a=r$ located.

\newcommand{\wuli}{0.24}
\begin{figure}[!]
	\includegraphics[width=\wuli\linewidth]{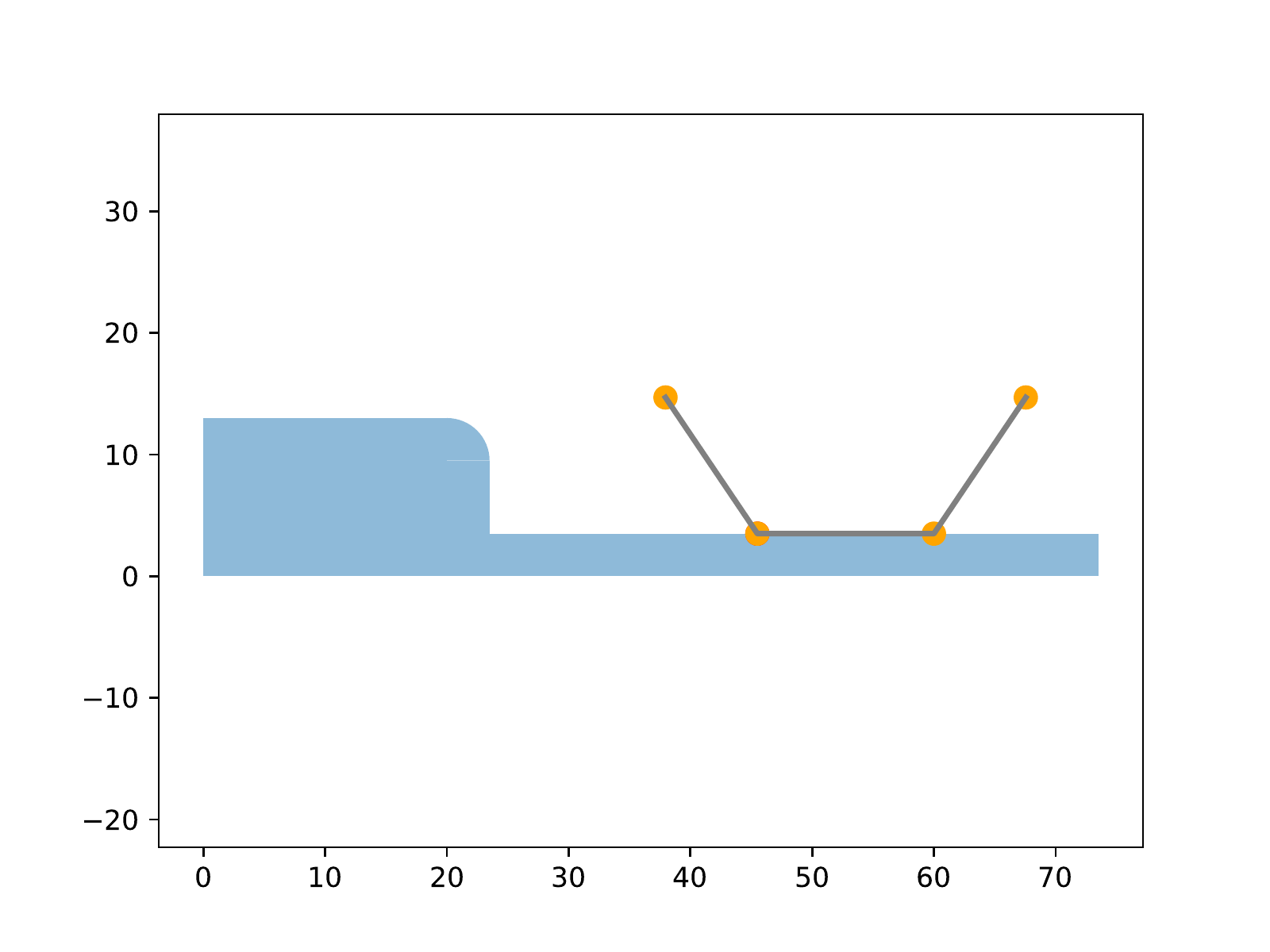}
	\includegraphics[width=\wuli\linewidth]{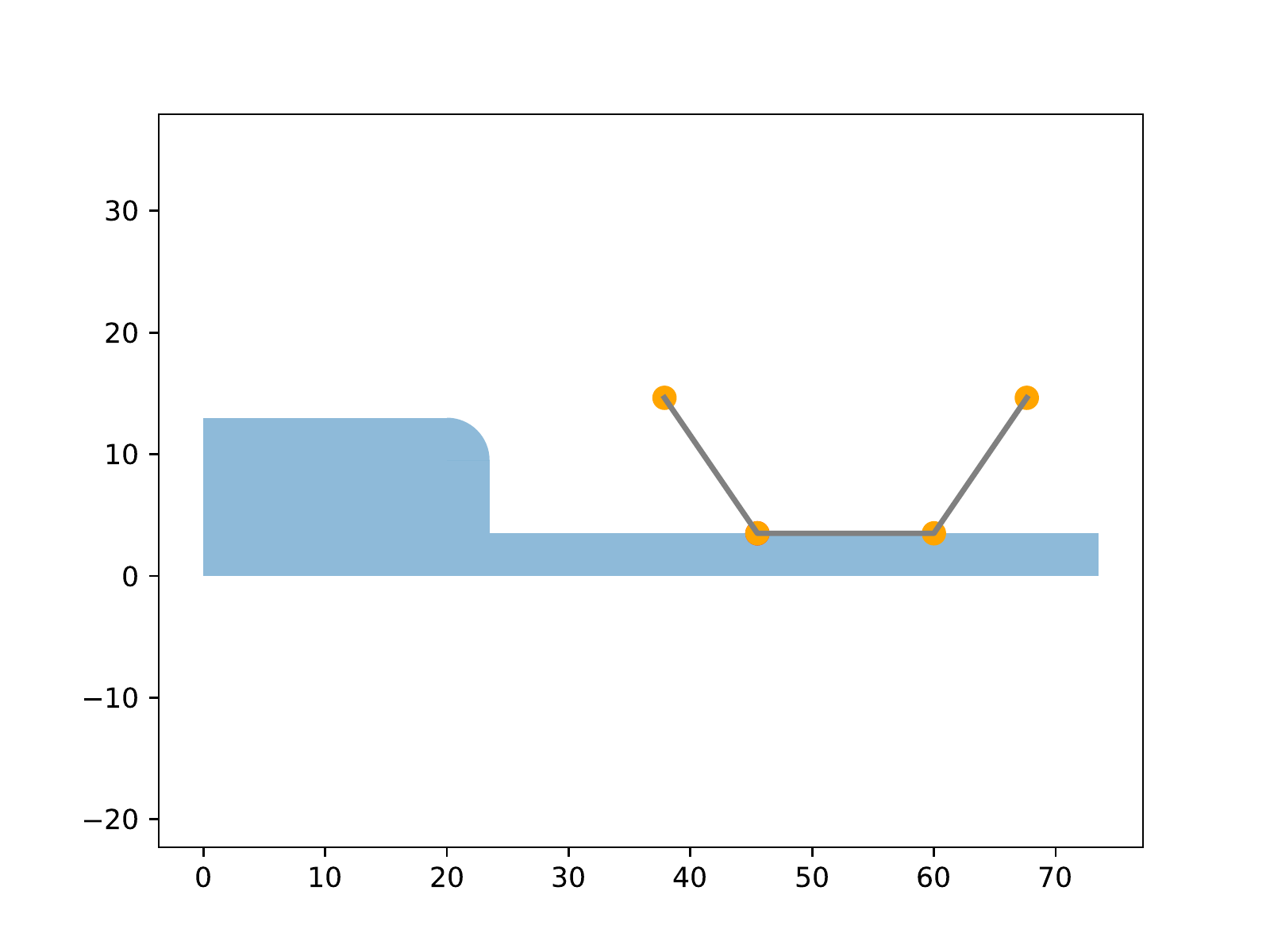}
	\includegraphics[width=\wuli\linewidth]{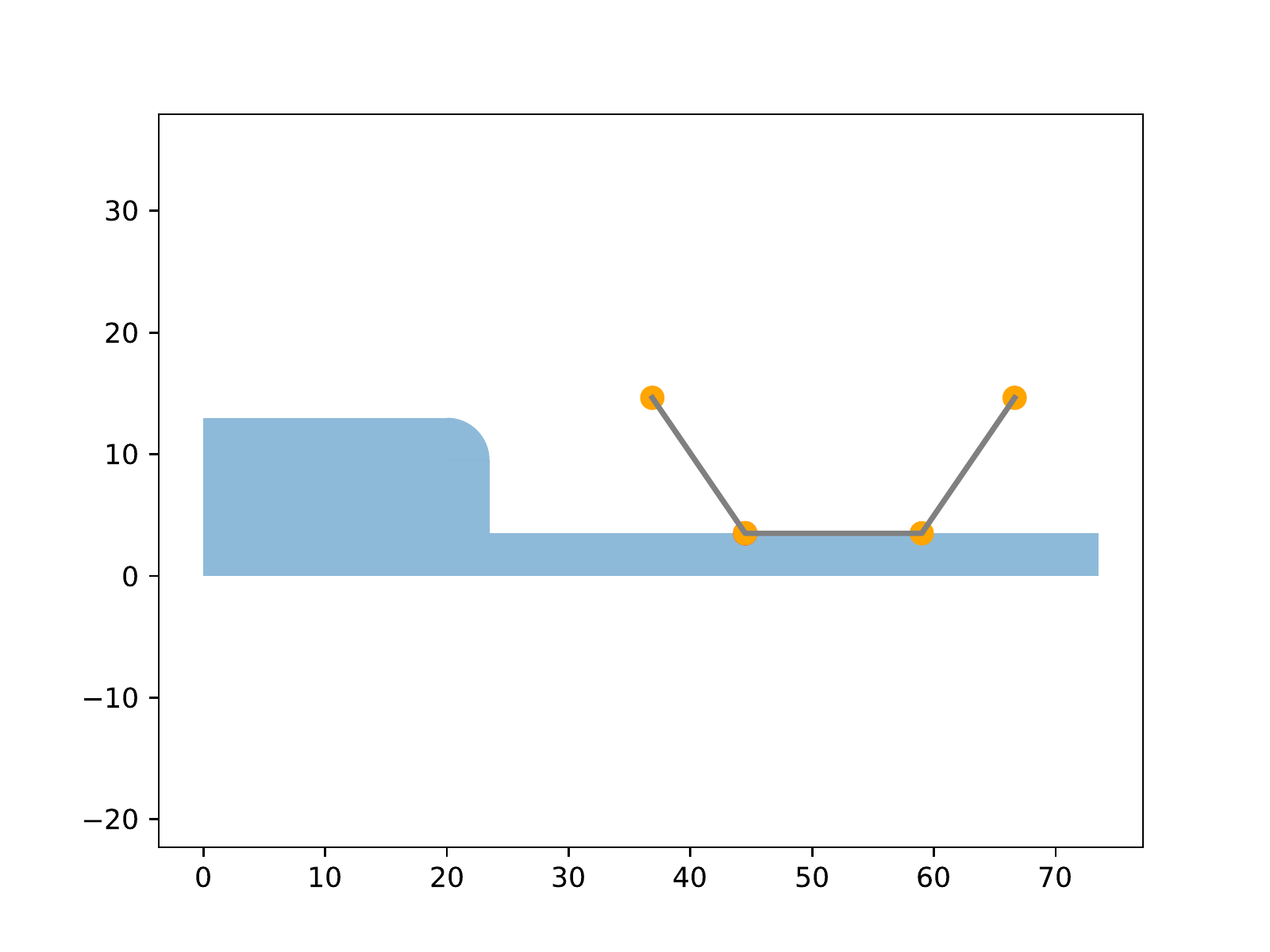}
	\includegraphics[width=\wuli\linewidth]{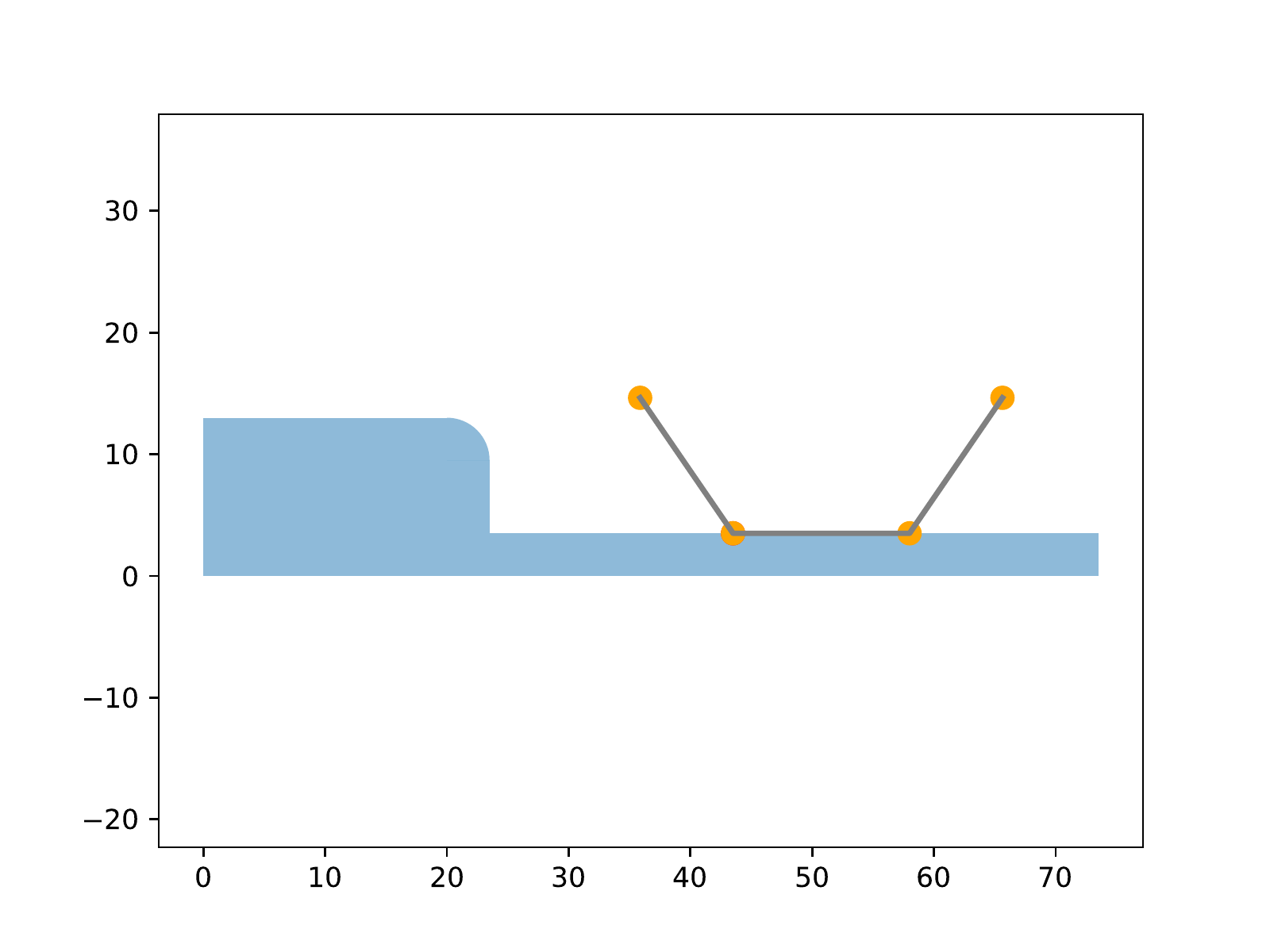}\\
	\includegraphics[width=\wuli\linewidth]{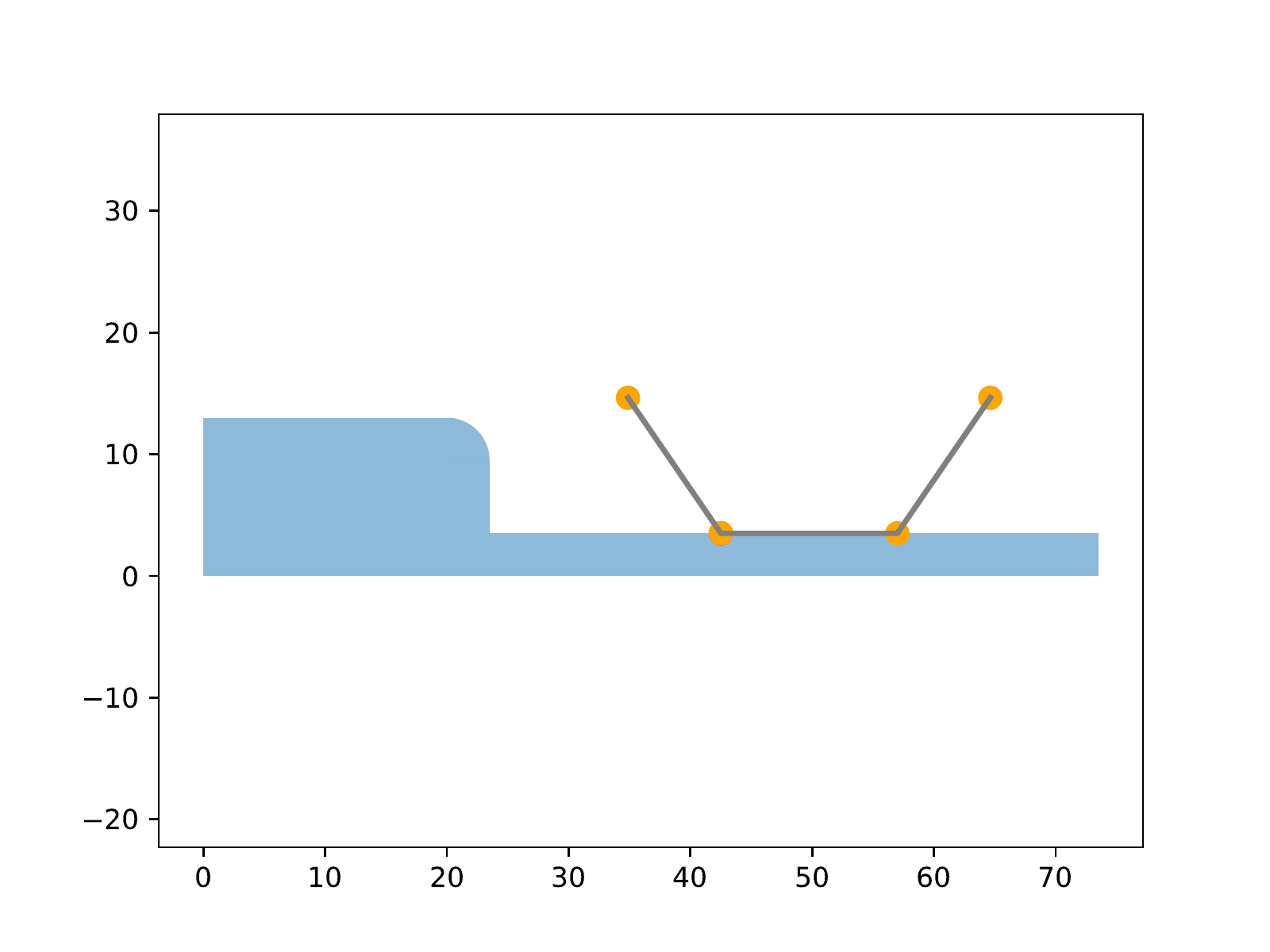}
	\includegraphics[width=\wuli\linewidth]{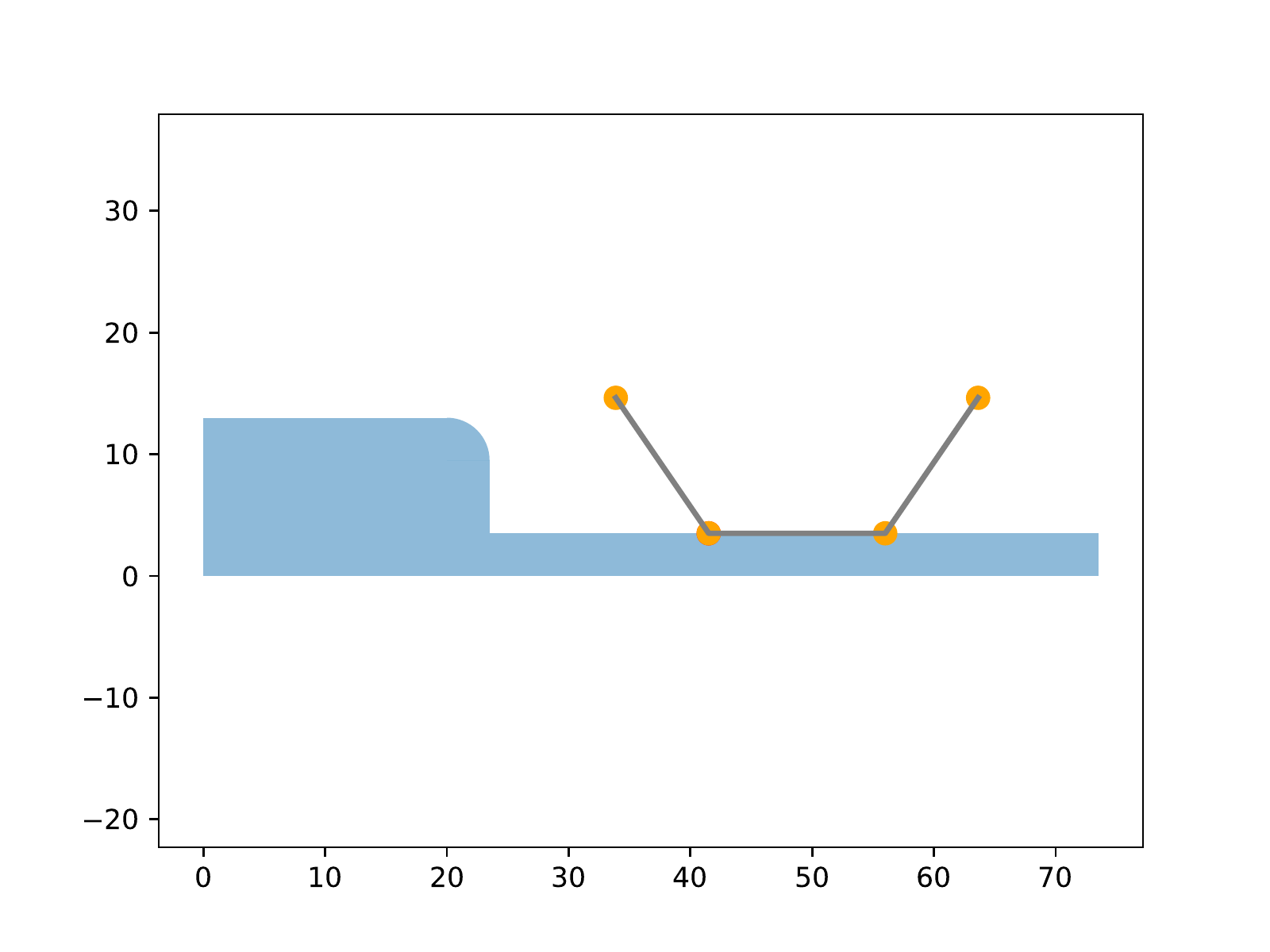}
	\includegraphics[width=\wuli\linewidth]{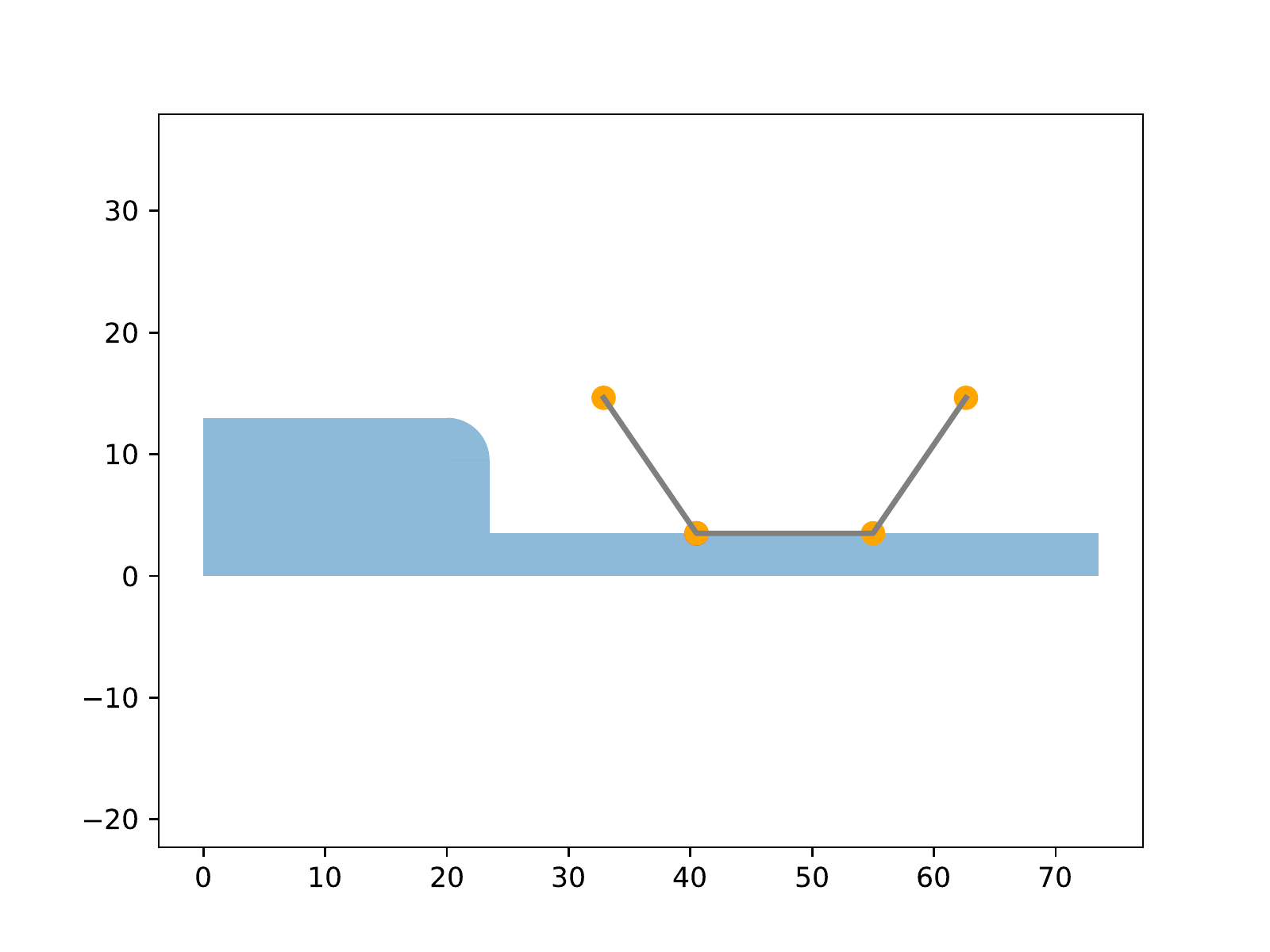}
	\includegraphics[width=\wuli\linewidth]{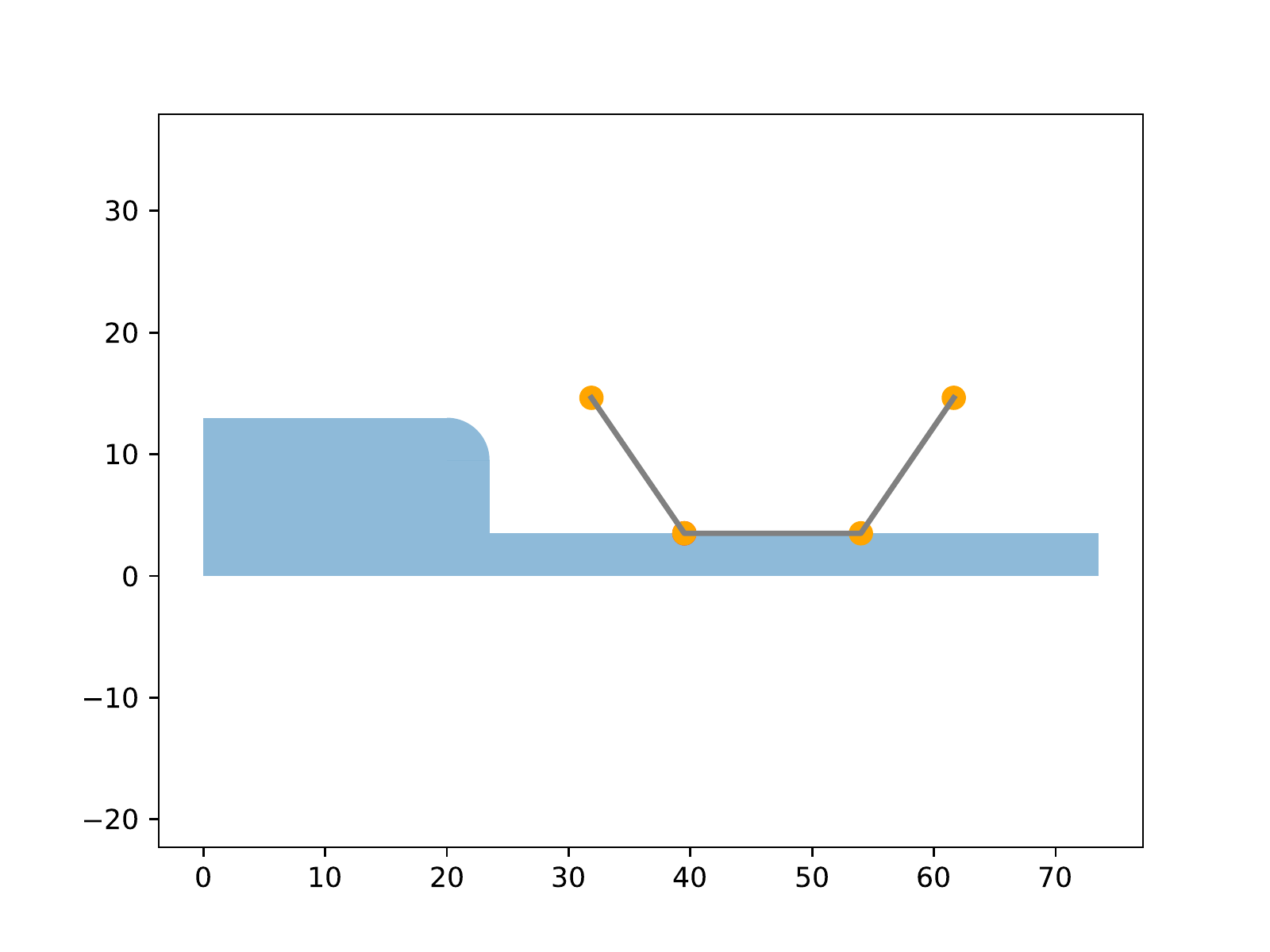}\\
	\includegraphics[width=\wuli\linewidth]{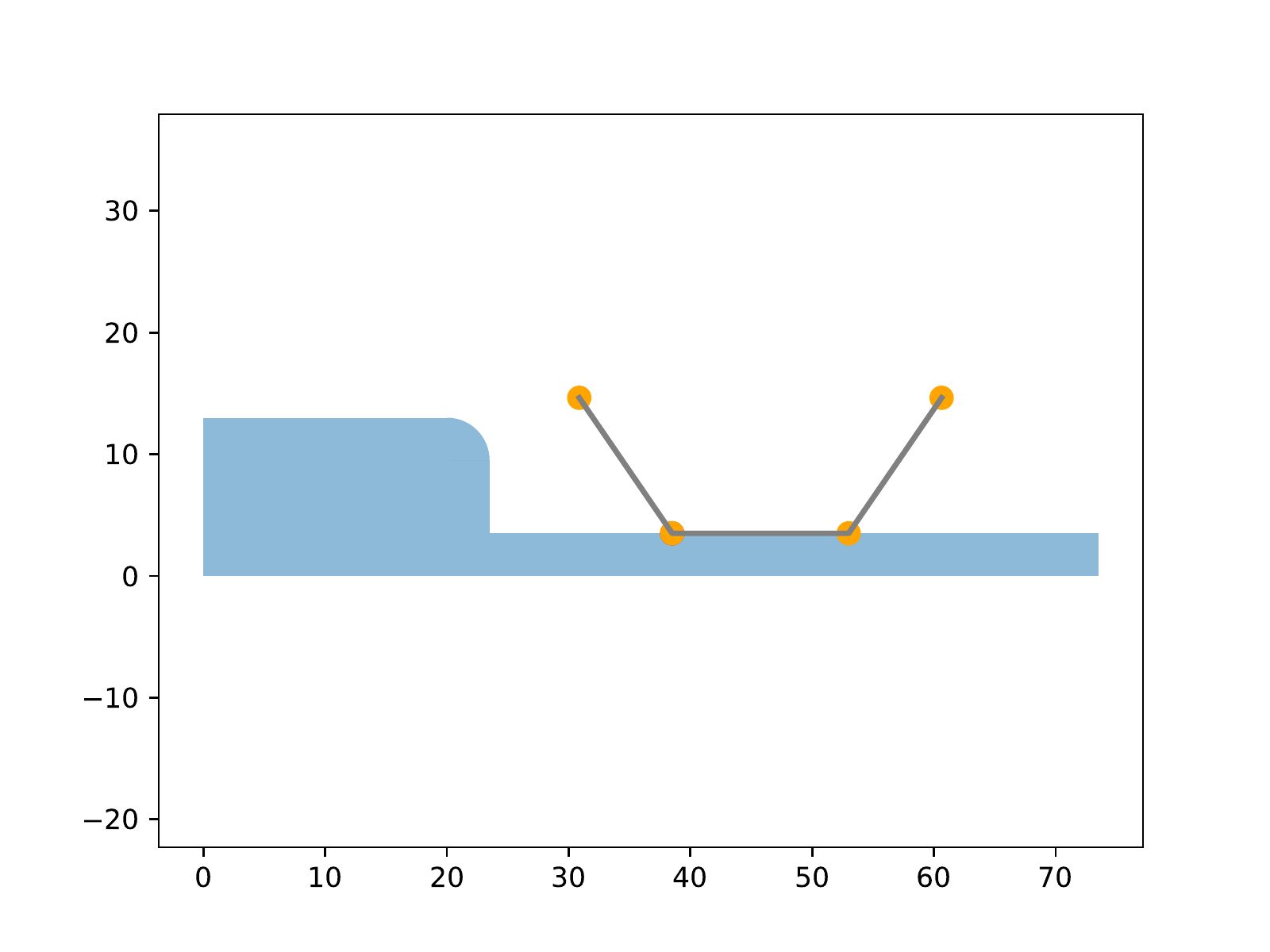}
	\includegraphics[width=\wuli\linewidth]{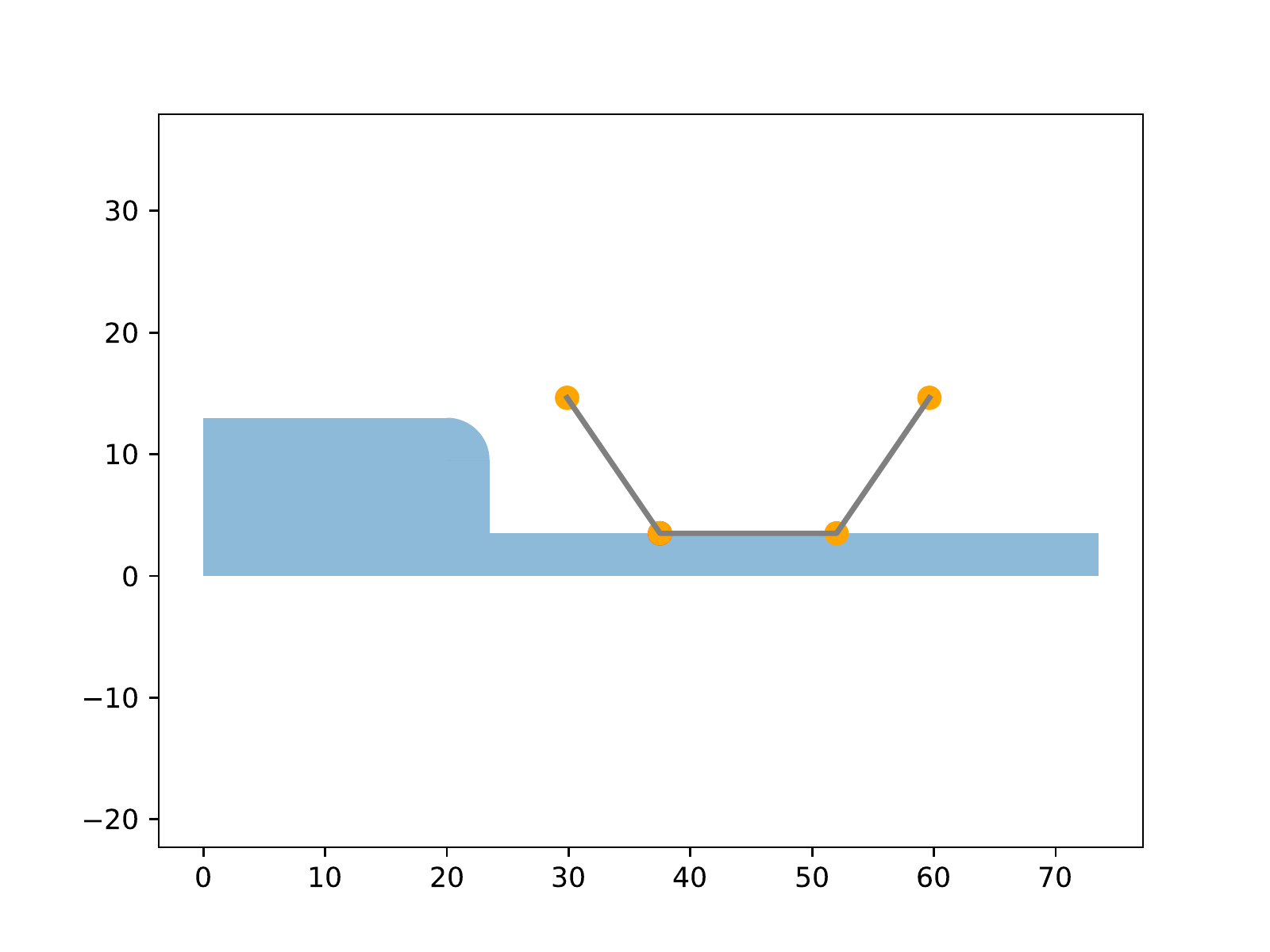}
	\includegraphics[width=\wuli\linewidth]{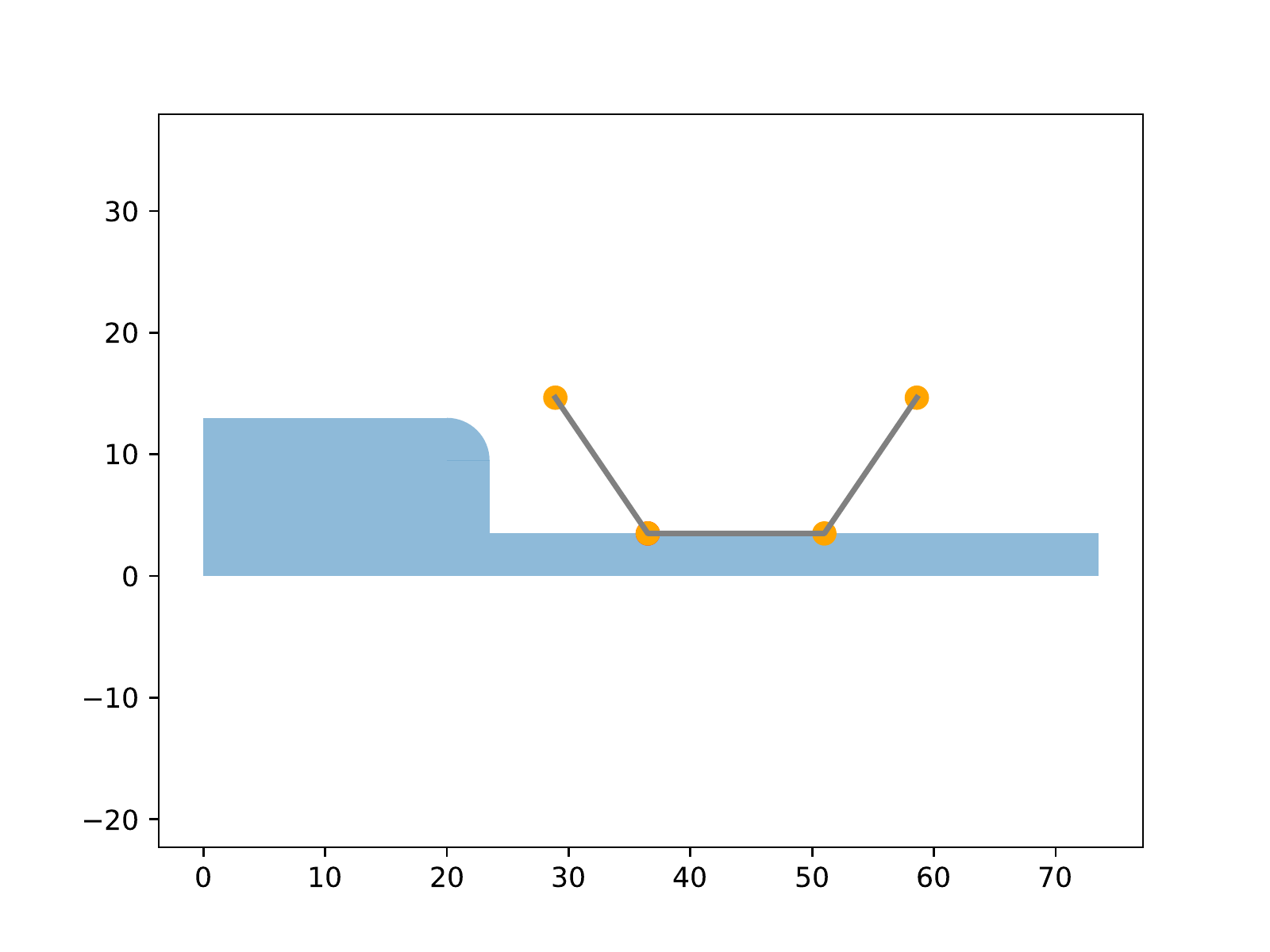}
	\includegraphics[width=\wuli\linewidth]{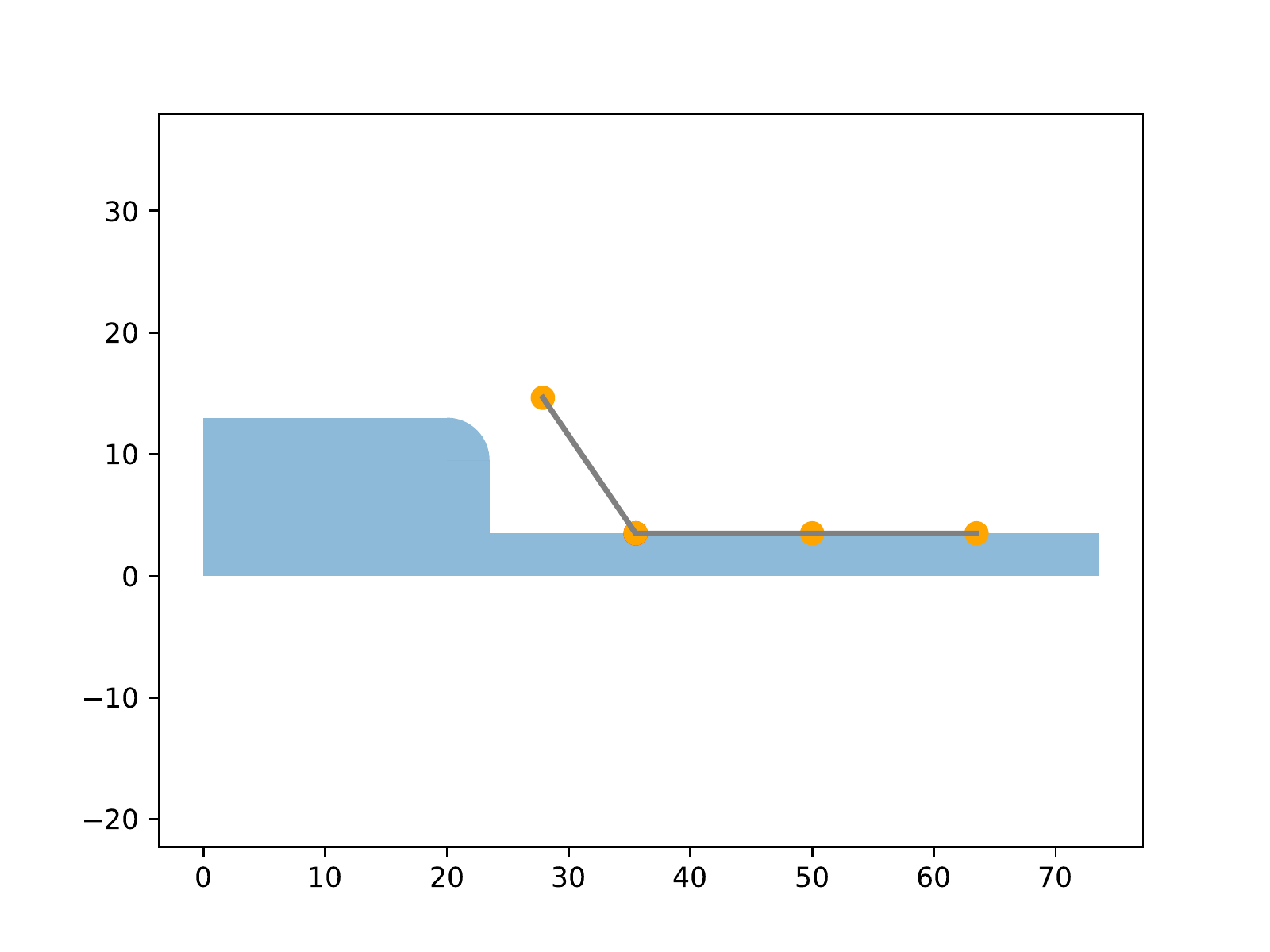}\\
	\includegraphics[width=\wuli\linewidth]{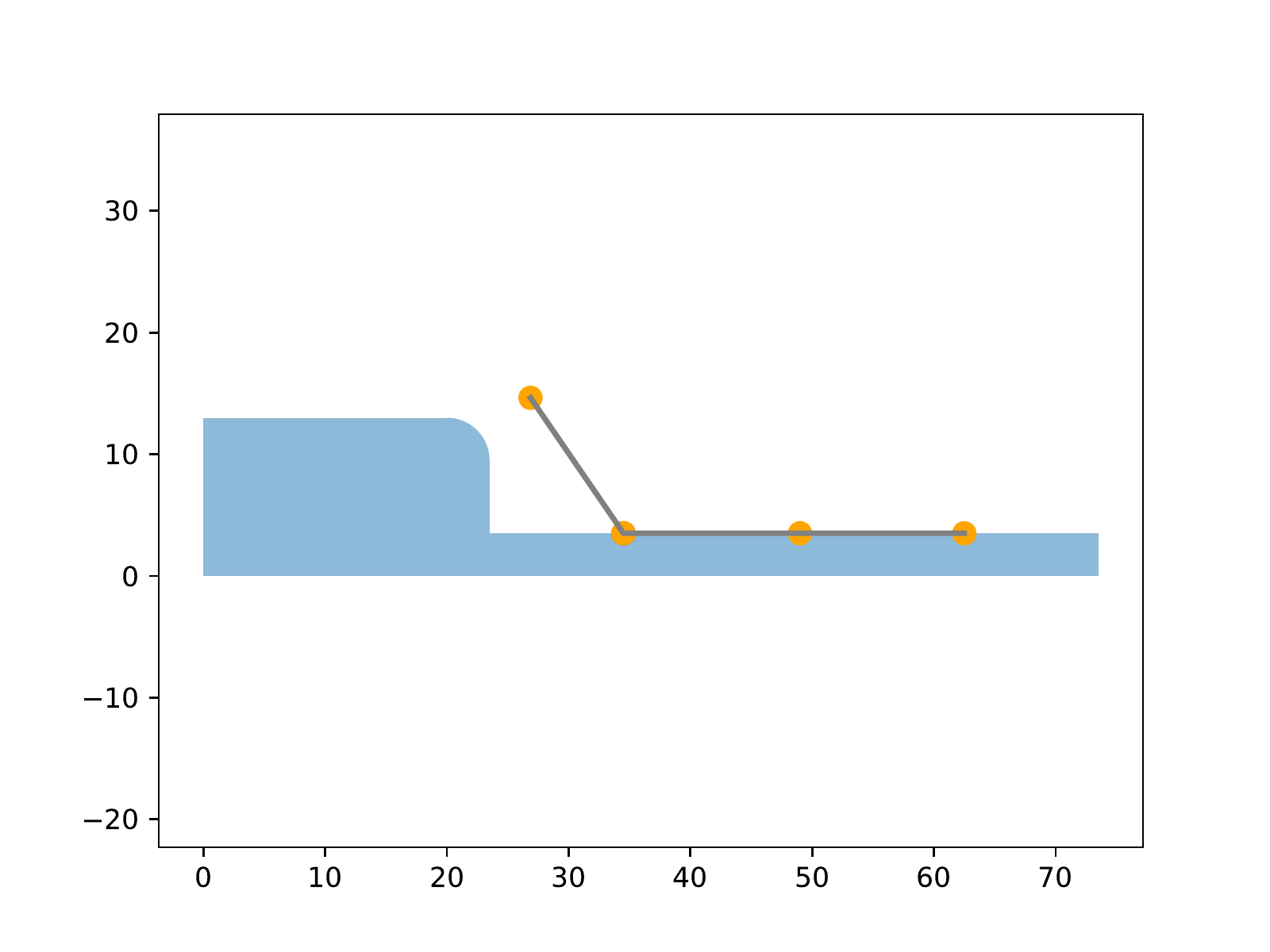}
	\includegraphics[width=\wuli\linewidth]{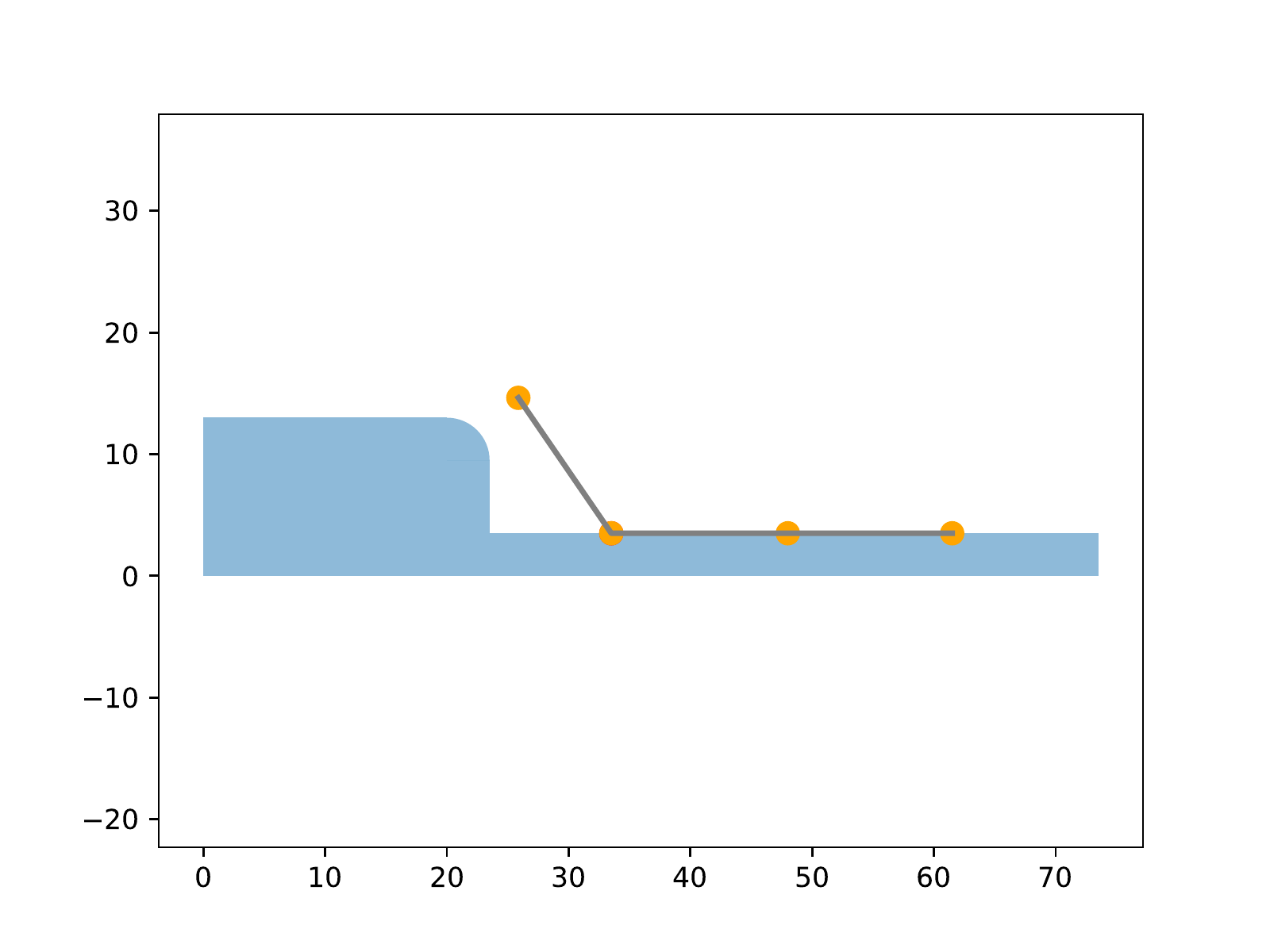}
	\includegraphics[width=\wuli\linewidth]{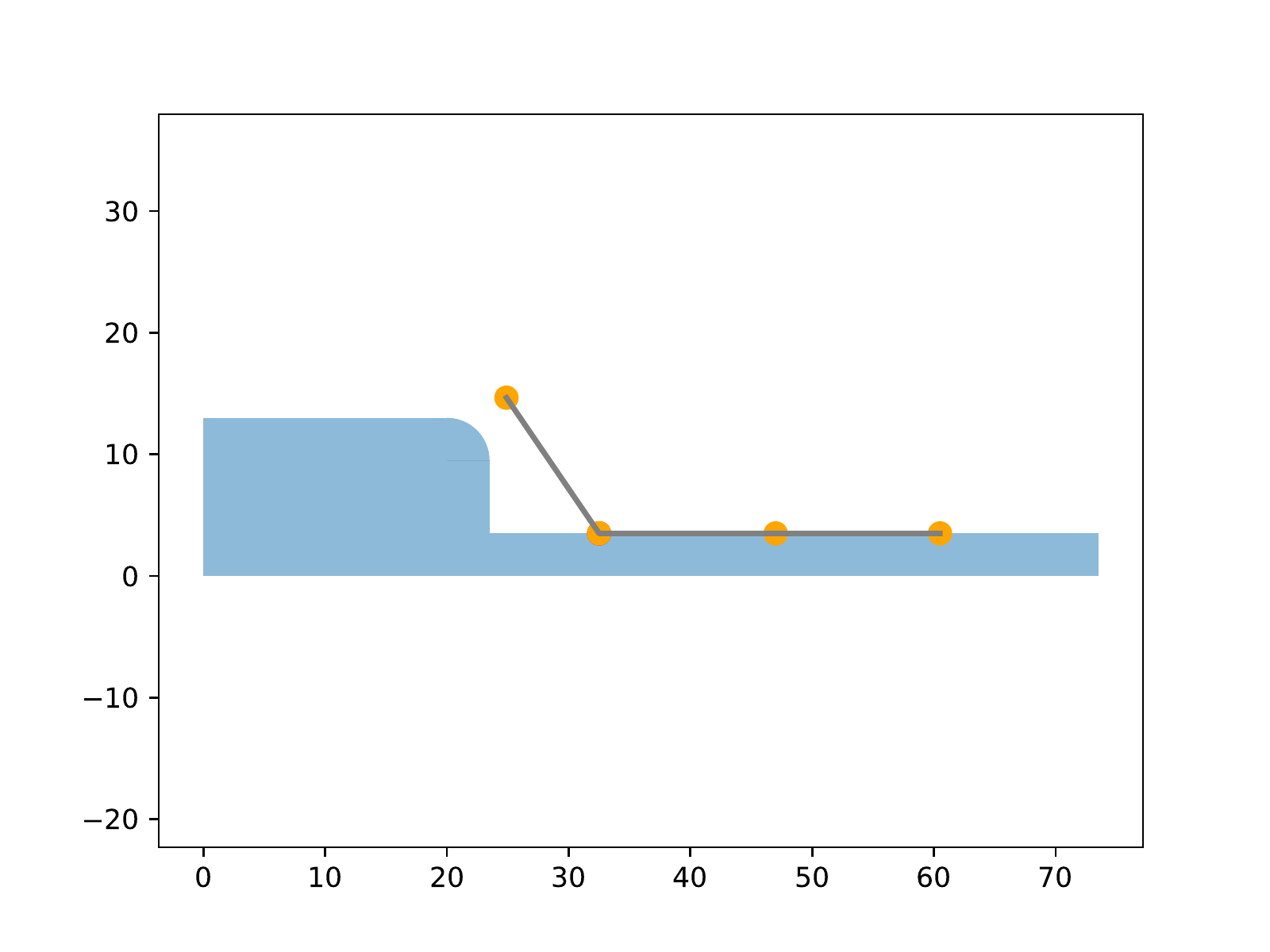}
	\includegraphics[width=\wuli\linewidth]{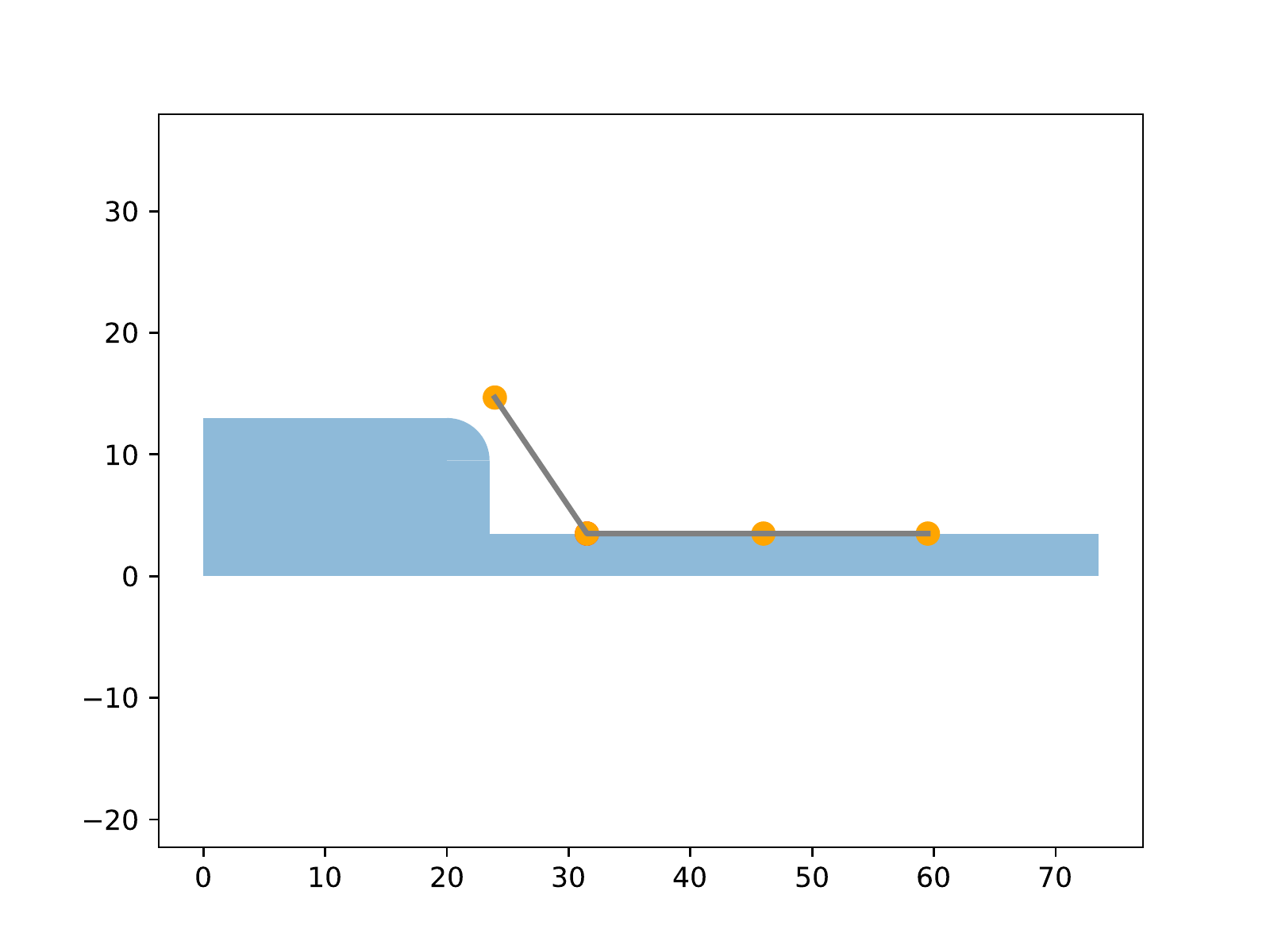}\\
	\includegraphics[width=\wuli\linewidth]{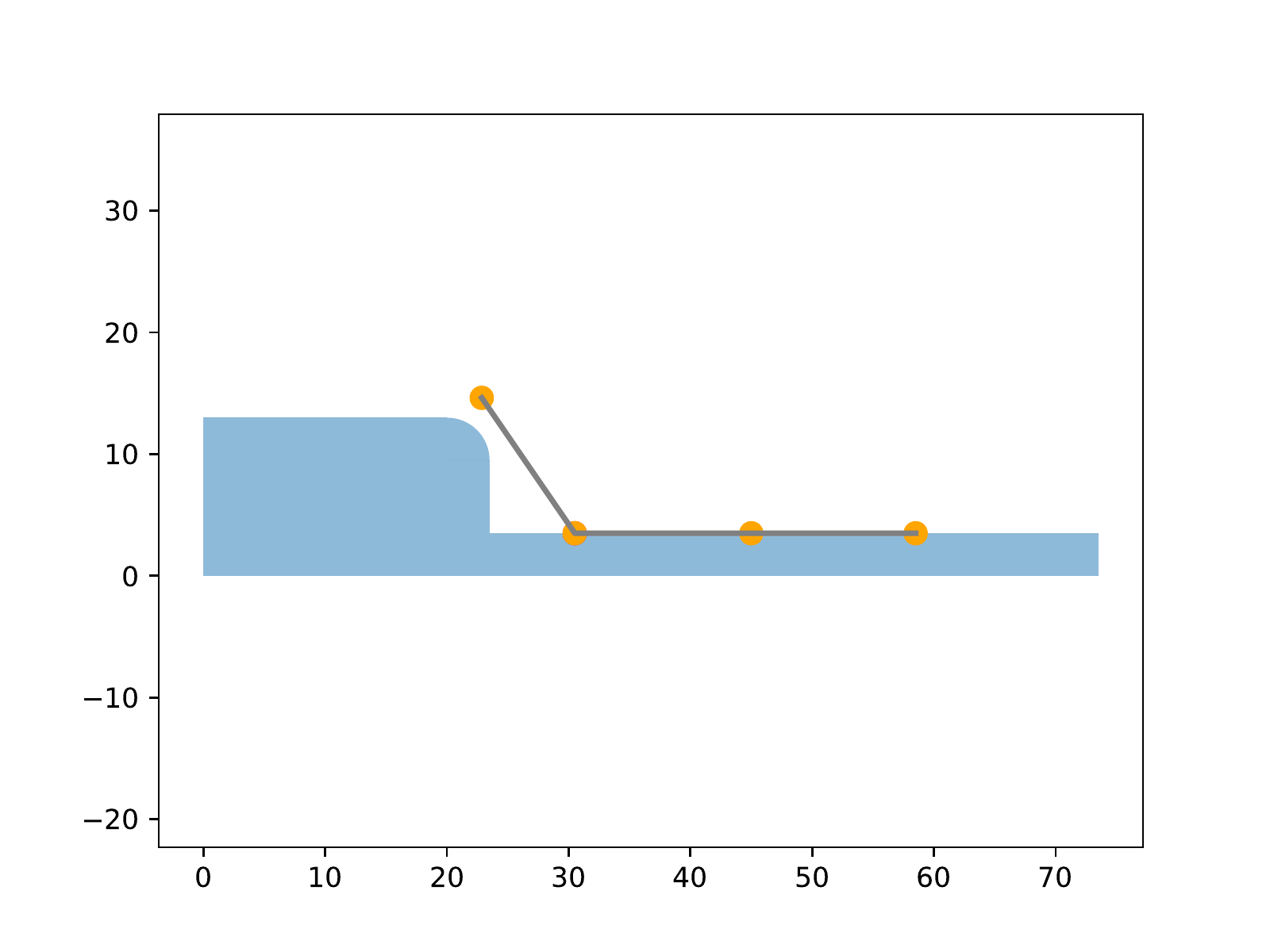}
	\includegraphics[width=\wuli\linewidth]{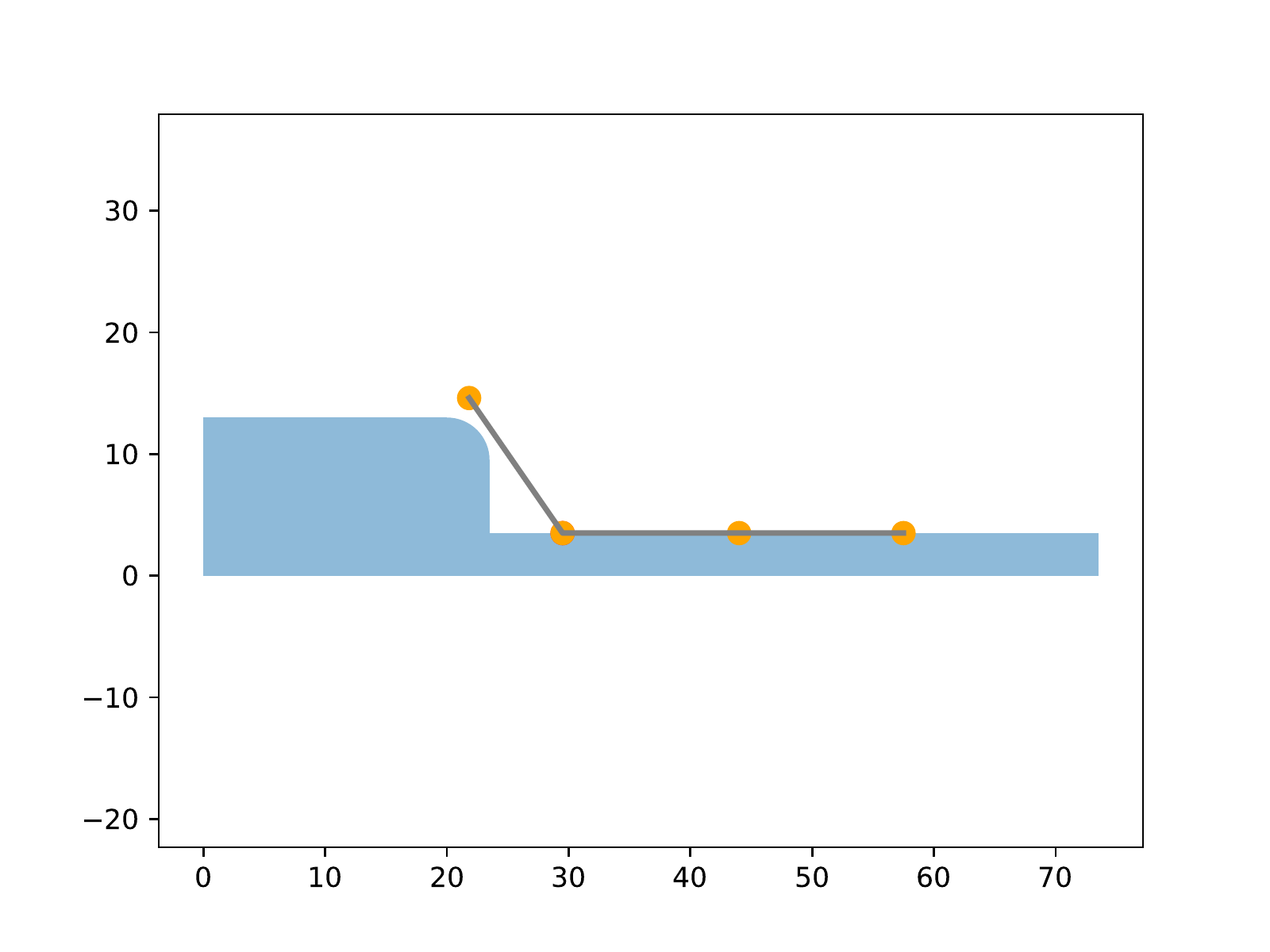}
	\includegraphics[width=\wuli\linewidth]{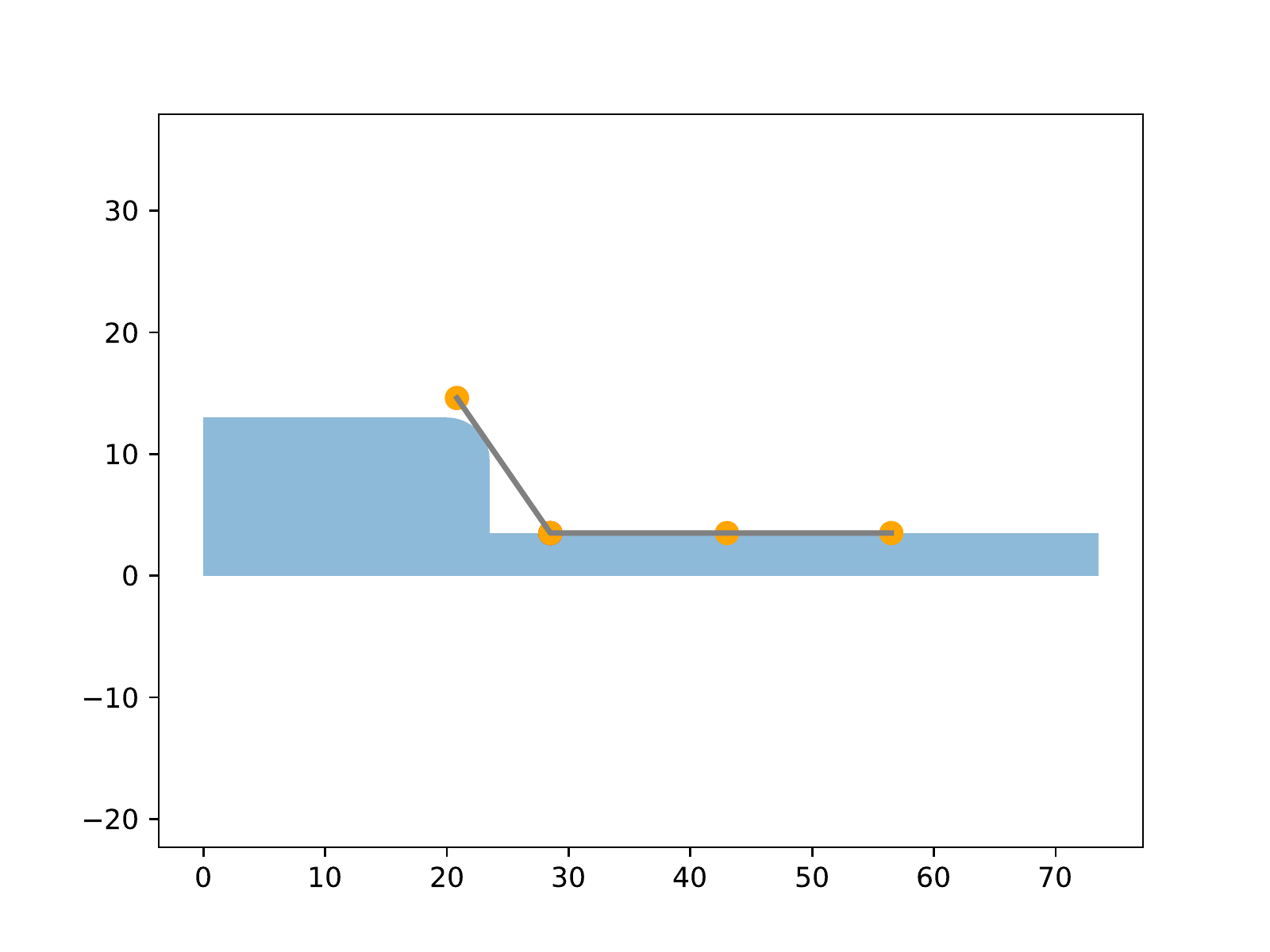}
	\includegraphics[width=\wuli\linewidth]{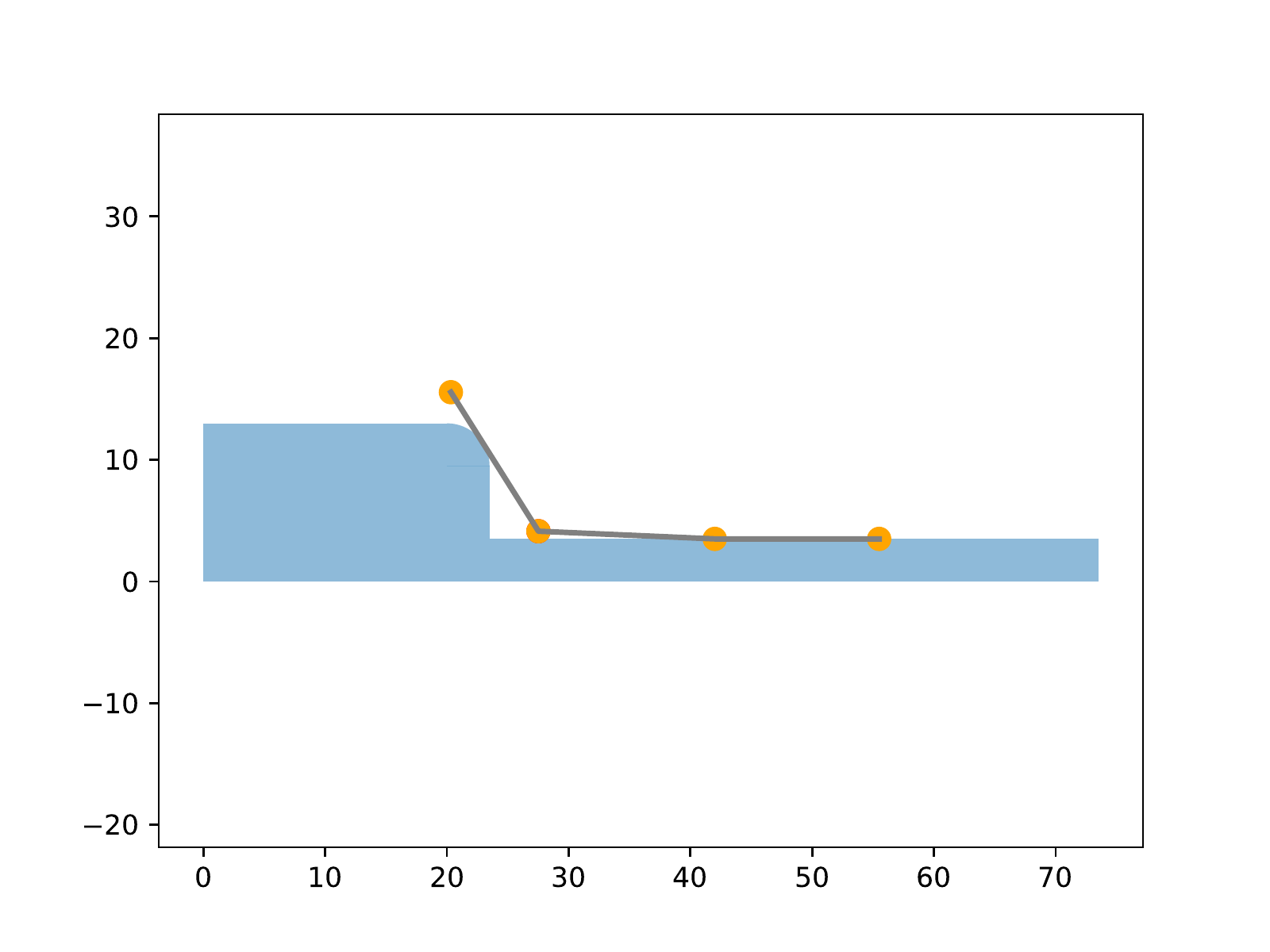}\\
	\includegraphics[width=\wuli\linewidth]{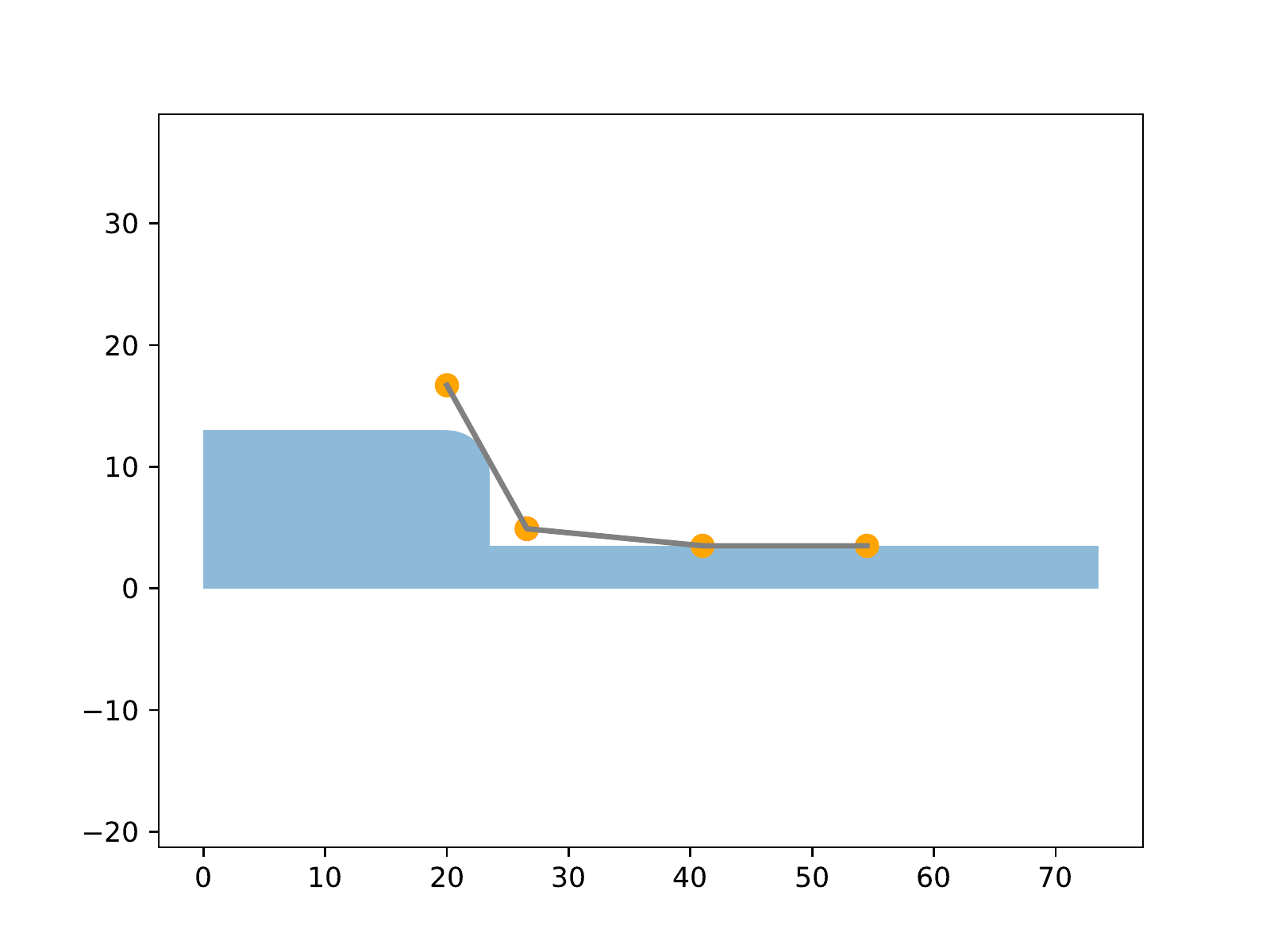}
	\includegraphics[width=\wuli\linewidth]{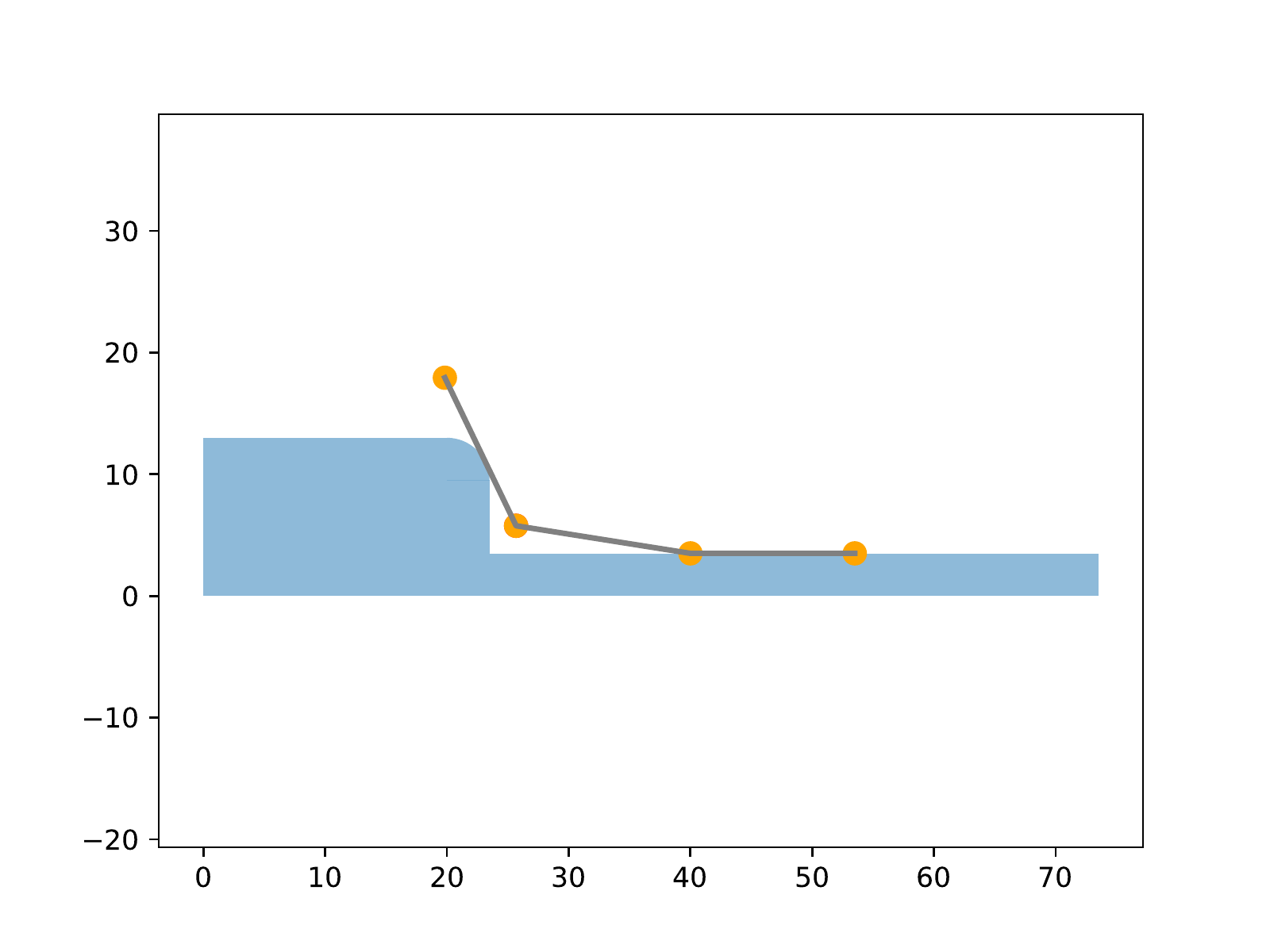}
	\includegraphics[width=\wuli\linewidth]{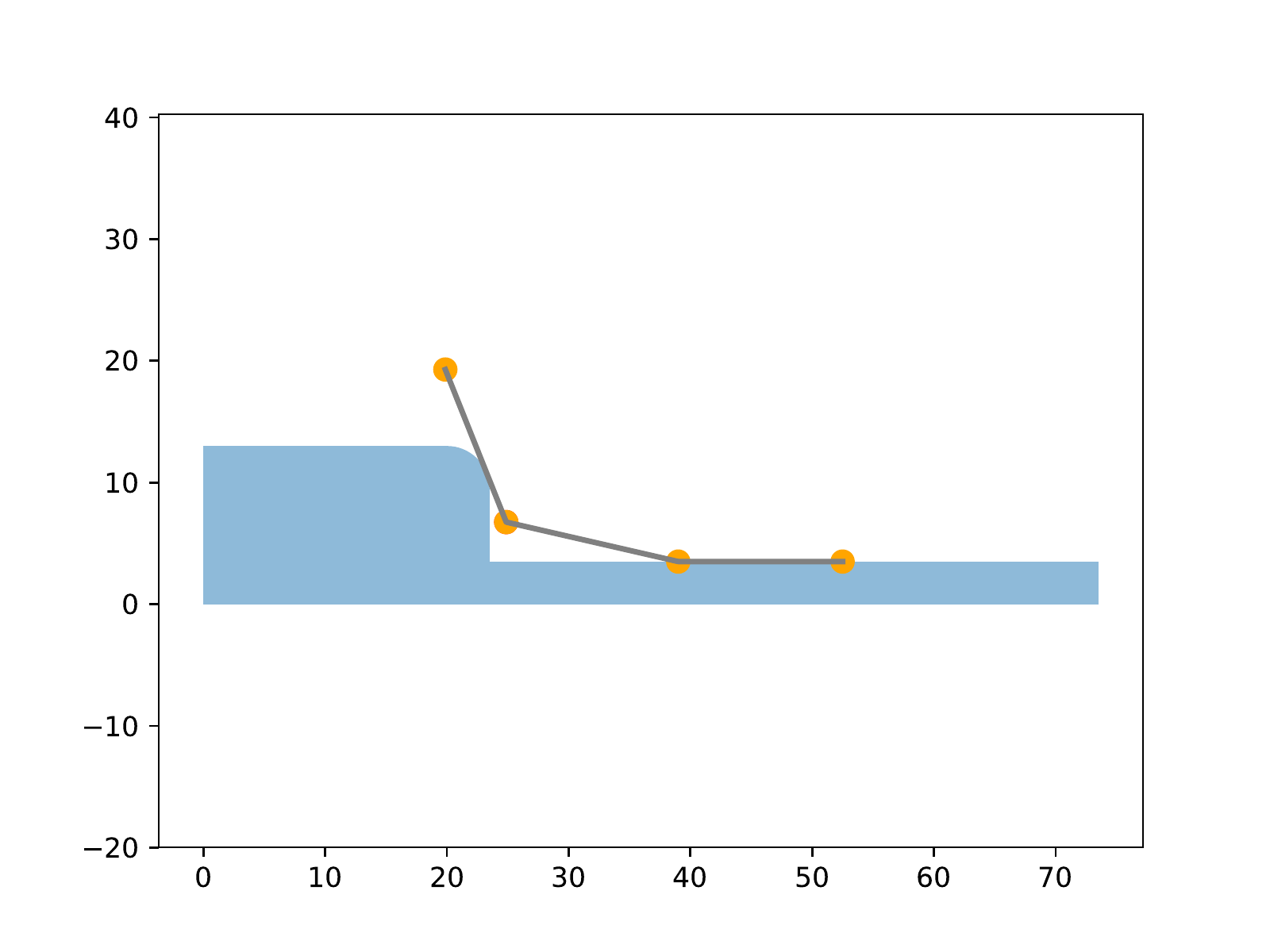}
	\includegraphics[width=\wuli\linewidth]{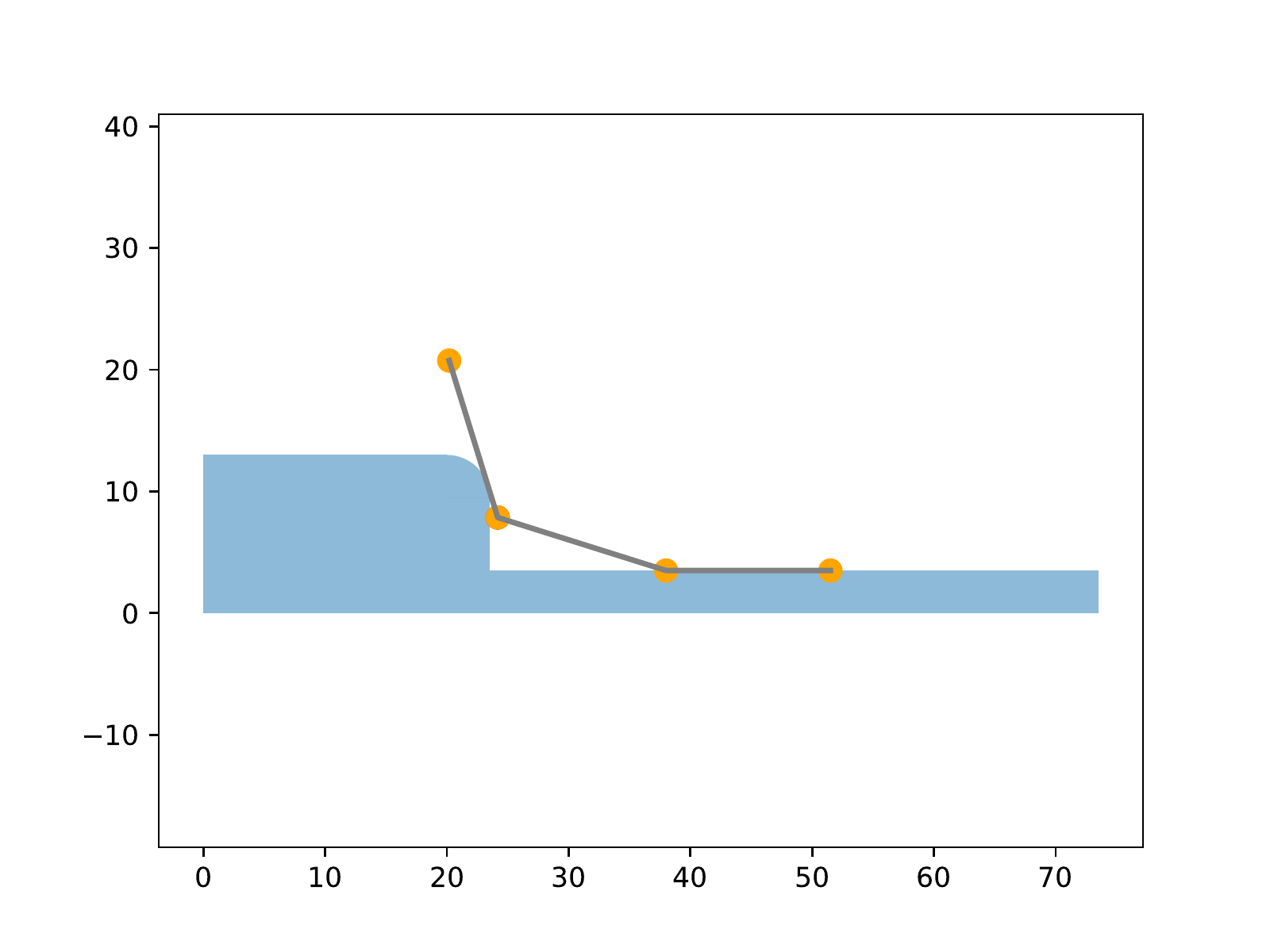}\\
	\includegraphics[width=\wuli\linewidth]{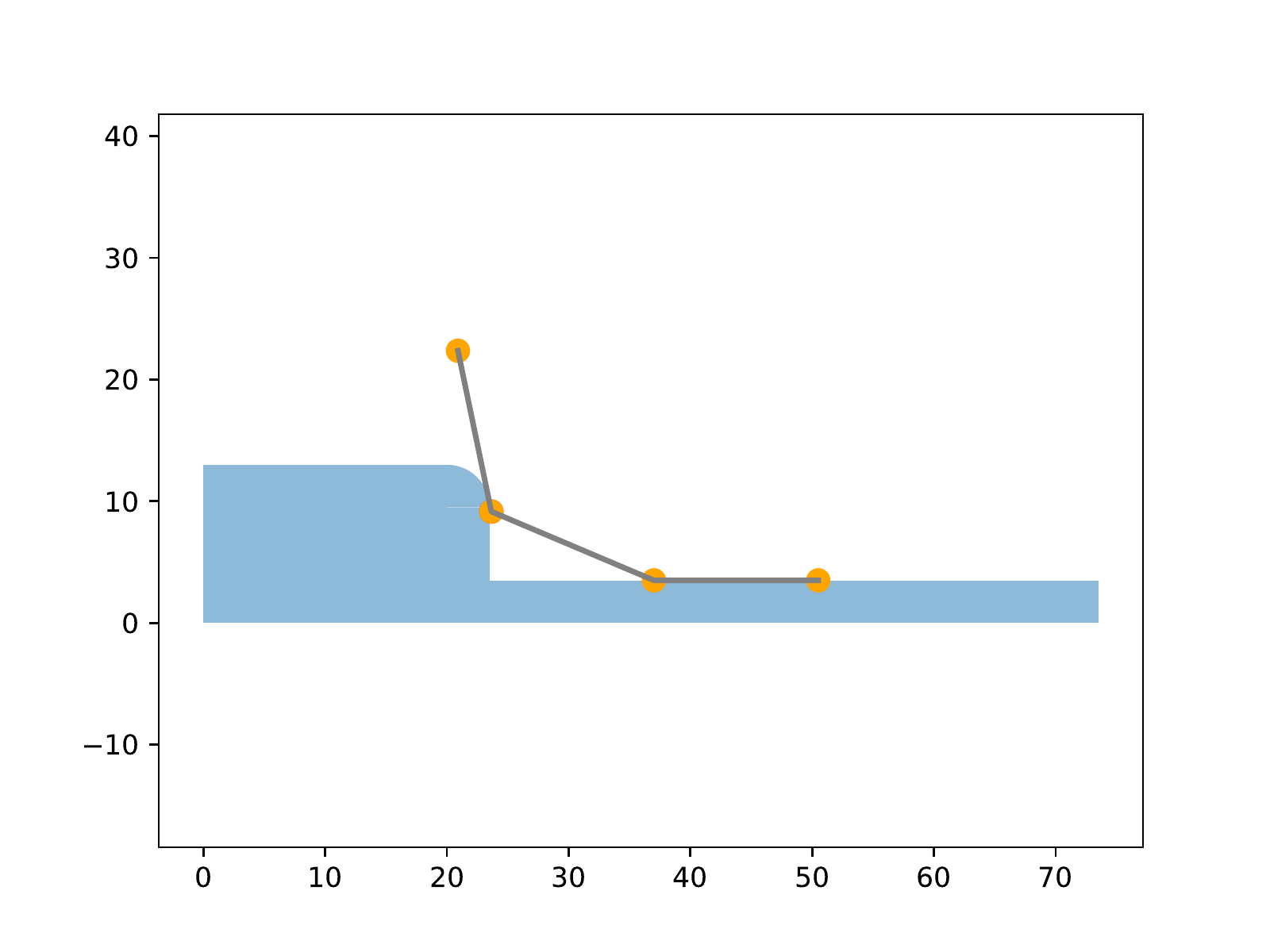}
	\includegraphics[width=\wuli\linewidth]{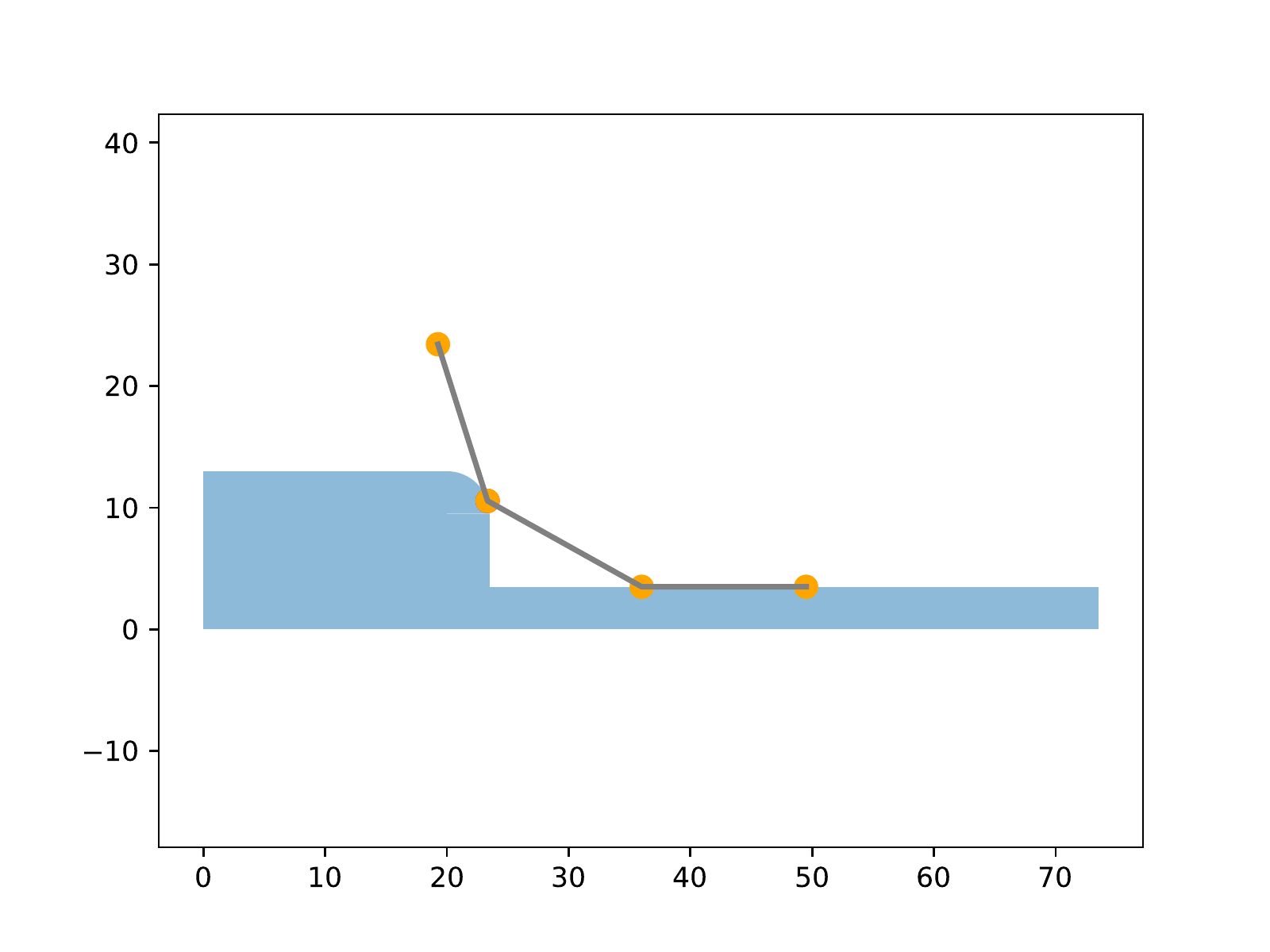}
	\includegraphics[width=\wuli\linewidth]{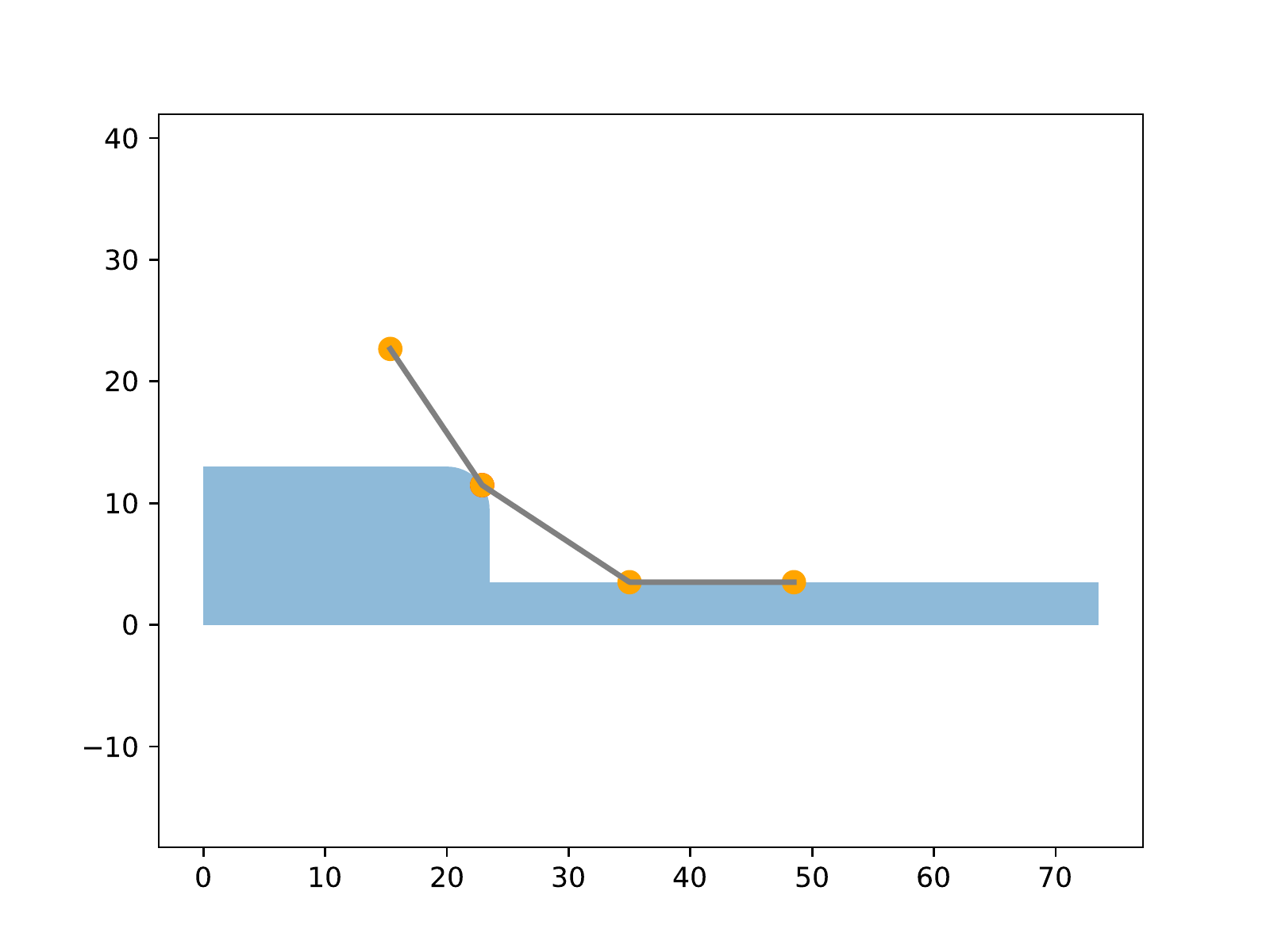}
	\includegraphics[width=\wuli\linewidth]{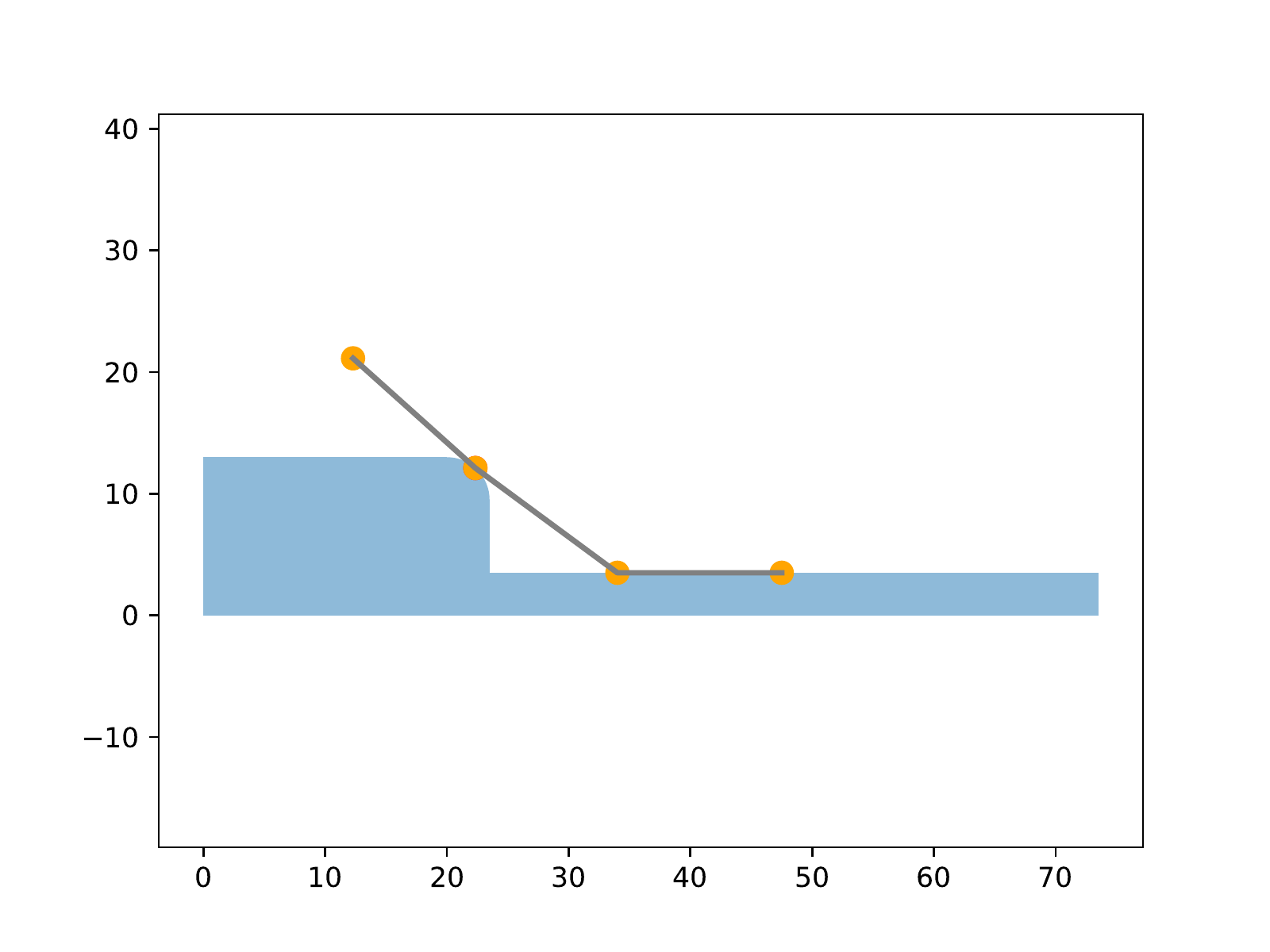}\\
	\includegraphics[width=\wuli\linewidth]{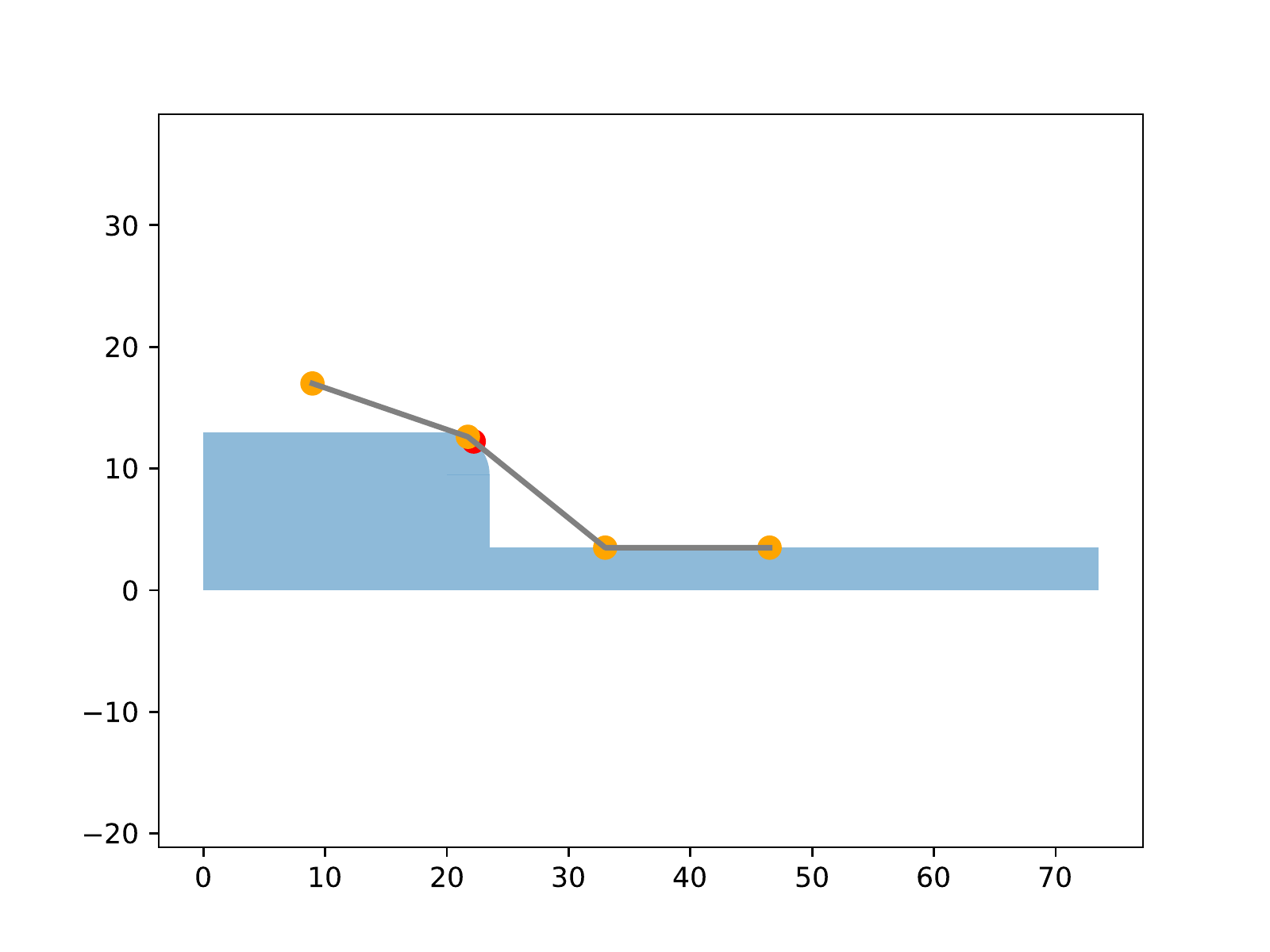}
	\includegraphics[width=\wuli\linewidth]{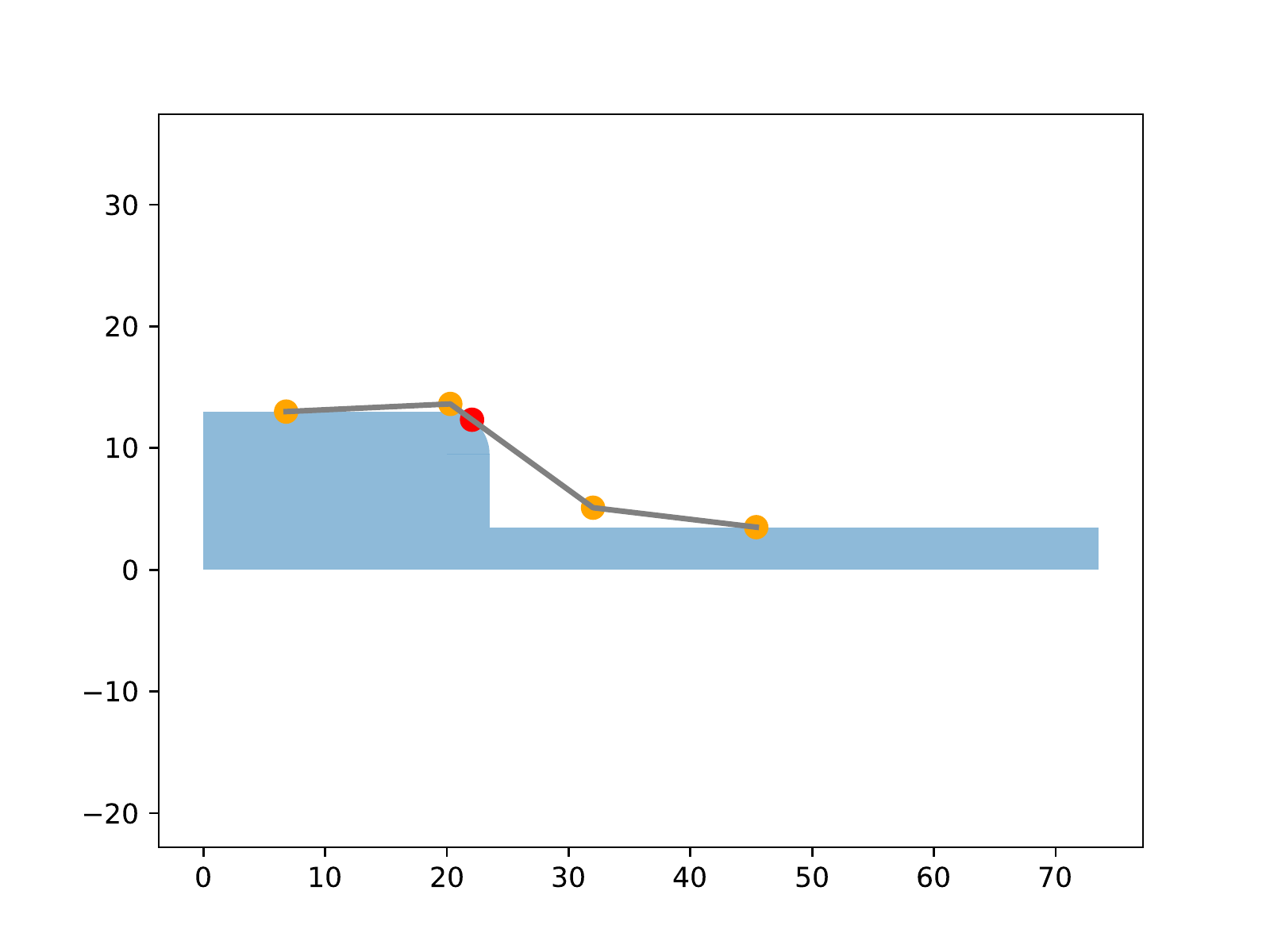}
	\includegraphics[width=\wuli\linewidth]{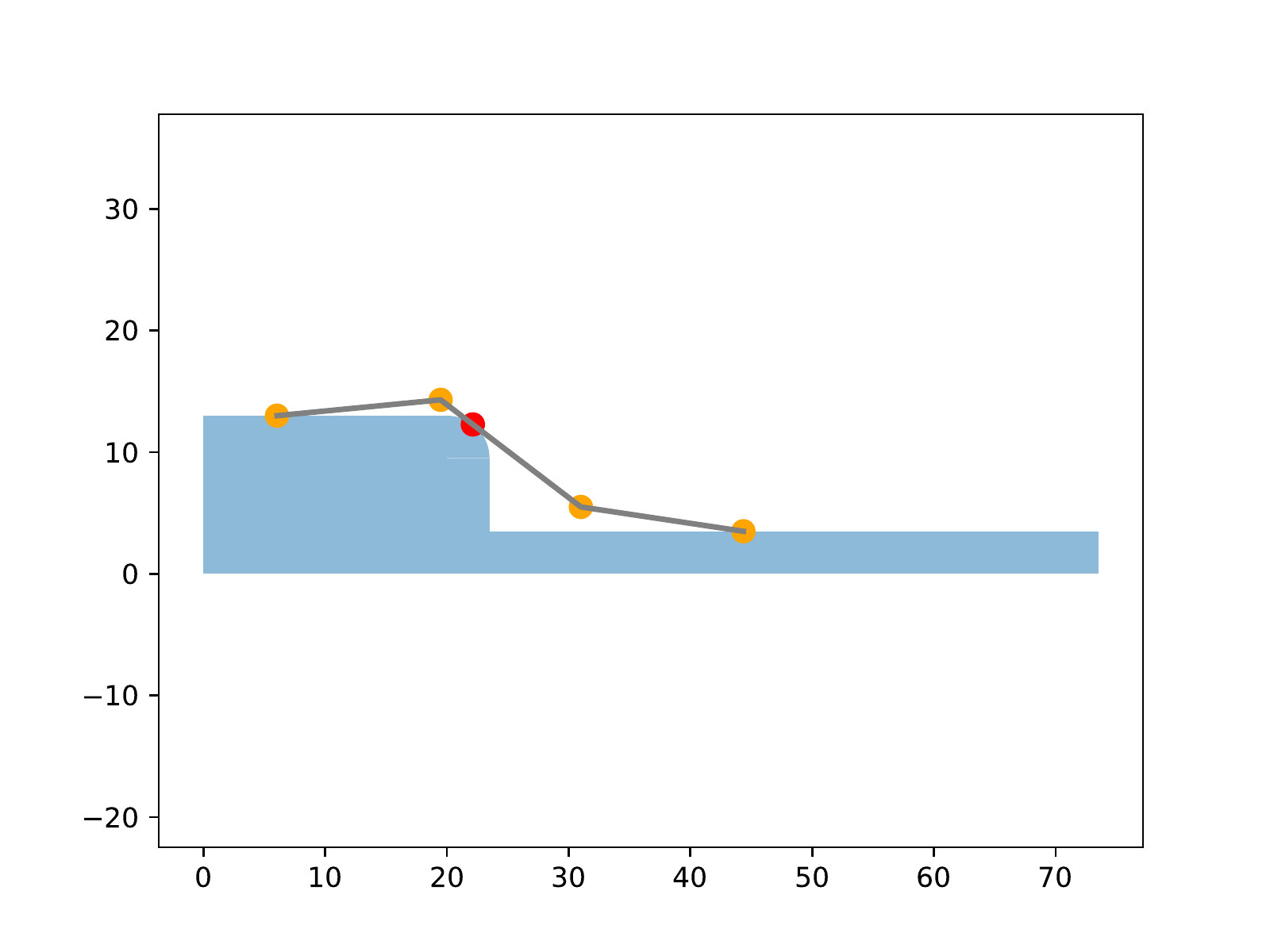}
	\includegraphics[width=\wuli\linewidth]{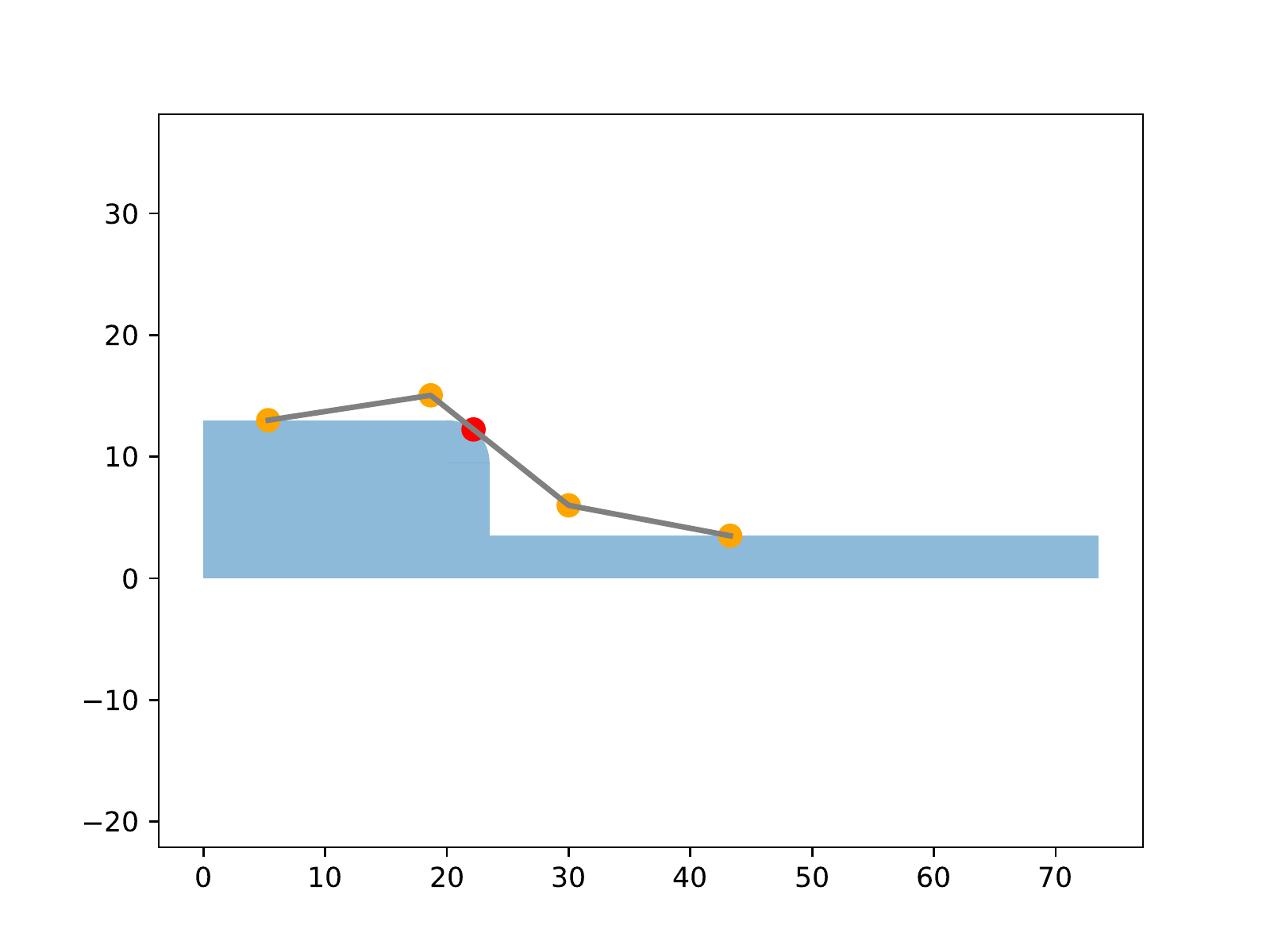}\\
	\includegraphics[width=\wuli\linewidth]{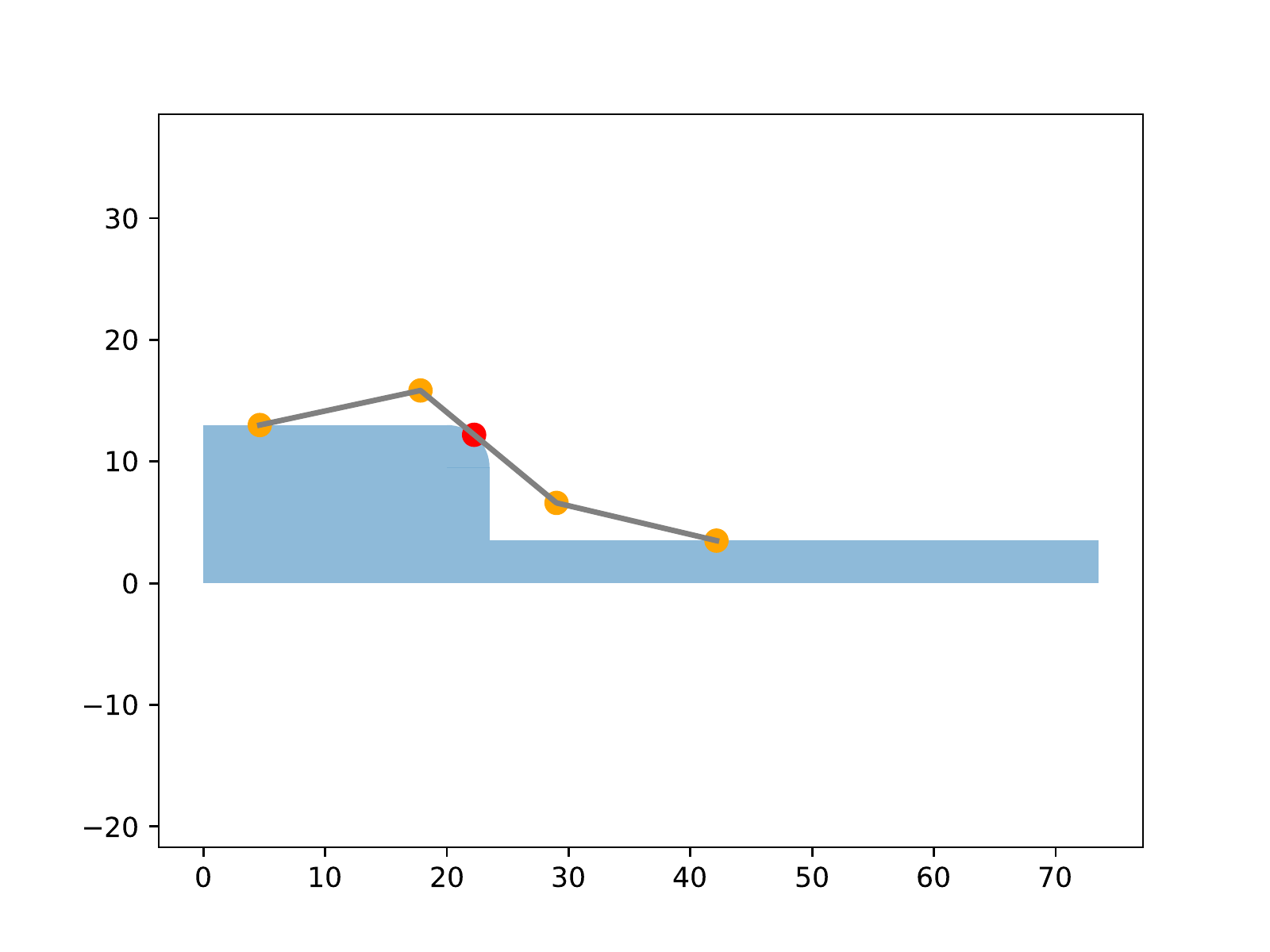}
	\includegraphics[width=\wuli\linewidth]{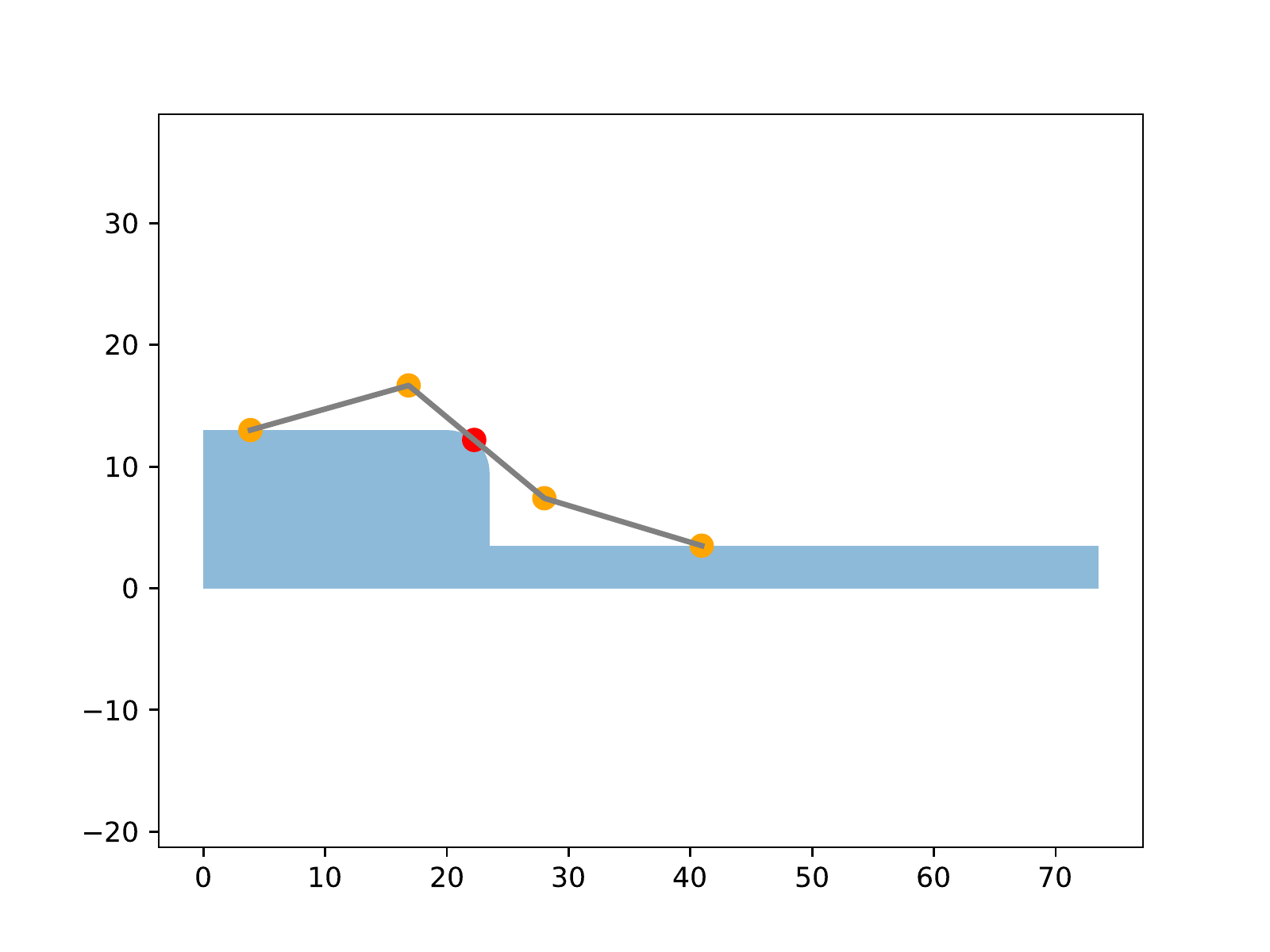}
	\includegraphics[width=\wuli\linewidth]{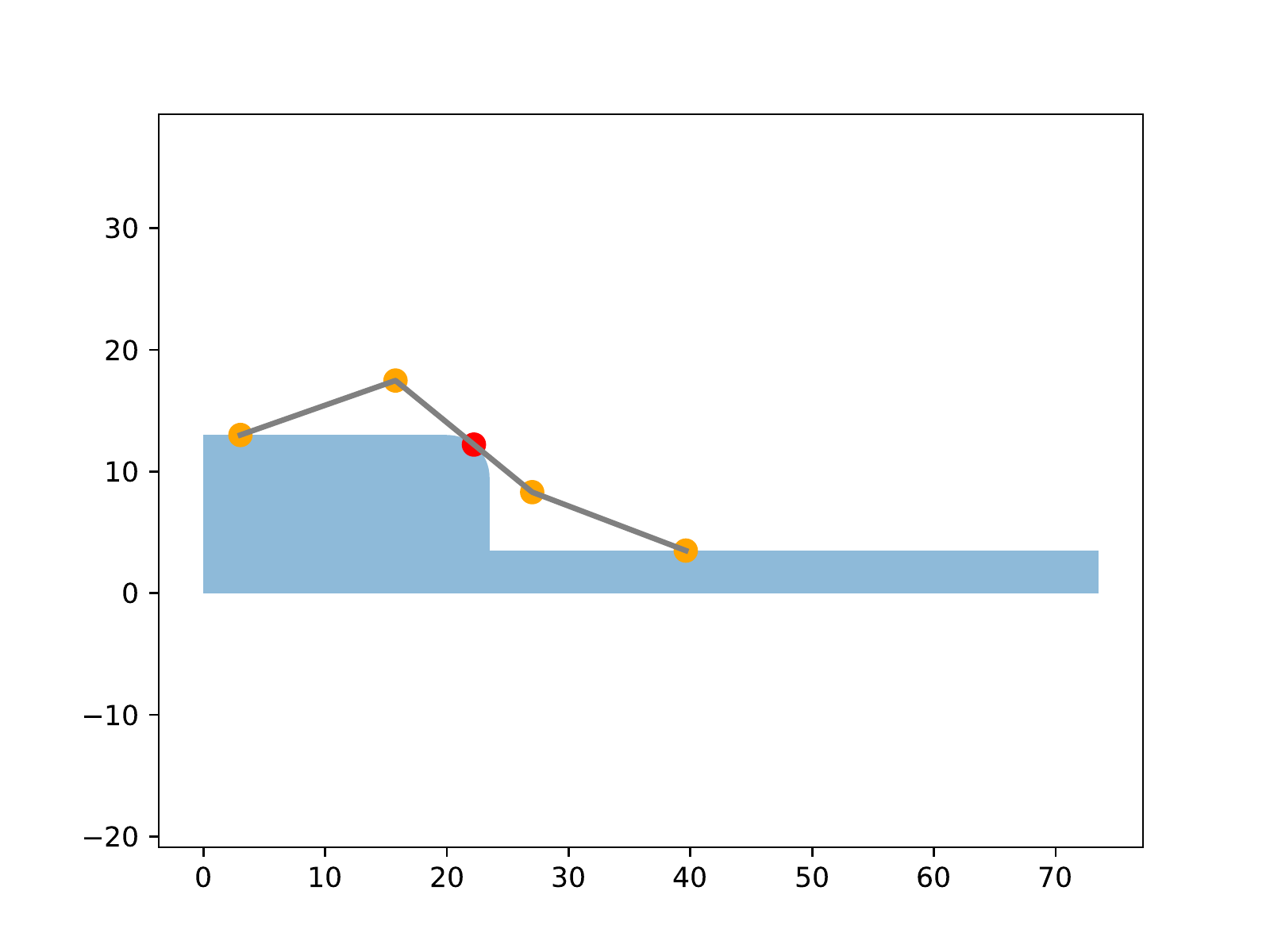}
	\includegraphics[width=\wuli\linewidth]{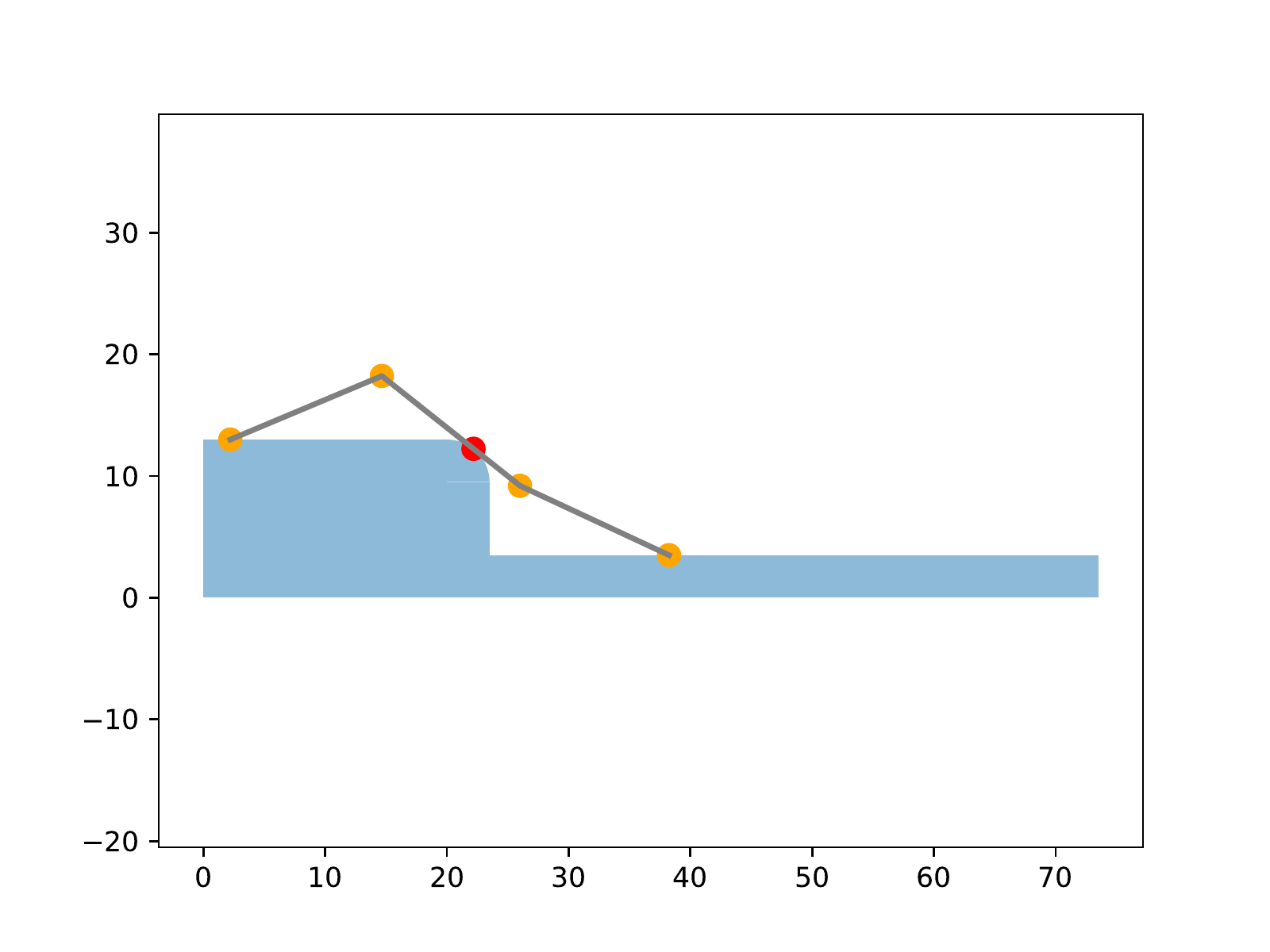}\\
	\includegraphics[width=\wuli\linewidth]{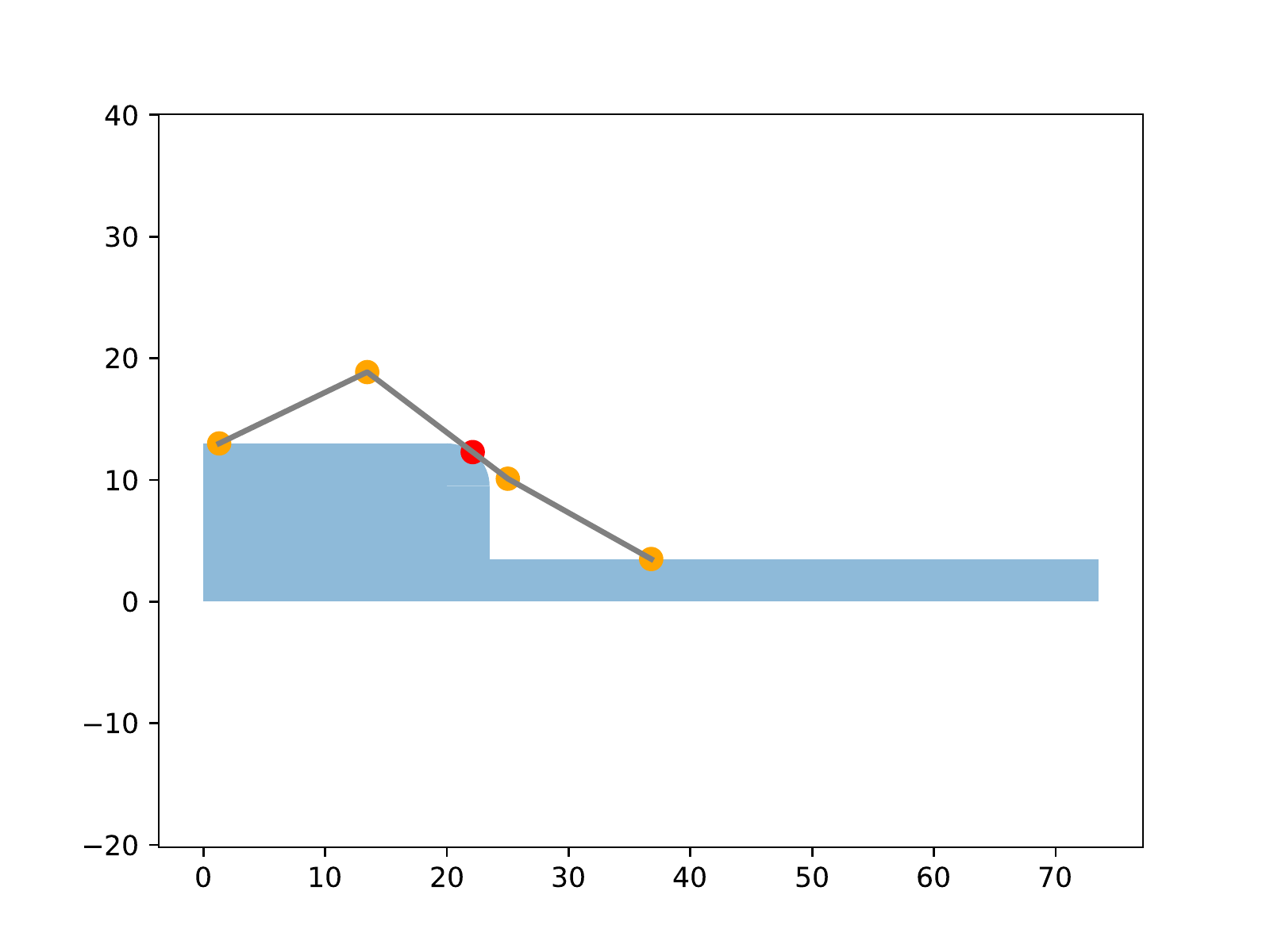}
	\includegraphics[width=\wuli\linewidth]{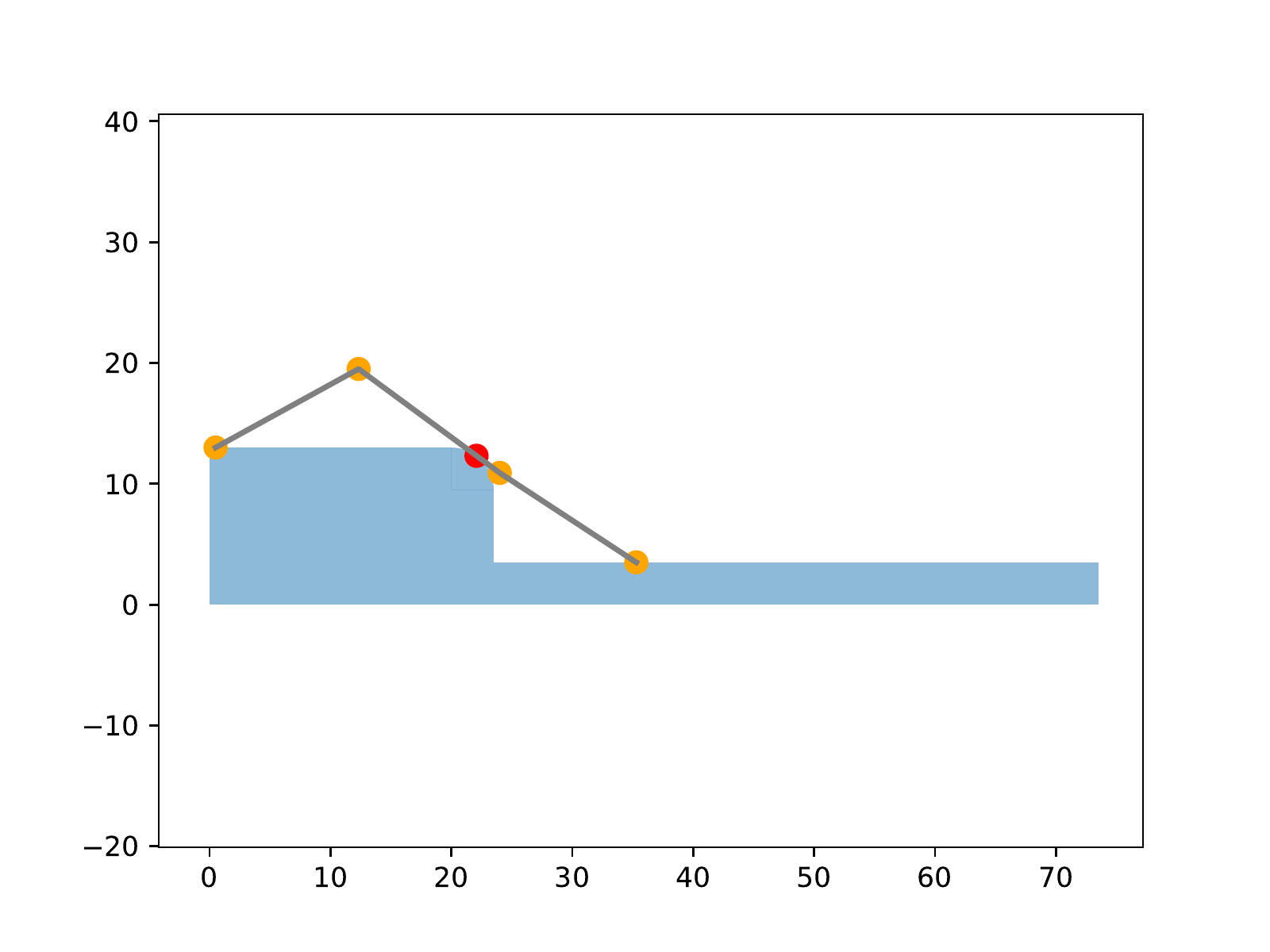}
	\includegraphics[width=\wuli\linewidth]{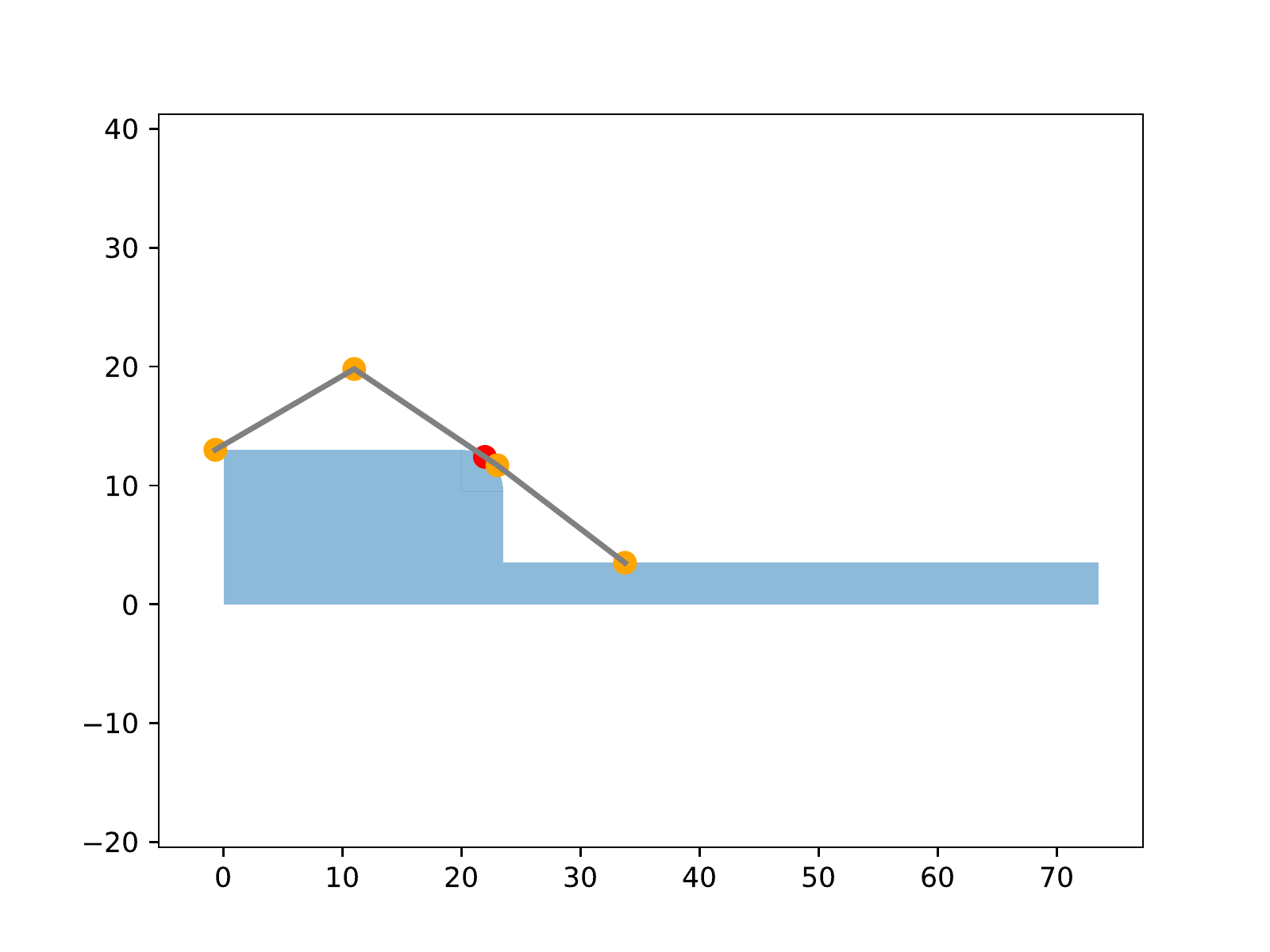}
	\includegraphics[width=\wuli\linewidth]{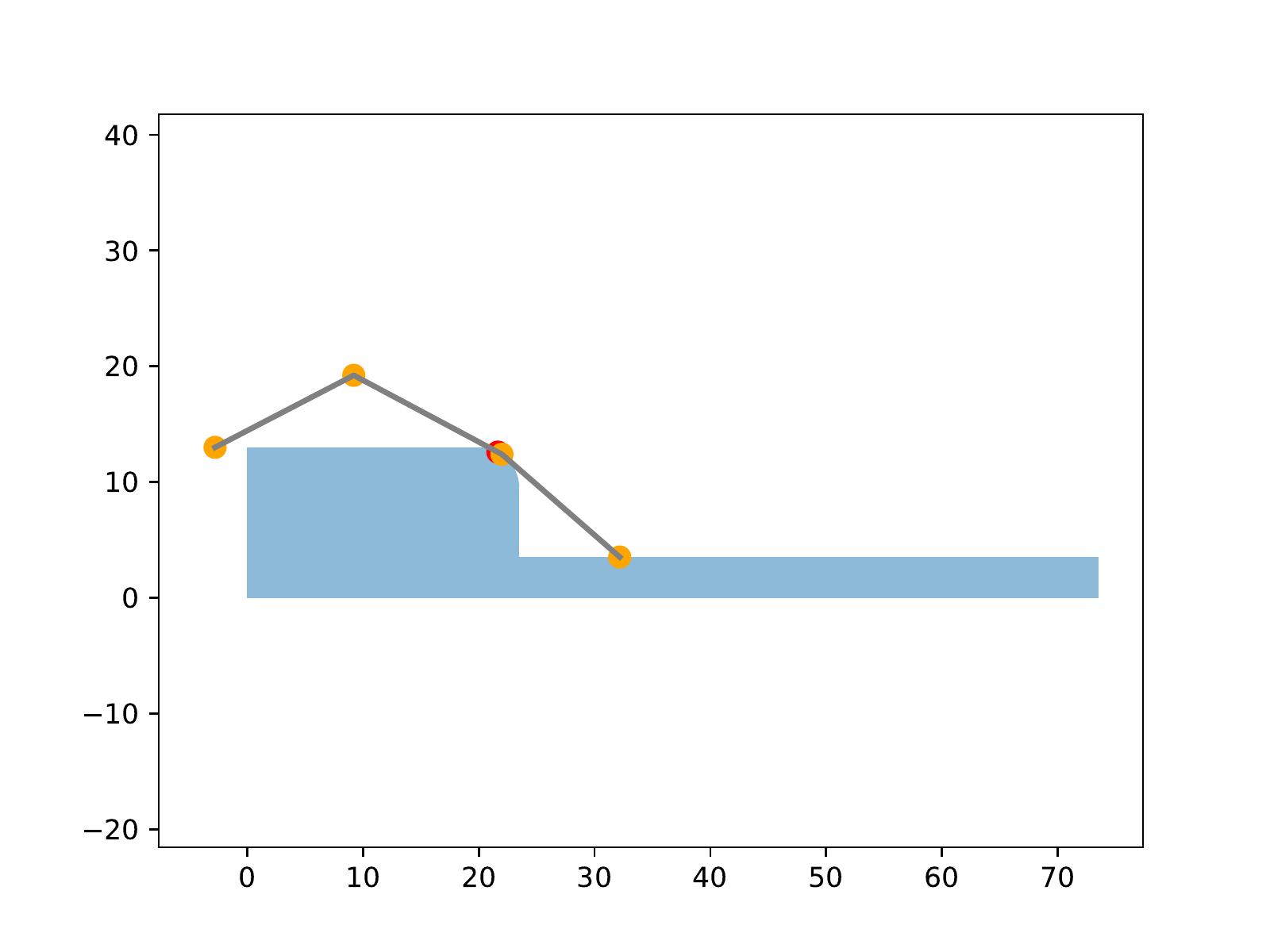}\\
	\includegraphics[width=\wuli\linewidth]{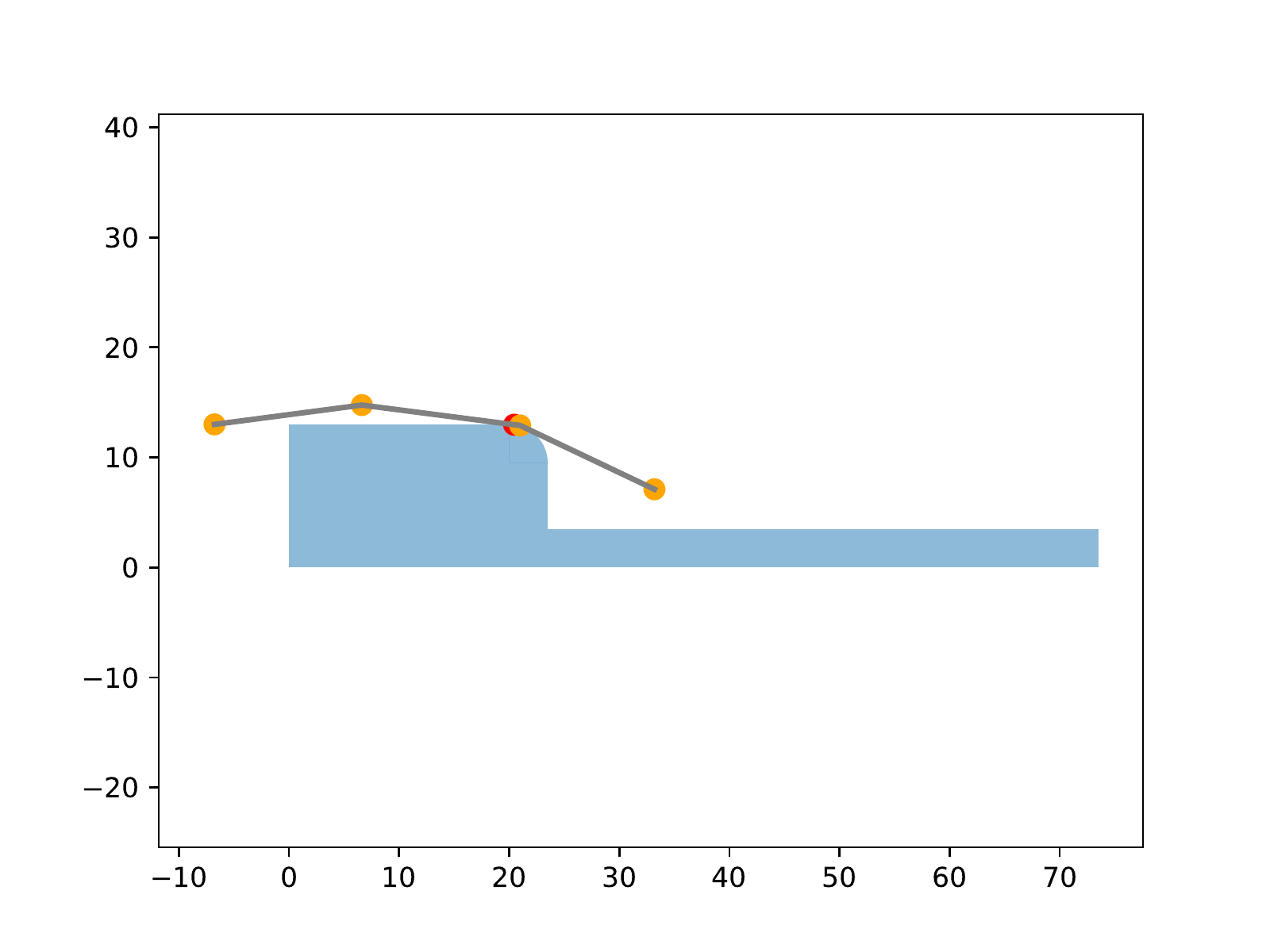}
	
	\caption{Morphology of points in path of Fig. \ref{fig:path}. The orange dot is joints, the red dot is the point that line $S1S2$ is tangent to the curve. }
	\label{fig:plot}
\end{figure}

Then with interval $0.01m$, we make a path in Fig. \ref{fig:path}. To check the correctness of each point we plot the morphology of each path point in Fig. \ref{fig:plot}.

Then we evaluate the performance of real robot path following as in Fig. \ref{fig:t}. The running of real robot is shown in the attached video.

To demonstrate the performance and safety of our algorithm, we make the robot run on three different height steps and record its $\alpha$, $\beta$, $d$ and $\theta$.

\newcommand{\sili}{0.34}
\begin{figure}[!]
	\centering
	\subfloat[]{
		\label{fig:alpha1}
		\includegraphics[width=\sili\linewidth]{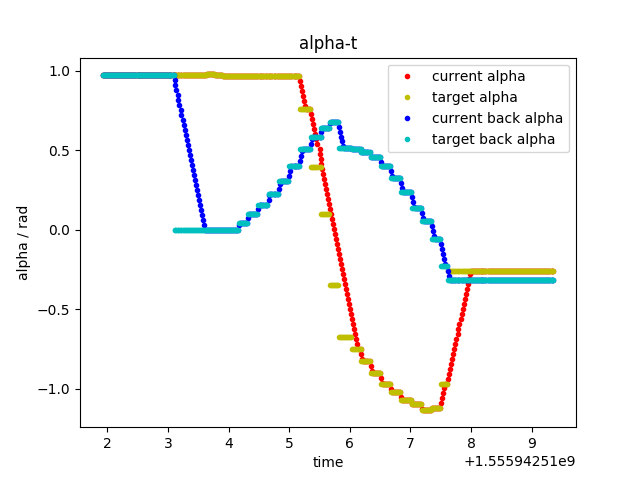}}
	\subfloat[]{
		\label{fig:alpha2}
		\includegraphics[width=\sili\linewidth]{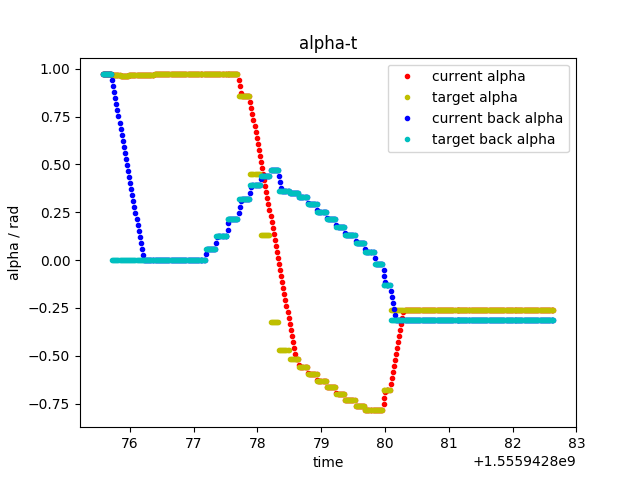}}
	\subfloat[]{
		\label{fig:alpha3}
		\includegraphics[width=\sili\linewidth]{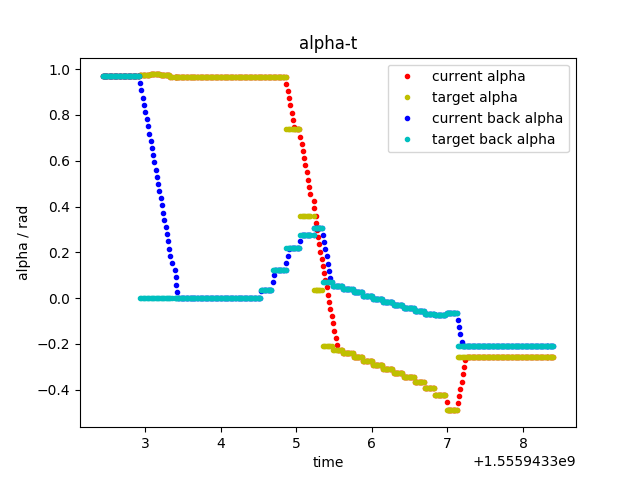}}	\\
	
	\subfloat[]{
		\label{fig:d1}
		\includegraphics[width=\sili\linewidth]{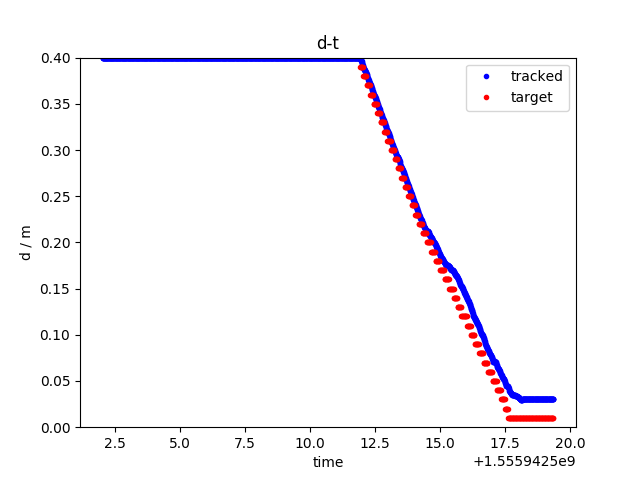}}
	\subfloat[]{
		\label{fig:d2}
		\includegraphics[width=\sili\linewidth]{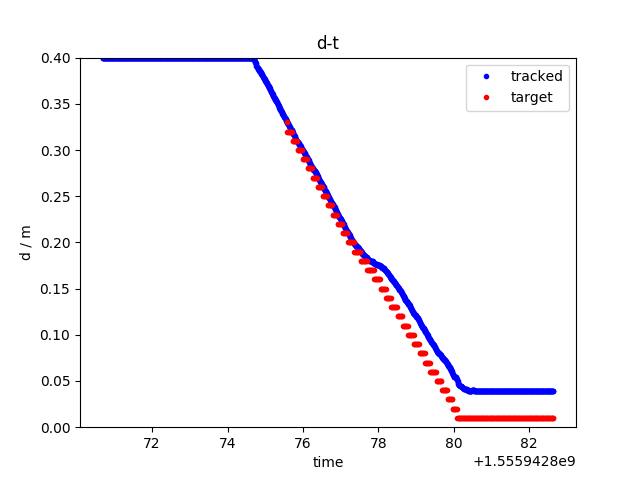}}
	\subfloat[]{
		\label{fig:d3}
		\includegraphics[width=\sili\linewidth]{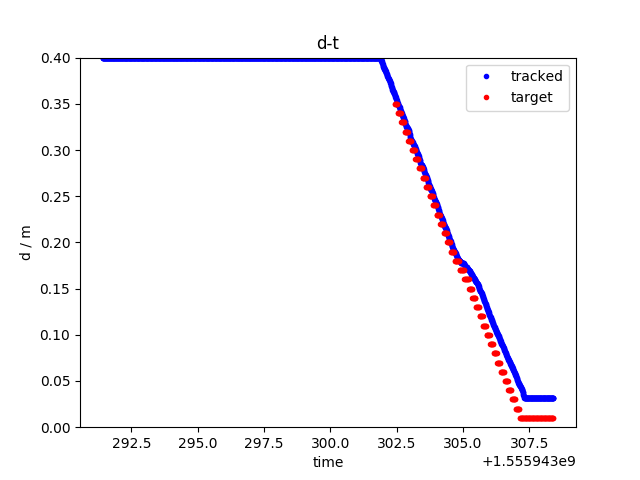}}	\\

	\subfloat[]{
		\label{fig:theta1}
		\includegraphics[width=\sili\linewidth]{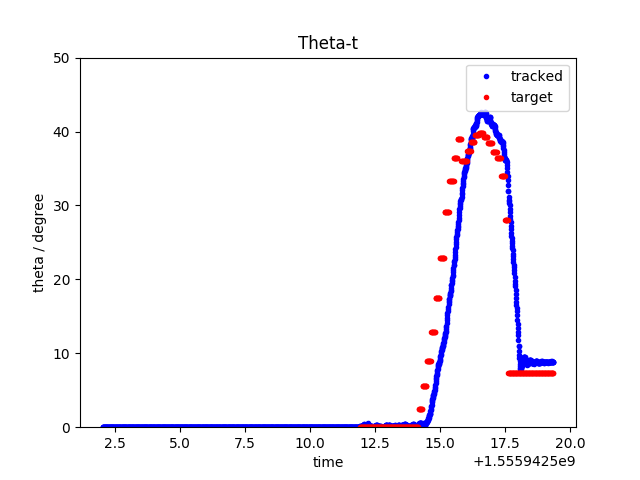}}
	\subfloat[]{
		\label{fig:theta2}
		\includegraphics[width=\sili\linewidth]{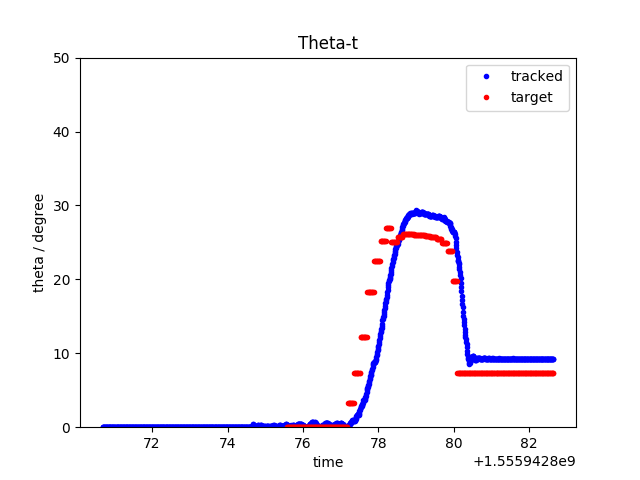}}
	\subfloat[]{
		\label{fig:theta3}
		\includegraphics[width=\sili\linewidth]{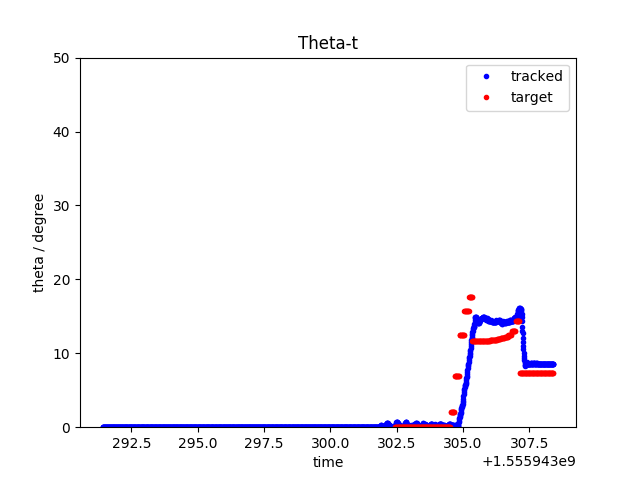}}
	\caption{The recorded and target flipper angle, $d$ and $\theta$ to $t$. The columns from left to right is with height $9.5$ cm, $6.7$ cm and $4.6$ cm.}
	\label{fig:t}
\end{figure}

In the path following, when the wheel moves adequate long for target $m$, it will trigger the next target $\alpha$, $\beta$, and $m$.

In Fig. \ref{fig:t}, the three rows are for flipper angle, $d$, robot elevation angle and elevation angle. The three column are for $9.5$ cm, $6.7$ cm and $4.6$ cm stairs, respectively. 

In the first row the current $\alpha$, $\beta$ and its targets at $t$ time are shown. We can find the current flipper angle closely follows the target angle. 

Then the second row shows the target $d$ and the tracked $d$ at each time. We find that the tracked $d$ always follows the target. However, the tracked $d$ does not reach the final target $d$, which is this way because of our implementation: the robot is following the target $m$ that is pre-assumed to be a simplified condition and only the base track will matter on movement. In addition, it do not have close loop on $d$. Thus error may happen with slip between track and ground. 

The third row shows the changing elevation angle as robot moving forward. We find that in both case, the angle drop is not sharp, we consider it make sense because in our design, while robot is on the stair edge, front flipper is required to touch the floor to avoid the sharp leaning and thus more safety for robot. However, in Fig. \ref{fig:theta3}, the target and tracked $\theta$ seems mismatched. And it is also raised by the following $m$ and non-close loop issue. To note that, the final $\theta$ is not $0$, since our final target $d$ as in Fig. \ref{fig:plot} is close but not $0$.

\begin{figure}
\subfloat[]{
	\label{fig:diffHLoc}
	\includegraphics[width=0.49\linewidth]{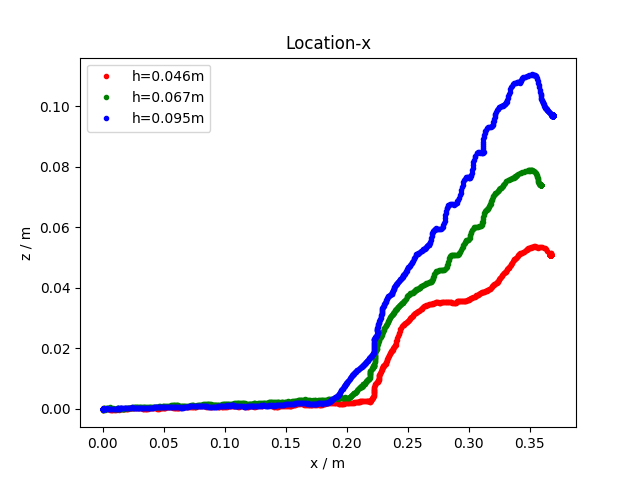}}
\subfloat[]{
	\label{fig:diffHEli}
	\includegraphics[width=0.49\linewidth]{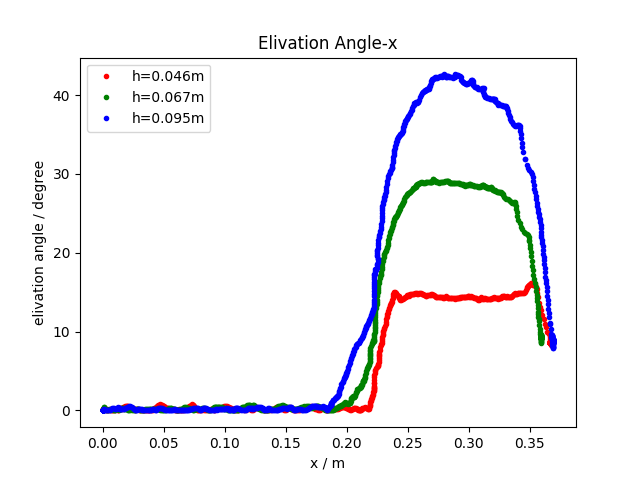}}
	\caption{Location of center and Elevation with different height.}
	\label{fig:diffH}
\end{figure}

In Fig. \ref{fig:diffH}, we shows the z-x and elevation angle-x plots for those three step cases. Here x and z is the movement and height for the center of robot (center of rectangle from the four joints , that are $S1$, $S2$ for both left and right sides, on real robot). Fig. \ref{fig:diffHLoc} demonstrates the trajectory of the center as it rises up and drop down, clearly revealed the movement of robot. Fig. \ref{fig:diffHEli} also shows the changing of elevation angle. We can find the higher step tends to require larger elevation angle and its changing is more smooth from the plot.

\begin{figure}[!]
	\includegraphics[width=0.9\linewidth]{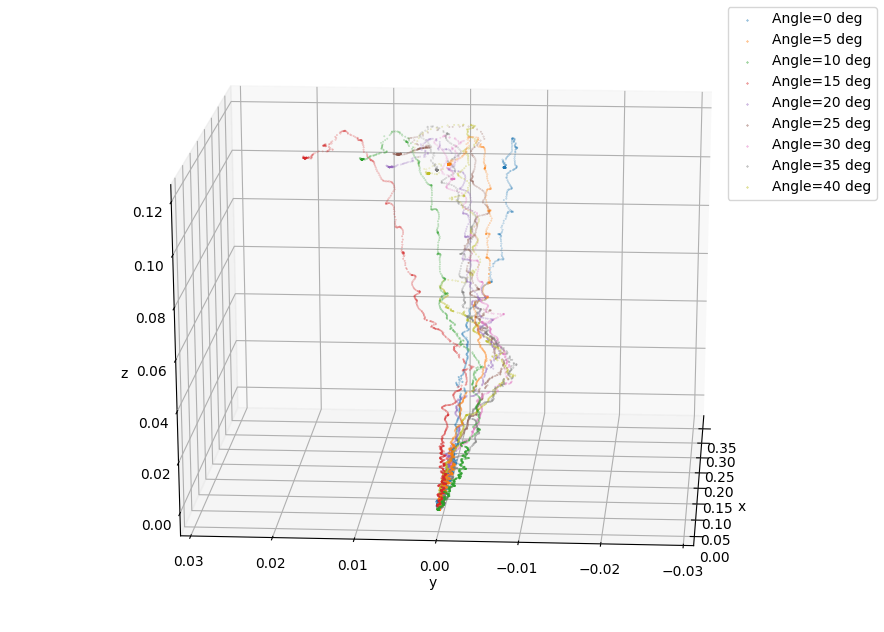}\\
	\includegraphics[width=0.9\linewidth]{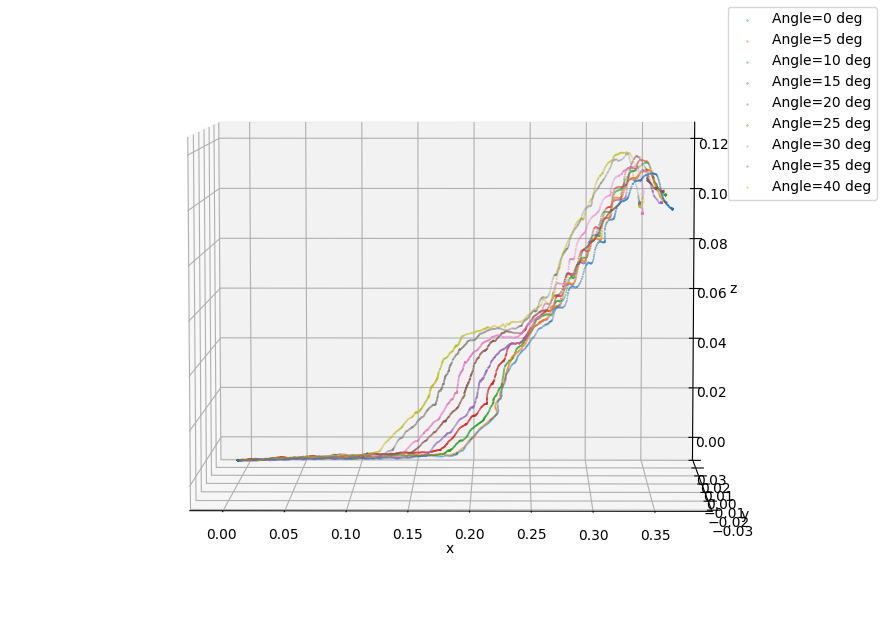}
	\caption{Location of center with different Angle. The plots are from different views.}
	\label{fig:diffAngle}
\end{figure}
To further test the mobility when stair is not perfectly straight to robot, we rotate the step with angle $\Omega$ from $5^{\circ}$ to $40^{\circ}$ and find it can also works well as shown in the video. The trajectories of the robot center are in Fig. \ref{fig:diffAngle}. They all finished the task with center on the same level of height as step. Also, the larger rotation tend to rise center earlier and end up little bit farther to the target.

\section{Conclusion}
\label{sec::conclusion}
In this work, we simplify the rescue robot control and represent its morphology with parameters $d$, $a$ and $\alpha$. We are thus able to construct the parameter space for climbing a step. This work is so far the first algorithm that can build the configuration space for rescue robots on step climbing with flippers and it allows for continuous changing of the robot morphology. From our experiment on the real robot, it can well climb on the step with the implemented path following on our obtained path. 

\section{Future Work}
This work is our first step to make global planning for the movement of tracked robot with flipper. Later on, closed loop should be considered in path following for better self-localization. Also, this work provides constraints for the morphology of rescue robot, so we will attempt to implement dynamic control on top of it. Additionally, more complicate and generalized scenes should be considered.

%\pagebreak

\bibliographystyle{IEEEtran}
\bibliography{references.bib}
\end{document}